%% file: main.tex
\documentclass[10pt,journal,compsoc]{IEEEtran}

\usepackage{amsmath,amsfonts}
\usepackage{algorithmic}
\usepackage{algorithm}
\usepackage{array}
\usepackage{textcomp}
\usepackage{stfloats}
\usepackage{url}
\usepackage{verbatim}
\usepackage{graphicx}
\usepackage{cite}
\usepackage[breaklinks=true,colorlinks=true,linkcolor=blue,citecolor=blue,urlcolor=blue,bookmarks=true,bookmarksnumbered=true]{hyperref}
\usepackage{caption}
\usepackage{mathtools}
\usepackage{amssymb}
\usepackage{multirow}
\usepackage{footmisc}
\usepackage{forest}
\usepackage{tikz}
\usepackage{tikz-3dplot}
\usetikzlibrary{shapes, arrows, arrows.meta, positioning, trees, shadows, calc, patterns, decorations.pathreplacing}
\input{preamble}

\begin{document}

\title{Advances in Global Solvers for 3D Vision}

\author{
Zhenjun Zhao,
Heng Yang,
Bangyan Liao,
Yingping Zeng,
Shaocheng Yan,\\
Yingdong Gu,
Peidong Liu,
Yi Zhou,
Haoang Li,
Javier Civera
\IEEEcompsocitemizethanks{
\IEEEcompsocthanksitem Z. Zhao and J. Civera are with University of Zaragoza, Zaragoza, Spain.\protect\\
E-mail: {\tt \{zhenjunz,jcivera\}@unizar.es}
\IEEEcompsocthanksitem H. Yang is with Harvard University, Cambridge, MA, USA.\protect\\
E-mail: {\tt hankyang@seas.harvard.edu}
\IEEEcompsocthanksitem B. Liao and P. Liu are with Westlake University, Hangzhou, China.\protect\\
E-mail: {\tt \{liaobangyan,liupeidong\}@westlake.edu.cn}
\IEEEcompsocthanksitem Y. Zeng and Y. Zhou are with Hunan University, Changsha, China.\protect\\
E-mail: {\tt \{yingpingzeng,eeyzhou\}@hnu.edu.cn}
\IEEEcompsocthanksitem S. Yan and Y. Gu are with Wuhan University, Wuhan, China.\protect\\
E-mail: {\tt \{shaochengyan,guyingdong\}@whu.edu.cn}
\IEEEcompsocthanksitem H. Li is with The Hong Kong University of Science and Technology (Guangzhou), Guangzhou, China.\protect\\
E-mail: {\tt haoangli@hkust-gz.edu.cn}
\IEEEcompsocthanksitem Corresponding author: Javier Civera
}
}

\input{sec/0_abstract}

\maketitle

\IEEEdisplaynontitleabstractindextext
\IEEEpeerreviewmaketitle

\input{sec/1_introduction}
\input{sec/2_background}
\input{sec/3_bnb}
\input{sec/4_relaxation}
\input{sec/5_gnc}
\input{sec/6_comparison}
\input{sec/7_task}
\input{sec/8_discussion}
\input{sec/9_social_impact}
\input{sec/10_conclusion}
\input{sec/acknowledgements}

{\small
\bibliographystyle{IEEEtran}
\bibliography{main}
}

\end{document}

%% file: preamble.tex
\usepackage{booktabs}
\usepackage{tabularx}
\usepackage{placeins}

\usepackage[capitalize]{cleveref}
\usepackage{xspace}
\usepackage[utf8]{inputenc}
\usepackage{newunicodechar}
\newunicodechar{∞}{\infty}
\crefname{section}{Sec.}{Secs.}
\Crefname{section}{Section}{Sections}
\Crefname{table}{Table}{Tables}
\crefname{table}{Tab.}{Tabs.}

\makeatletter
\DeclareRobustCommand\onedot{\futurelet\@let@token\@onedot}
\def\@onedot{\ifx\@let@token.\else.\null\fi\xspace}
\def\eg{\emph{e.g}\onedot,\xspace} 
\def\ie{\emph{i.e}\onedot,\xspace}

\def\etal{\emph{et al}\onedot}
\let\@authorsaddresses\@empty
\makeatother

\newcommand{\bA}{\mathbf{A}}
\newcommand{\ba}{\mathbf{a}}

\newcommand{\bbb}{\mathbf{b}}
\newcommand{\bC}{\mathbf{C}}

\newcommand{\bd}{\mathbf{d}}
\newcommand{\bE}{\mathbf{E}}

\newcommand{\bF}{\mathbf{F}}
\newcommand{\bI}{\mathbf{I}}
\newcommand{\bK}{\mathbf{K}}

\newcommand{\bM}{\mathbf{M}}
\newcommand{\bn}{\mathbf{n}}
\newcommand{\bN}{\mathbf{N}}

\newcommand{\bp}{\mathbf{p}}
\newcommand{\bQ}{\mathbf{Q}}
\newcommand{\bq}{\mathbf{q}}
\newcommand{\bR}{\mathbf{R}}

\newcommand{\bS}{\mathbf{S}}

\newcommand{\bT}{\mathbf{T}}
\newcommand{\bt}{\mathbf{t}}
\newcommand{\bV}{\mathbf{V}}
\newcommand{\bU}{\mathbf{U}}
\newcommand{\bv}{\mathbf{v}}
\newcommand{\bX}{\mathbf{X}}
\newcommand{\bx}{\mathbf{x}}
\newcommand{\bY}{\mathbf{Y}}

\newcommand{\bz}{\mathbf{z}}
\newcommand{\bZ}{\mathbf{Z}}

\newcommand{\calF}{{\cal F}}

\newcommand{\calI}{{\cal I}}
\newcommand{\calJ}{{\cal J}}

\newcommand{\nnR}{\bbR_{+}}
\newcommand{\pR}{\bbR_{++}}
\renewcommand{\int}{\mathbb{Z}}
\newcommand{\nnint}{\int_{+}}
\newcommand{\pint}{\int_{++}}

\newcommand{\zero}{{\mathbf 0}}
\newcommand{\eye}{{\mathbf I}}

\newcommand{\valpha}{\boldsymbol{\alpha}}
\newcommand{\vlambda}{\boldsymbol{\lambda}}
\newcommand{\ceil}[1]{\left\lceil #1 \right\rceil}

\newcommand{\bbR}{\mathbb{R}}
\newcommand{\bbS}{\mathbb{S}}
\newcommand{\bbI}{\mathbb{I}}

\newcommand{\vcat}{\,;\,}

\newcommand{\monoleq}[2]{\left[ #1 \right]_{#2}}
\newcommand{\dimbasis}[2]{m_{#1}(#2)}

\DeclareMathOperator*{\trace}{trace}
\DeclareMathOperator*{\rank}{rank}

\newcommand{\PAR}[1]{\vspace{0.1cm}\noindent{\bf #1} }

%% file: sec/0_abstract.tex
\IEEEtitleabstractindextext{
\begin{abstract}
Global solvers have emerged as a powerful paradigm for 3D vision, offering certifiable solutions to nonconvex geometric optimization problems traditionally addressed by local or heuristic methods.
This survey presents the first systematic review of global solvers in geometric vision, unifying the field through a comprehensive taxonomy of three core paradigms: Branch-and-Bound (BnB), Convex Relaxation (CR), and Graduated Non-Convexity (GNC).
We present their theoretical foundations, algorithmic designs, and practical enhancements for robustness and scalability, examining how each addresses the fundamental nonconvexity of geometric estimation problems.
Our analysis spans ten core vision tasks, from Wahba problem to bundle adjustment, revealing the optimality-robustness-scalability trade-offs that govern solver selection.
We identify critical future directions: scaling algorithms while maintaining guarantees, integrating data-driven priors with certifiable optimization, establishing standardized benchmarks, and addressing societal implications for safety-critical deployment.
By consolidating theoretical foundations, practical advances, and broader impacts, this survey provides a unified perspective and roadmap toward certifiable, trustworthy perception for real-world applications.
A continuously-updated literature summary and companion code tutorials are available at \href{https://github.com/ericzzj1989/Awesome-Global-Solvers-for-3D-Vision}{Awesome Global Solvers for 3D Vision}.
\end{abstract}
}

%% file: sec/1_introduction.tex
\ifCLASSOPTIONcompsoc
\IEEEraisesectionheading{\section{Introduction}\label{sec:introduction}}
\else
\section{Introduction}
\label{sec:intro}
\fi

\IEEEPARstart{G}{eometric} modeling and optimization are central challenges in computer vision and robotics, concerned with estimating model parameters that best fit observed data.
These problems underpin a wide range of applications, including visual localization~\cite{sattler2018benchmarking}, 3D reconstruction~\cite{choi2015robust}, simultaneous localization and mapping (SLAM)~\cite{campos2021orb}, structure-from-motion (SfM)~\cite{schonberger2016structure}, virtual and augmented reality~\cite{huang2024ar,chen2025neurosymbolic}, autonomous driving~\cite{geiger2012we}, and medical imaging~\cite{audette2000algorithmic}.
In practice, these tasks are often formulated as nonlinear optimization problems and solved using local methods such as Gauss-Newton~\cite{bjorck2024numerical} and Levenberg-Marquardt~\cite{more2006levenberg}.
However, the highly nonlinear and nonconvex nature of both objectives and constraints makes local methods sensitive to initialization.
Furthermore, outlier-robust estimation techniques such as consensus maximization~\cite{chin2018robust,chin2017maximum} and M-estimation~\cite{huber19811981,hampel1986} introduce additional nonconvexity and combinatorial structure, significantly complicating the optimization landscape.

Over the past two decades, global solvers have emerged as a vital research direction in 3D vision, catalyzing significant advancements.
Existing approaches can be broadly categorized into three core paradigms, namely \textbf{Branch-and-Bound (BnB)}~\cite{lawler1966branch}, \textbf{Convex Relaxation (CR)}~\cite{shor1987quadratic,lasserre2001global}, and \textbf{Graduated Non-Convexity (GNC)}~\cite{black1996unification}.
Conceptually, these methods employ distinct strategies to achieve global or near-global solutions.
\textbf{BnB} systematically explores the solution space through recursive partitioning and pruning, guaranteeing global optimality by eliminating provably suboptimal regions.
\textbf{CR methods}, including Shor's relaxation, Moment-SOS hierarchies (where SOS denotes sum-of-squares), and other specialized techniques, reformulate the original nonconvex problem as a tractable convex surrogate, whose global solution can be computed efficiently and often yields tight or certifiable bounds on the original problem.
\textbf{GNC} constructs a continuous transformation (\ie a homotopy) from an easy-to-solve convex problem to the original nonconvex one, tracking a solution path that often converges to near-global solutions in practice.
These global methods have been successfully applied and theoretically analyzed across diverse 3D vision problems, including Wahba problem~\cite{wahba1965least}, vanishing point estimation~\cite{barnard1982interpreting}, absolute and relative pose estimation~\cite{lepetit2009ep,hartley1997defense}, 3D registration~\cite{arun1987least}, rotation and translation averaging~\cite{hartley2013rotation,govindu2001combining}, triangulation~\cite{hartley1997triangulation}, pose graph optimization~\cite{lu1997globally}, and bundle adjustment~\cite{triggs1999bundle}.

Despite substantial progress, a unified perspective on global solvers in 3D vision remains elusive.
Existing surveys address only partial aspects of the field.
For example, \cite{hartley2007optimal} is limited to techniques such as $L_2 / L_{\infty}$ and BnB, with applications like relative/absolute pose estimation and triangulation.
\cite{majumdar2020recent} focuses on scalability improvements for semidefinite programming (SDP) solvers.
\cite{rosen2021advances} surveys inference and representation methods for SLAM, addressing certifiable approaches among other techniques.
\cite{carlone2023estimation} is a full monograph focused specifically on outlier-contaminated 3D problems.
\cite{rosen2026certifiably} provides a dedicated treatment of certifiable methods and their theoretical properties in SLAM.
While valuable, these works do not provide a comprehensive and up-to-date perspective that
(i) unifies global methods under a structured taxonomy,
(ii) systematically compares their theoretical guarantees and practical trade-offs,
(iii) covers applications across diverse 3D vision tasks, and
(iv) identifies open challenges and future research directions.
This lack of a consolidated view hinders both newcomers and established researchers from navigating and advancing the field.
To address these gaps, we present the first systematic survey specifically dedicated to global solvers for 3D vision.

\input{fig/taxonomy}

\PAR{Contributions.} The primary contributions of this survey are:
\begin{itemize}
    \item \textbf{A unified taxonomy.}
    We categorize global solvers into three main families based on their underlying principles: BnB, CR, and GNC.
    For each family, we review core formulations, theoretical properties, and practical variants enhanced for robustness and scalability.

    \item \textbf{Comparative analysis.}
    We provide a systematic comparison of global solvers, examining trade-offs in optimality guarantees, computational complexity and scalability, outlier robustness, and practical deployment considerations across different method families and problem scales.
    
    \item \textbf{Task-driven coverage.}
    We survey how global solvers have been applied across classical 3D vision problems, including Wahba problem, vanishing point estimation, absolute and relative pose estimation, 3D registration, rotation and translation averaging, triangulation, pose graph optimization, and bundle adjustment, highlighting representative algorithms, typical performance trade-offs, and open gaps.
    
    \item \textbf{Fundamental challenges and future directions.}
    We highlight open challenges in the field and identify research directions that bridge theory and practice.

    \item \textbf{Practical resources and reproducibility.}
    We provide a continuously-updated online repository with curated literature, implementation pointers, and tutorial code demonstrating convex relaxation formulations for representative problems using the sparse polynomial optimization toolbox (SPOT) in Python~\cite{kang2025global,kang2024fast}.
\end{itemize}

\PAR{Scope.}
We focus on general-purpose global solvers for nonconvex optimization in 3D vision, covering methods that provide certified global optimality (BnB, convex relaxations), as well as continuation-based approaches that often achieve near-global solutions in practice (GNC).
While sampling-based methods such as RANSAC~\cite{fischler1981random} and its variants~\cite{chum2003locally,raguram2012usac,barath2020magsac++} also offer a complementary ``global'' estimation paradigm (in the sense of being less sensitive to initializations), they fall outside our scope as they are not optimization-based and have been extensively surveyed elsewhere~\cite{choi2009performance,raguram2008comparative}.
Our scope encompasses foundational theories and algorithms, robustness and scalability enhancements, comparative analysis of different approaches, and practical applications across diverse 3D vision tasks.
By elucidating these aspects, this survey fills a critical knowledge gap and serves as a foundational reference for researchers and practitioners working on geometric optimization problems in 3D vision.

\PAR{Organization.}
\cref{fig:taxonomy} provides an overview of this survey's structure and the taxonomy of global solvers for 3D vision.
Following the formal background and notation in~\cref{sec:background}, we review the three primary paradigms of global solvers: BnB in~\cref{sec:bnb}, CR in~\cref{sec:relaxation}, and GNC in~\cref{sec:gnc}.
For each paradigm, we present the foundational formulation, theoretical properties, and practical enhancements for robustness and scalability.
\cref{sec:compare} presents a comparative analysis of different methods, examining the trade-offs in optimality, computational complexity and scalability, outlier robustness, and method selection and deployment both across and within solver families.
\cref{sec:app} surveys applications and tasks where global methods have been particularly influential, analyzing per-task modeling choices and algorithmic recipes.
\cref{sec:dis} discusses major challenges and highlights promising future research directions, paving the way for continued innovation in the field.
\cref{sec:social} examines the societal impact of this technology, emphasizing responsible deployment and transparent modeling practices.
Finally, \cref{sec:conclusion} summarizes the key discussions of this survey.

\medskip
\noindent
In summary, global optimization has evolved from a niche theoretical pursuit into a practical design principle for reliable geometric perception.
By bringing together theories, algorithms, and applications, this survey aims to provide a coherent roadmap for future work that closes the gap between certifiable theory and practical systems.

%% file: fig/taxonomy.tex
\begin{figure*}[t]
\centering
\scriptsize

\definecolor{solverBase}{HTML}{E67E22}
\definecolor{solverLight}{HTML}{F0B27A}
\definecolor{solverVeryLight}{HTML}{F8E6D6}

\definecolor{analysisBase}{HTML}{27AE60}
\definecolor{analysisLight}{HTML}{7DCEA0}
\definecolor{analysisVeryLight}{HTML}{D5F4E6}

\definecolor{appBase}{HTML}{3498DB}
\definecolor{appLight}{HTML}{85C1E9}
\definecolor{appVeryLight}{HTML}{D6EAF8}

\definecolor{rootColor}{HTML}{95A5A6}

\forestset{
  taxonomy/.style={
    for tree={
      grow'=0,
      parent anchor=east,
      child anchor=west,
      anchor=west,
      edge path'={(!u.parent anchor) -- +(3pt,0) |- (.child anchor)},
      edge={draw, line width=0.6pt, color=black!60},
      rounded corners=2pt,
      draw,
      minimum height=5.5mm,
      inner xsep=3.5pt,
      inner ysep=2pt,
      s sep=3pt,
      l sep=10pt,
      align=center,
      font=\small
    },
    level1/.style={draw, rounded corners=2.5pt, fill=rootColor!30, 
                   minimum height=7mm, minimum width=45mm,
                   align=center, font=\bfseries\normalsize,
                   draw=rootColor!80, line width=0.7pt},
    level2solver/.style={draw, rounded corners=2pt, fill=solverBase!60,
                         minimum height=6mm, minimum width=36mm,
                         align=center, font=\small\bfseries,
                         draw=solverBase!90, line width=0.6pt},
    level2analysis/.style={draw, rounded corners=2pt, fill=analysisBase!60,
                           minimum height=6mm, minimum width=36mm,
                           align=center, font=\small\bfseries,
                           draw=analysisBase!90, line width=0.6pt},
    level2app/.style={draw, rounded corners=2pt, fill=appBase!60,
                      minimum height=6mm, minimum width=36mm,
                      align=center, font=\small\bfseries,
                      draw=appBase!90, line width=0.6pt},
    level3solver/.style={draw, rounded corners=2pt, fill=solverLight!50,
                         minimum height=5.5mm, minimum width=45mm,
                         align=center, font=\footnotesize,
                         draw=solverBase!70, line width=0.5pt},
    level3analysis/.style={draw, rounded corners=2pt, fill=analysisLight!50,
                           minimum height=5.5mm, minimum width=45mm,
                           align=center, font=\footnotesize,
                           draw=analysisBase!70, line width=0.5pt},
    level3app/.style={draw, rounded corners=2pt, fill=appLight!50,
                      minimum height=5.5mm, minimum width=45mm,
                      align=center, font=\footnotesize,
                      draw=appBase!70, line width=0.5pt},
    level4solver/.style={draw, rounded corners=2pt, fill=solverVeryLight!60,
                         minimum height=5mm, minimum width=28mm,
                         align=center, font=\footnotesize,
                         draw=solverBase!50, line width=0.5pt}
  }
}

\resizebox{1\textwidth}{!}{
\begin{forest} taxonomy
[{Global Solvers for 3D Vision}, level1 
  [{Global Solvers\\(\S~\ref{sec:bnb}--\ref{sec:gnc})}, level2solver
    [{Branch-and-Bound\\(\S~\ref{sec:bnb})}, level3solver]
    [{Convex Relaxation\\(\S~\ref{sec:relaxation})}, level3solver
      [{Shor's\\(\S~\ref{subsec:shor})}, level4solver]
      [{Moment-SOS\\(\S~\ref{subsec:moment-sos})}, level4solver]
      [{Other Relaxations\\(\S~\ref{subsec:other_relaxation})}, level4solver]
    ]
    [{Graduated Non-Convexity\\(\S~\ref{sec:gnc})}, level3solver]
  ]
  [{Comparative Analysis\\(\S~\ref{sec:compare})}, level2analysis
    [{Optimality Guarantees\\(\S~\ref{subsec:optimality})}, level3analysis]
    [{Complexity \& Scalability\\(\S~\ref{subsec:complexity})}, level3analysis]
    [{Robustness to Outliers\\(\S~\ref{subsec:robustness})}, level3analysis]
    [{Method Selection \& Deployment\\(\S~\ref{subsec:selection})}, level3analysis]
  ]
  [{Tasks \& Applications\\(\S~\ref{sec:app})}, level2app
    [{Wahba Problem\\(\S~\ref{subsec:wahba})}, level3app]
    [{Vanishing Point Estimation\\(\S~\ref{subsec:vp})}, level3app]
    [{Absolute Pose Estimation\\(\S~\ref{subsec:pnp})}, level3app]
    [{Relative Pose Estimation\\(\S~\ref{subsec:relpose})}, level3app]
    [{3D Registration\\(\S~\ref{subsec:registration})}, level3app]
    [{Rotation Averaging\\(\S~\ref{subsec:rot-avg})}, level3app]
    [{Translation Averaging\\(\S~\ref{subsec:trans-avg})}, level3app]
    [{Triangulation\\(\S~\ref{subsec:triangulation})}, level3app]
    [{Pose Graph Optimization\\(\S~\ref{subsec:pgo})}, level3app]
    [{Bundle Adjustment\\(\S~\ref{subsec:bundle-adjustment})}, level3app]
  ]
]
\end{forest}
}

\caption{%
    Taxonomy of global solvers for 3D vision problems.
    The field is organized around three main components:
    \textbf{(1)~\textcolor{solverBase}{Global Solvers}} (\cref{sec:bnb,sec:relaxation,sec:gnc}), encompassing Branch-and-Bound, Convex Relaxation (Shor's relaxation, Moment-SOS relaxation, and other relaxation techniques), and Graduated Non-Convexity;
    \textbf{(2)~\textcolor{analysisBase}{Comparative Analysis}} (\cref{sec:compare}), examining optimality guarantees, computational complexity and scalability, robustness to outliers, and method selection and deployment; and
    \textbf{(3)~\textcolor{appBase}{Tasks and Applications}} (\cref{sec:app}), covering ten fundamental problems in 3D computer vision.
}

\label{fig:taxonomy}
\end{figure*}
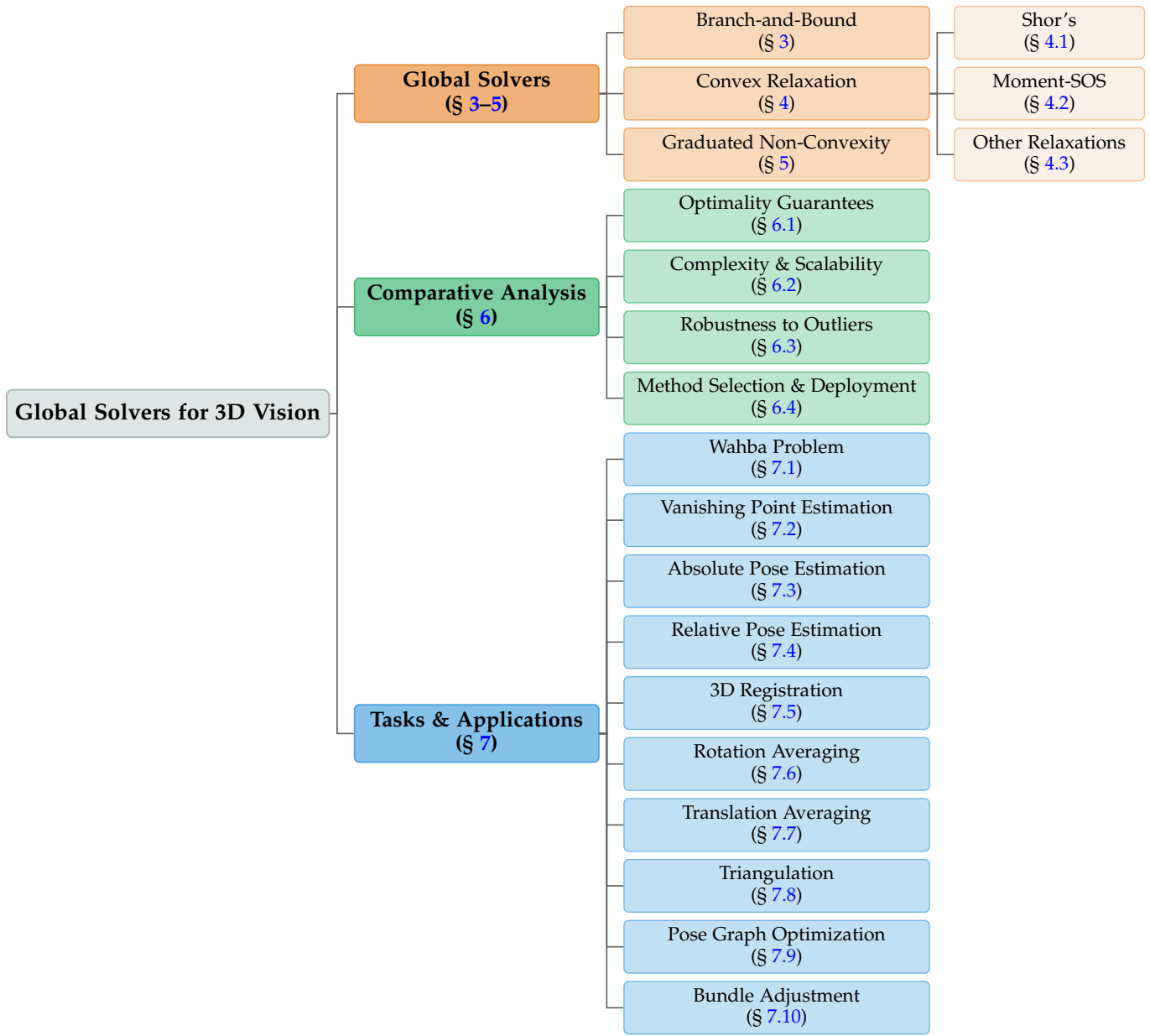

%% file: sec/2_background.tex
\section{Background and Problem Formulation}
\label{sec:background}

\input{tab/notation}

This section introduces notation, mathematical preliminaries, and canonical geometric estimation problems that underpin global methods. 
We first summarize symbols and conventions, then present typical problem formulations, highlight their intrinsic nonconvexity, and finally discuss strategies for robustness against outliers and noise.

\subsection{Notation and Conventions}
We adopt standard linear algebra notation, summarized in~\cref{tab:notation}.

\PAR{Scalars, vectors, matrices.} 
Real scalars are denoted by lowercase letters (\eg $a$), real column vectors by bold lowercase letters (\eg $\bv$), and real matrices by bold uppercase letters (\eg $\bM$).
$\eye_d$ denotes the $d \!\times\! d$ identity matrix, and $\zero$ denotes a zero vector or matrix of appropriate size.  
Let $\bM \!\in\! \bbR^{m \!\times\! n}$, and let $m_{ij}$ denote its $(i,j)$-th entry.
For index sets $\calI \!\subseteq\! [m]$ and $\calJ \!\subseteq\! [n]$, we denote by $[\bM]_{\calI,\calJ} \!\in\! \bbR^{|\calI| \!\times\! |\calJ|}$ the submatrix of $\bM$ consisting of the rows indexed by $\calI$ and the columns indexed by $\calJ$.
For a vector $\bv \!\in\! \bbR^n$, $v_i$ denotes its $i$-th entry and $\bv_\calI$ the subvector indexed by $\calI \!\subseteq\! [n]$.
The standard inner product between vectors is $\langle \ba,\bbb \rangle \!=\! \sum_i a_i b_i \!=\! \ba^\top \bbb$, and the Frobenius inner product between matrices is $\langle\bM,\bN\rangle \!=\! \sum_{i,j} m_{ij} n_{ij}$.
Norms are denoted by $\|\bv\|_1 \!=\! \sum_i |v_i|$, $\|\bv\|_2 \!=\! \sqrt{\langle \bv, \bv \rangle}$, and $\|\bM\|_F \!=\! \sqrt{\langle\bM, \bM\rangle}$.
The trace is $\trace(\bM) \!=\! \sum_i m_{ii}$, $\operatorname{vec}(\bM)$ is column-stacking vectorization, $\otimes$ is the Kronecker product, and $\operatorname{diag}([\bM_1,\dots,\bM_k])$ is block-diagonal concatenation.
Horizontal concatenation is $[\bM, \bN]$, vertical concatenation $[\bM \vcat \bN]$, and $[\bv]_\times$ denotes the skew-symmetric matrix associated with $\bv \!\in\! \bbR^3$.
For $a \!\in\! \bbR$, the symbol $\ceil{a}$ returns the smallest integer above $a$.

\PAR{Sets.} 
$\mathcal{X}$ and $\mathcal{Y}$ denote general sets.
$\bbS^n$ is the set of $n \!\times\! n$ real symmetric matrices, $\bbS^n_+$ the positive semidefinite (PSD) matrices, and $\bbS^n_{++}$ the positive definite (PD) matrices.
We write $\bX \!\succeq\! 0$ (resp. $\bX \!\succ\! 0$) for PSD (resp. PD).
The unit sphere in $\bbR^d$ is $\mathcal{S}^{d-1} \!=\! \{\bv \!\in\! \bbR^d \mid \|\bv\|_2 \!=\! 1\}$.
The $d$-dimensional special orthogonal group is $\mathrm{SO}(d) \!=\! \{\bR {\in} \bbR^{d \times d} \mid \bR^\top \bR \!=\! \eye, \det(\bR)=1\}$, and the special Euclidean group is $\mathrm{SE}(d) \!=\! \left\{ \begin{bmatrix} \bR & \bt \\ \zero^\top & 1 \end{bmatrix} \mid \bR \!\in\! \mathrm{SO}(d), \bt \!\in\! \bbR^d \right\}$.
We denote by $\mathbf{UT}^+(d)$ the set of $d \!\times\! d$ upper triangular matrices with positive diagonal entries, and $\mathcal{K} \!=\! \{ \bM \!\in\! \mathbf{UT}^+(3) \mid [\bM]_{3,3} = 1 \}$ the set of camera calibration matrices.
$\nnR$ and $\pR$ denote nonnegative and positive real numbers.
Nonnegative and positive integers are denoted $\nnint$ and $\pint$, respectively.

\PAR{Polynomials.} 
$\bbR[\bx]$ denotes the ring of real-valued polynomials in variables $\bx \!=\!(x_1,\dots,x_n)$.
Following standard notation~\cite{lasserre2009moments}, $f \!\in\! \bbR[\bx]$ is written as $f \!=\! \sum_{\valpha \in \calF} c(\valpha) \bx^{\valpha}$, where $\calF \!\subseteq\! \nnint$ is a finite set of exponents, $c(\valpha) \!\in\! \bbR$ are real coefficients, and $\bx^{\valpha} \!=\! \prod_i x_i^{\alpha_i}$ are standard monomials.
Let $\monoleq{\bx}{\calF} \!=\! \{\bx^{\valpha}: \valpha \in \calF \}$ denote the monomial basis and $\dimbasis{}{\calF}$ its dimension.
A polynomial $q(\bx)$ is a \emph{SOS} if $q(\bx) \!=\! \monoleq{\bx}{\calF}^\top \bQ \monoleq{\bx}{\calF}$ for some PSD matrix $\bQ \!\succeq\! 0$, which guarantees $q(\bx) \!\ge\! 0$ for all $\bx \!\in\! \bbR^n$.

\subsection{Geometric Problem Formulation and Structure}
Many geometric estimation problems can be abstracted as a generic nonlinear least-squares problem:
\begin{equation}
\label{eq:generic-ls}
    \min_{\bx \in \mathcal{X}} \sum_{i=1}^N r^2(\bx, y_i),
    \tag{Geometric Problem}
\end{equation}
where $\bx \!\in\! \mathcal{X}$ parametrizes the model, $\{y_i \!\in\! \mathcal{Y}\}_{i=1}^N$ are measurements, and $r: \mathcal{X} \!\times\! \mathcal{Y} \!\to\! \bbR_+$ is a scalar residual function quantifying the disagreement between the model parameters $\bx$ and measurement $y_i$.
This template captures reprojection error in absolute pose estimation, triangulation, and bundle adjustment; epipolar error in relative pose; alignment error in Wahba problem and point-set registration; and distance-based costs in vanishing point estimation and rotation/translation averaging.

\PAR{Sources of nonconvexity.}
In~\eqref{eq:generic-ls}, nonconvexity stems from two primary sources.
First, the \emph{constraint set} $\mathcal{X}$ introduces structural nonconvexity through:
(i) \emph{rotation constraints}, with nonlinear manifold $\mathrm{SO}(3)$,
(ii) \emph{unit-norm constraints}, \eg translation directions and line directions on $\mathcal{S}^{2}$,
(iii) \emph{low-rank or bilinear constraints}, \eg essential or fundamental matrices $\bE,\bF$ with rank restrictions or adjugate relations $\mathrm{adj}(\bE) \!=\! \bq \bt^\top$, and 
(iv) \emph{discrete or combinatorial variables}, \eg when selecting subsets of measurements or assigning measurements to multiple structures.
Second, the \emph{residual function} $r(\bx, y_i)$ contributes nonconvexity through nonlinear measurement models, such as perspective projection in absolute pose estimation and bundle adjustment, or angular distances in rotation averaging.

\subsection{Robust Estimation}
In practice, measurements are often incomplete, noisy, or contaminated by outliers.
Robust strategies complement canonical formulations of~\eqref{eq:generic-ls} by enhancing resilience to outliers and noise while retaining computational tractability.
Two predominant paradigms are \emph{consensus maximization}~\cite{chin2017maximum} and \emph{M-estimation}~\cite{huber19811981}.

\PAR{Consensus maximization.}
\emph{Consensus maximization} (a.k.a.\ \emph{inlier set maximization})~\cite{chin2018robust,chin2017maximum} seeks model parameters maximizing the number of measurements consistent with a predefined error threshold $\tau$:
\begin{equation}
\label{eq:cm}
    \max_{\bx \in \mathcal{X}} \sum_{i=1}^N \bbI \big(|r(\bx, y_i)| \le \tau \big),
    \tag{CM}
\end{equation}
where $\bbI$ is the indicator function that returns 1 if the condition inside is true and 0 otherwise.
\eqref{eq:cm} is combinatorial and NP-hard~\cite{chin2017maximum}.

\PAR{M-estimation.}
\emph{M-estimation}~\cite{huber19811981,hampel1986} reduces the influence of large residuals by replacing the least-squares cost with a robust function $\rho: \bbR \!\rightarrow\! \bbR_+$ of subquadratic growth:
\begin{equation}
\label{eq:m-est}
    \min_\bx \sum_{i=1}^N \rho\big(r(\bx,y_i)\big),
    \tag{M-Estimation}
\end{equation}
where typical choices for $\rho$ are Huber, Cauchy, Geman--McClure (GM), or truncated least squares (TLS)~\cite{barron2019general}.
Unlike consensus maximization, \eqref{eq:m-est} yields a continuous optimization problem, which remains challenging to solve globally due to the nonconvexity of both the robust cost $\rho$ and the feasible set $\mathcal{X}$.

%% file: tab/notation.tex
\begin{table}[t]
    \centering
    \caption{Summary of notation used throughout the survey.}
    \label{tab:notation}
    \resizebox{\columnwidth}{!}{%
        \begin{tabular}{ll}
            \toprule
            Symbol & Meaning \\
            \midrule
            $a$ & Real scalar \\
            $\bv$ & Real column vectors \\
            $\bM$ & Real matrix \\
            $\mathbf{I}_d$ & Identity matrix of size $d$\\
            $\mathbf{0}$ & All-zero vector or matrix \\
            $m_{ij}$ & Entry indexing \\
            $[\bM]_{I,J}$ & Submatrix indexing \\
            $\langle \cdot,\cdot\rangle$ & Inner product \\
            $\|\bv\|_1,\|\bv\|_2$ & Vector $\ell_1$, $\ell_2$ norms \\
            $\|\bM\|_F$ & Frobenius norm \\
            $\trace(\bM)$ & Trace \\
            $\det(\bM)$ & Determinant \\
            $\rank(\bM)$ & Rank \\
            $\mathrm{adj}(\bM)$ & Adjugate \\
            $\operatorname{vec}(\bM)$ & Column-stacking operator \\
            $\operatorname{diag}([\bM_1,\dots,\bM_k])$ & Block-diagonal \\
            $\otimes$ & Kronecker product \\
            $[\bM, \bN]$ & Horizontal concatenation \\
            $[\bM \vcat \bN]$ & Vertical concatenation \\
            $[\bv]_\times$ & Skew-symmetric operator: $ \bbR^{3} \to \bbR^{3\times 3} $ \\
            $\ceil{a}$ & Smallest integer above $a$ \\
            $\bbS^n,\,\bbS^n_+,\,\bbS^n_{++}$ & Symmetric, PSD, PD matrices \\
            $\bX \!\succeq\! 0$ & Positive semidefinite constraint \\
            $\bX \!\succ\! 0$ & Positive definite constraint \\
            $\mathcal{S}^{d-1}$ & Unit sphere in $\bbR^d$ \\
            $\mathrm{SO}(d)$ & Special orthogonal group (rotations) \\
            $\mathrm{SE}(d)$ & Special Euclidean group (rigid motions) \\
            $\mathbf{UT}^+(d)$ & Upper triangular matrices with positive diagonal \\
            $\mathcal{K}$ & Camera calibration matrices \\
            $\nnR$, $\pR$ & Nonnegative, positive real numbers \\
            $\nnint$, $\pint$ & Nonnegative, positive integers \\
            $\bbR[x]$ & Polynomial ring \\
            $\monoleq{\bx}{\calF}$ & Monomial basis \\
            $q(\bx) \!=\! \monoleq{\bx}{\calF}^\top \bQ \monoleq{\bx}{\calF}$, $\bQ\succeq 0$ & SOS polynomial \\
            $\tau$ & Inlier threshold for robust estimation \\
            $\varepsilon$ & Algorithm convergence tolerance \\
            \bottomrule
        \end{tabular}
    }
\end{table}

%% file: sec/3_bnb.tex
\section{Global Solver 1: Branch-and-Bound (BnB)}
\label{sec:bnb}

\emph{Branch-and-Bound (BnB)}~\cite{lawler1966branch,clausen1999branch,land2009automatic} is a deterministic exhaustive search framework for global optimization in nonconvex problems.
Unlike sampling-based methods (\eg RANSAC~\cite{fischler1981random}) that provide probabilistic guarantees, BnB systematically partitions the search space (\emph{branching}) and computes provable bounds on the objective within each partition (\emph{bounding}), discarding regions that provably cannot contain the global optimum (\emph{pruning}).
Under mild assumptions, validity of bounds and sufficient refinement, BnB converges to a certified global optimum with $\varepsilon$-suboptimality guarantees.

\PAR{Modeling and solution ingredients.}
A BnB solver for geometric vision requires three core components:

\begin{itemize}
    \item \textbf{Search domain and parameterization.}  
    The domain $\mathcal{X}$ represents the parameter space: rotations $\mathrm{SO}(3)$, translations $\bbR^3$, or poses $\mathrm{SE}(3)$.
    Parameterization critically affects bound tightness and branching efficiency. 
    Common choices include angle-axis~\cite{hartley2009bnb}, and axis-aligned bounding boxes for translations~\cite{bazin2012globally,yang2013go}.
    The initial search region $\mathcal{X}_0$ is typically chosen to cover all plausible solutions based on problem constraints or data distribution (\eg $[-\pi,\pi]$ for angular parameters).
    
    \item \textbf{Bounding functions.}
    The core of BnB lies in constructing tight and efficiently computable lower bounds for the objective over each subdomain.
    Key techniques include:
    (i) Lipschitz continuity~\cite{li20073d},
    (ii) interval arithmetic and interval stabbing~\cite{de2008computational,parra2015guaranteed,bustos2017guaranteed},
    (iii) convex relaxations~\cite{ivanov2025fast}, and
    (iv) geometric bounds~\cite{hartley2009bnb,yang2013go,yang2015go}.
    The bound quality determines pruning effectiveness: tighter bounds enable earlier elimination of suboptimal regions.

    \item \textbf{Branching and node selection.}  
    Branching strategies include uniform bisection~\cite{straub2017efficient}, adaptive splitting~\cite{zhang2024accelerating,zheng2011deterministically}, and problem-specific heuristics~\cite{lian2016efficient}.
    Node selection policies include best-first, depth-first, and hybrid schemes~\cite{cai2018deterministic,cai2019consensus}, balancing early pruning and memory efficiency.
\end{itemize}

\input{fig/bnb}

\cref{fig:bnb-algorithm} illustrates the BnB search process on a one-dimensional example, showing how branching recursively subdivides the search space, bounding computes provable bounds on each region, and pruning eliminates regions that provably cannot contain the global optimum.

\PAR{Theoretical properties.}
When valid lower bounds are employed and branching refines subdomains to arbitrarily small diameters, BnB guarantees convergence to a global optimum (or within any prescribed tolerance $\varepsilon$).
The correctness follows from the monotonicity of bounds: pruning never eliminates the region containing the true minimizer, as any region whose lower bound exceeds the current best feasible solution cannot harbor the global optimum.
Consequently, BnB naturally provides \emph{a posteriori} optimality certificates once the bound gap falls below a threshold.
However, the worst-case complexity is exponential in the dimension of the model parameter space, while the actual performance strongly depends on bound tightness and problem structure.

\PAR{Computational complexity.}
The worst-case complexity of BnB is exponential in the parameter dimension.
Specifically, to achieve tolerance $\varepsilon$ in a $d$-dimensional domain $\mathcal{X}_0$ with diameter $D$, the number of subregions explored can grow as $O((D/\delta)^d)$, where $\delta \!\sim\! \varepsilon$ is the required subdivision resolution.
This corresponds to branching depth $O\big(\log(D/\varepsilon)\big)$ with branching factor $2^d$ per level, yielding $O\big(2^{d \log(D/\varepsilon)}\big) \!=\! O\big((D/\varepsilon)^d\big)$ nodes in the worst case.
In practice, tight bounding functions and problem structure (such as low intrinsic dimensionality or smoothness) enable aggressive pruning, substantially reducing the explored search tree.

\PAR{Robust estimation via consensus maximization.}
A major extension of BnB addresses robust estimation under outliers through consensus maximization~\eqref{eq:cm}.
BnB computes \emph{upper bounds} on the maximum inlier count over each subdomain $\mathcal{X}$ via interval arithmetic: for each measurement $y_i$, compute the interval $[r_i^{\text{lb}}, r_i^{\text{ub}}]$ bounding $r(\mathbf{x}, y_i)$ for all $\mathbf{x} \in \mathcal{X}$.
Then, an upper bound on the inlier count is:
\begin{equation*}
    \sum_{i=1}^N \bbI\bigl(r_i^{\text{lb}} \leq \tau\bigr) 
    \geq \max_{\mathbf{x} \in \mathcal{X}} 
    \sum_{i=1}^N \bbI\bigl(r(\mathbf{x}, y_i) \leq \tau\bigr).
\end{equation*}
A lower bound is obtained by evaluating the inlier count at any point in $\mathcal{X}$ (\eg the center).
Regions whose upper bound falls below the current best lower bound are pruned.
Despite its combinatorial nature, efficient bounding and adaptive subdivision enable globally optimal inlier set identification for moderate-scale problems~\cite{chin2017maximum,cai2019consensus}.

\PAR{Hybrid frameworks.}
To mitigate exponential complexity, many solvers integrate BnB with local refinement, trading strict global optimality for improved scalability.

\begin{itemize}
    \item \textbf{BnB initialization.}  
    A coarse BnB search over a low-dimensional subspace (\eg rotation-only) provides a high-quality initialization for nonlinear least-squares refinement~\cite{brown2019family,campbell2019alignment}.
    
    \item \textbf{Iterative alternation.}  
    Alternating between local optimization (to update the upper bound) and BnB search (to refine the lower bound) progressively tightens the bound gap~\cite{yang2013go,wu2024scalable}.
\end{itemize}

\PAR{Strengths and limitations.}

\begin{itemize}
    \item \textbf{Strengths.}
    BnB offers three key advantages:
    \begin{enumerate}
        \item \emph{Certified global optimality}: The algorithm provides provable $\varepsilon$-suboptimality certificates upon termination.
        \item \emph{Deterministic convergence}: Unlike randomized methods, BnB's exhaustive search guarantees finding the optimum given sufficient computation.
        \item \emph{Natural robustness integration}: The framework extends seamlessly to consensus maximization without fundamental algorithmic changes.
    \end{enumerate}

    \item \textbf{Limitations.}
    The primary limitation is \emph{exponential scaling} with the dimension of the model parameter space and tight tolerance requirements.
    Bound construction often demands problem-specific geometric analysis, limiting plug-and-play applicability.
    In hybrid designs, reliance on local refinement or coarse initialization may partially sacrifice strict optimality guarantees, though practical performance often remains excellent.
    Finally, high outlier ratios ($>50\%$) in consensus problems can lead to large search trees, necessitating adaptive strategies or hybrid approaches.
\end{itemize}

%% file: fig/bnb.tex
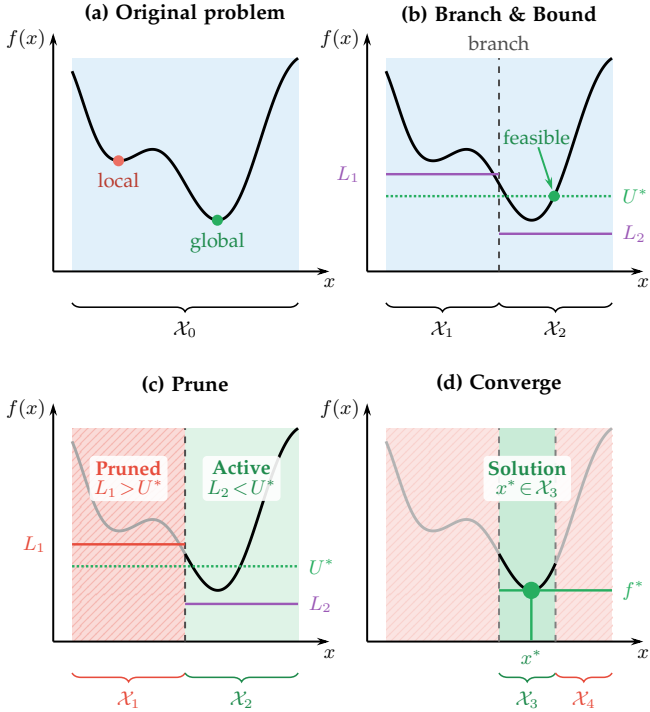
\begin{figure}[t]
\centering

\definecolor{domainColor}{HTML}{3498DB}
\definecolor{relaxColor}{HTML}{E67E22}
\definecolor{prunedColor}{HTML}{E74C3C}
\definecolor{activeColor}{HTML}{27AE60}
\definecolor{boundColor}{HTML}{9B59B6}

\resizebox{\columnwidth}{!}{
\begin{tikzpicture}[
    >={Stealth[length=2mm, width=1.2mm]},
    font=\footnotesize,
    every node/.style={font=\footnotesize}
]

\def\func#1{ 
    -0.7*sin(3*#1 r) + 0.18*cos(3*#1 r) + 
    0.15*exp(-5*(#1-0.515)*(#1-0.515)) + 
    0.5*exp(-5*(#1+1.25)*(#1+1.25)) + 
    1.304*exp(-5*(#1+1.8)^2) + 
    0.4*#1*#1 + 0.15*#1 + 1.178 
}

\def\hspacing{5.0}
\def\vspacing{5.9}
\def\xmin{-1.8}
\def\xmax{1.8}
\def\ymin{0}
\def\ymax{3.4}

\begin{scope}[shift={(0,\vspacing)}]

    \fill[domainColor!15] (\xmin, 0) rectangle (\xmax, \ymax);
    
    \draw[->, line width=0.9pt] (\xmin-0.3, 0) -- (\xmax+0.5, 0);
    \draw[->, line width=0.9pt] (\xmin-0.3, 0) -- (\xmin-0.3, \ymax+0.4);
    
    \node[below, font=\small] at (\xmax+0.5, 0) {$x$};
    \node[left, font=\small] at (\xmin-0.3, \ymax+0.3) {$f(x)$};
    
    \draw[line width=1.3pt, black] plot[domain=\xmin:\xmax, samples=100, smooth] (\x, {\func{\x}});

    \fill[activeColor] (0.515, {\func{0.515}}) circle (2.5pt);
    \node[below=2pt, font=\small, activeColor!80!black] at (0.515, {\func{0.515}}) {global};
    \fill[prunedColor!80] (-1.06, {\func{-1.06}}) circle (2.5pt);
    \node[below=2pt, font=\small, prunedColor!80!black] at (-1.06, {\func{-1.06}}) {local};
    
    \draw[decorate, decoration={brace, amplitude=4pt, mirror}, line width=0.7pt] 
        (\xmin, -0.5) -- (\xmax, -0.5) node[midway, below=5pt, font=\small] {$\mathcal{X}_0$};
    
    \node[font=\normalsize\bfseries] at ({(\xmin+\xmax)/2}, \ymax+0.7) {(a) Original problem};
\end{scope}

\begin{scope}[shift={(\hspacing,\vspacing)}]

    \fill[domainColor!15] (\xmin, 0) rectangle (0, \ymax);
    \fill[domainColor!15] (0, 0) rectangle (\xmax, \ymax);
    
    \draw[->, line width=0.9pt] (\xmin-0.3, 0) -- (\xmax+0.5, 0);
    \draw[->, line width=0.9pt] (\xmin-0.3, 0) -- (\xmin-0.3, \ymax+0.4);
    
    \node[below, font=\small] at (\xmax+0.5, 0) {$x$};
    \node[left, font=\small] at (\xmin-0.3, \ymax+0.3) {$f(x)$};
    
    \draw[dashed, line width=0.9pt, black!70] (0, 0) -- (0, \ymax);
    \node[above, font=\small, black!70] at (0, \ymax) {branch};
    
    \draw[line width=1.3pt, black] plot[domain=\xmin:\xmax, samples=100, smooth] 
        (\x, {\func{\x}});
    
    \def\LBleft{1.55}
    \def\LBright{0.6}
    
    \draw[line width=1.1pt, boundColor] (\xmin, \LBleft) -- (0, \LBleft);
    \draw[line width=1.1pt, boundColor] (0, \LBright) -- (\xmax, \LBright);
    
    \node[left, font=\small, boundColor] at (\xmin-0.3, \LBleft) {$L_1$};
    \node[right, font=\small, boundColor] at (\xmax+0.05, \LBright) {$L_2$};
    
    \def\UBbest{1.2}
    \draw[line width=1.1pt, activeColor, densely dotted] 
        (\xmin, \UBbest) -- (\xmax, \UBbest);
    \node[right, font=\small, activeColor] at (\xmax+0.05, \UBbest) {$U^*$};
    
    \pgfmathsetmacro{\xfeas}{0.8789}
    \fill[activeColor] (\xfeas, {\func{\xfeas}}) circle (2.5pt);
    \draw[->, activeColor, line width=0.9pt] (0.626, 1.98) -- (0.848, 1.3);
    \node[font=\small, activeColor!80!black] at (0.6, 2.15) {feasible};
    
    \draw[decorate, decoration={brace, amplitude=4pt, mirror}, line width=0.7pt] 
        (\xmin, -0.5) -- (0, -0.5) 
        node[midway, below=5pt, font=\small] {$\mathcal{X}_1$};
    \draw[decorate, decoration={brace, amplitude=4pt, mirror}, line width=0.7pt] 
        (0, -0.5) -- (\xmax, -0.5) 
        node[midway, below=5pt, font=\small] {$\mathcal{X}_2$};
    
    \node[font=\normalsize\bfseries] at ({(\xmin+\xmax)/2}, \ymax+0.7) 
        {(b) Branch \& Bound};
\end{scope}

\begin{scope}[shift={(0,0)}]

    \fill[prunedColor!20] (\xmin, 0) rectangle (0, \ymax);
    \fill[pattern=north east lines, pattern color=prunedColor!40] 
        (\xmin, 0) rectangle (0, \ymax);
    
    \fill[activeColor!15] (0, 0) rectangle (\xmax, \ymax);
    
    \draw[->, line width=0.9pt] (\xmin-0.3, 0) -- (\xmax+0.5, 0);
    \draw[->, line width=0.9pt] (\xmin-0.3, 0) -- (\xmin-0.3, \ymax+0.4);
    
    \node[below, font=\small] at (\xmax+0.5, 0) {$x$};
    \node[left, font=\small] at (\xmin-0.3, \ymax+0.3) {$f(x)$};
    
    \draw[dashed, line width=0.9pt, black!70] (0, 0) -- (0, \ymax);
    
    \draw[line width=1.3pt, black!35] plot[domain=\xmin:0, samples=50, smooth] 
        (\x, {\func{\x}});
    \draw[line width=1.3pt, black] plot[domain=0:\xmax, samples=50, smooth] 
        (\x, {\func{\x}});
    
    \def\LBleft{1.55}
    \def\LBright{0.6}
    
    \draw[line width=1.1pt, prunedColor] (\xmin, \LBleft) -- (0, \LBleft);
    \node[left, font=\small, prunedColor] at (\xmin-0.3, \LBleft) {$L_1$};
    
    \draw[line width=1.1pt, boundColor] (0, \LBright) -- (\xmax, \LBright);
    \node[right, font=\small, boundColor] at (\xmax+0.05, \LBright) {$L_2$};
    
    \def\UBbest{1.2}
    \draw[line width=1.1pt, activeColor, densely dotted] 
        (\xmin, \UBbest) -- (\xmax, \UBbest);
    \node[right, font=\small, activeColor] at (\xmax+0.05, \UBbest) {$U^*$};
    
    \node[font=\small, align=center, prunedColor, fill=white, inner sep=2pt, 
      fill opacity=0.9, text opacity=1, rounded corners=2pt] 
        at (-0.9, 2.6) {\textbf{Pruned}\\[-1pt]$L_1 \!>\! U^*$};
    \node[font=\small, align=center, activeColor!80!black, fill=white, inner sep=2pt, 
      fill opacity=0.9, text opacity=1, rounded corners=2pt] 
        at (0.9, 2.6) {\textbf{Active}\\[-1pt]$L_2 \!<\! U^*$};
    
    \draw[decorate, decoration={brace, amplitude=4pt, mirror}, 
          line width=0.7pt, prunedColor] 
        (\xmin, -0.5) -- (0, -0.5) 
        node[midway, below=5pt, font=\small, prunedColor] {$\mathcal{X}_1$};
    \draw[decorate, decoration={brace, amplitude=4pt, mirror}, 
          line width=0.7pt, activeColor!80!black] 
        (0, -0.5) -- (\xmax, -0.5) 
        node[midway, below=5pt, font=\small, activeColor!80!black] {$\mathcal{X}_2$};
    
    \node[font=\normalsize\bfseries] at ({(\xmin+\xmax)/2}, \ymax+0.7) 
        {(c) Prune};
\end{scope}

\begin{scope}[shift={(\hspacing,0)}]

    \fill[prunedColor!15] (\xmin, 0) rectangle (0, \ymax);
    \fill[pattern=north east lines, pattern color=prunedColor!25] 
        (\xmin, 0) rectangle (0, \ymax);
    
    \fill[activeColor!25] (0, 0) rectangle (0.9, \ymax);
    
    \fill[prunedColor!15] (0.9, 0) rectangle (\xmax, \ymax);
    \fill[pattern=north east lines, pattern color=prunedColor!25] 
        (0.9, 0) rectangle (\xmax, \ymax);
    
    \draw[->, line width=0.9pt] (\xmin-0.3, 0) -- (\xmax+0.5, 0);
    \draw[->, line width=0.9pt] (\xmin-0.3, 0) -- (\xmin-0.3, \ymax+0.4);
    
    \node[below, font=\small] at (\xmax+0.5, 0) {$x$};
    \node[left, font=\small] at (\xmin-0.3, \ymax+0.3) {$f(x)$};
    
    \draw[dashed, line width=0.9pt, black!50] (0, 0) -- (0, \ymax);
    \draw[dashed, line width=0.9pt, black!50] (0.9, 0) -- (0.9, \ymax);
    
    \draw[line width=1.3pt, black!30] plot[domain=\xmin:0, samples=75, smooth] 
        (\x, {\func{\x}});
    \draw[line width=1.3pt, black] plot[domain=0:0.9, samples=75, smooth] 
        (\x, {\func{\x}});
    \draw[line width=1.3pt, black!30] plot[domain=0.9:\xmax, samples=30, smooth] 
        (\x, {\func{\x}});
    
    \def\xopt{0.515}
    \def\yopt{0.825}
    \pgfmathsetmacro{\yopt}{\func{\xopt}}
    
    \draw[activeColor, line width=1.1pt] (\xopt, 0.02) -- (\xopt, \yopt);
    
    \fill[activeColor] (\xopt, \yopt) circle (3.5pt);
    \draw[activeColor, line width=0.8pt] (\xopt, \yopt) circle (3.5pt);
    
    \draw[line width=1.1pt, activeColor] (0, \yopt) -- (\xmax, \yopt);
    \node[right, font=\small, activeColor] at (\xmax+0.05, \yopt) {$f^*$};

    \node[below, font=\small, activeColor!80!black] at (\xopt, -0.01) {$x^*$};
    
    \node[font=\small, align=center, activeColor!80!black, fill=white, inner sep=2pt, 
      fill opacity=0.9, text opacity=1, rounded corners=2pt] 
        at (0.45, 2.6) {\textbf{Solution}\\[-1pt]$x^* \!\in\! \mathcal{X}_3$};
    
    \draw[decorate, decoration={brace, amplitude=4pt, mirror}, 
          line width=0.7pt, activeColor!80!black] 
        (0, -0.5) -- (0.9, -0.5) 
        node[midway, below=5pt, font=\small, activeColor!80!black] {$\mathcal{X}_3$};
    \draw[decorate, decoration={brace, amplitude=4pt, mirror}, 
          line width=0.7pt, prunedColor] 
        (0.9, -0.5) -- (\xmax, -0.5) 
        node[midway, below=5pt, font=\small, prunedColor] {$\mathcal{X}_4$};
    
    \node[font=\normalsize\bfseries] at ({(\xmin+\xmax)/2}, \ymax+0.7) 
        {(d) Converge};
\end{scope}

\end{tikzpicture}
}
\caption{%
    Branch-and-Bound (BnB) algorithm on a one-dimensional non-convex minimization problem.
    (a)~The original problem: minimize $f(x)$ over domain $\mathcal{X}_0$.
    The function has a \textcolor{activeColor}{global minimum} and a \textcolor{prunedColor}{local minimum}.
    (b)~\textit{Branching} partitions the domain into subregions $\mathcal{X}_1$ and $\mathcal{X}_2$, while \textit{bounding} computes lower bounds $L_1$ and $L_2$ for the respective regions and an upper bound $U^*$ from a feasible point.
    (c)~\textit{Pruning} eliminates region $\mathcal{X}_1$ since $L_1 \!>\! U^*$, while region $\mathcal{X}_2$ remains active since $L_2 \!<\! U^*$.
    (d)~After further iterations, the algorithm converges to the global optimum $x^*$ with certified optimal value $f^*$.
}
\label{fig:bnb-algorithm}
\end{figure}

%% file: sec/4_relaxation.tex
\section{Global Solver 2: Convex Relaxation (CR)}
\label{sec:relaxation}

\emph{Convex relaxation} transforms nonconvex optimization problems into convex surrogates that can be solved to global optimality, with certificates of optimality or provable bounds on suboptimality.
The core strategy is to reformulate the problem in an approximate, simpler version where nonconvex constraints are relaxed to convex ones, yielding tractable convex programs.

Two foundational relaxation techniques dominate the landscape:
\textbf{Shor's relaxation}~\cite{shor1987quadratic}, discussed in~\cref{subsec:shor}, handles quadratically constrained quadratic programs (QCQPs) via matrix lifting and rank relaxation, while
\textbf{Moment-SOS relaxation}~\cite{lasserre2001global}, detailed in~\cref{subsec:moment-sos}, addresses polynomial optimization through SOS certificates and moment matrices.
Both techniques have been extensively applied in 3D vision, providing certifiable solutions for rotation averaging, pose estimation, registration, and other fundamental problems.
Beyond these unified frameworks, a variety of specialized relaxation techniques exploiting problem-specific structure for efficiency are discussed in~\cref{subsec:other_relaxation}.

\subsection{Shor's Relaxation}
\label{subsec:shor}

\emph{Shor's relaxation}~\cite{shor1987quadratic} is a foundational technique for globally solving QCQPs through SDP.
To apply Shor's relaxation to problems of the form~\eqref{eq:generic-ls}, the constrained set $\mathcal{X}$ (such as $\mathrm{SO}(3)$ or $\mathrm{SE}(3)$) is first reformulated using polynomial constraints over $\bbR^n$, transforming the problem into a QCQP.
The key idea is to lift the model parameter $\bx \!\in\! \bbR^n$ to a matrix $\bX \!=\! \bx\bx^\top$ and relax the nonconvex rank-one constraint, yielding a convex SDP that can be solved to global optimality.
When the relaxation is tight (rank-one solution), Shor's relaxation provides certifiable global solutions; otherwise, rounding and local refinement yield high-quality approximate solutions.

\PAR{Formulation: from QCQP to SDP.}
A broad family of formulations in 3D vision can be cast as a QCQP:
\begin{equation}
\label{eq:qcqp}
    \begin{aligned}
        \min_{\bx \in \bbR^n} \quad & \bx^\top \bC \bx \\
        \text{s.t.}           \quad & \bx^\top \bA_i \bx = b_i, \quad i = 1,\dots,m,
    \end{aligned}
    \tag{QCQP}
\end{equation}
where $\bC, \bA_1,\dots,\bA_m \!\in\! \bbS^n$ are real symmetric matrices.
\eqref{eq:qcqp} is nonconvex in general due to the quadratic objective and constraints.
Introducing the lifted matrix $\bX \!=\! \bx\bx^\top$, which is a rank-one symmetric positive semidefinite (PSD) matrix, yields the rank-constrained matrix problem:
\begin{equation*}
    \begin{aligned}
        \min_{\bX \in \bbS^n} \quad & \trace(\bC \bX) \\
        \text{s.t.}           \quad & \trace(\bA_i \bX) = b_i, \quad i = 1,\dots,m, \\
                                    & \bX \succeq \mathbf{0}, \quad \rank(\bX) = 1.
    \end{aligned}
\end{equation*}
Dropping the nonconvex rank constraint yields Shor's SDP relaxation:
\begin{equation}
\label{eq:sdp}
    \begin{aligned}
        \min_{\bX \in \bbS^n} \quad & \trace(\bC \bX) \\
        \text{s.t.}           \quad & \trace(\bA_i \bX) = b_i, \quad i = 1,\dots,m, \\
                                    & \bX \succeq \mathbf{0}, 
    \end{aligned}
    \tag{SDP}
\end{equation}
which is convex and solvable via standard interior-point methods (IPMs) implemented in solvers such as MOSEK~\cite{aps2019mosek}, SDPT3~\cite{tutuncu2003solving}, and SEDUMI~\cite{sturm1999using}.

The convex SDP~\eqref{eq:sdp} admits a dual formulation that plays a central role in both theory and practice.
The dual problem is given by:
\begin{equation}
\label{eq:sdp_dual}
    \begin{aligned}
        \max_{\vlambda \in \bbR^m} \quad & \bbb^\top \vlambda \\
        \text{s.t.}                \quad & \bQ(\vlambda) = \bC - \sum_{i=1}^m \lambda_i \bA_i \succeq \mathbf{0},
    \end{aligned}
    \tag{Dual}
\end{equation}
where $\bbb \!\in\! \bbR^m$ and $\vlambda \!\in\! \bbR^m$ are the dual variables. 
The matrix $\bQ(\vlambda)$ is the Hessian of the Lagrangian of~\eqref{eq:qcqp}, and~\eqref{eq:sdp_dual} coincides with the Lagrangian dual of~\eqref{eq:qcqp}.
By SDP duality, the primal and dual optimal values coincide when Slater's condition~\cite{boyd2004convex} holds, enabling certification of global optimality through zero duality gap.

\cref{fig:shor-relaxation} illustrates the relationships between the original QCQP, the primal and dual SDP formulations, and the recovery of global solutions.

\input{fig/shor}

\PAR{Modeling and solution ingredients.}
Applying Shor's relaxation to geometric problems involves:
\begin{itemize}
    \item \textbf{Lifting and structure preservation.}
    The lifting $\bX \!=\! \bx\bx^\top$ is designed to retain problem-specific  structure (\eg block structure from averaging~\cite{eriksson2018rotation} and pose graph topology~\cite{tian2019block}), enabling tighter relaxations and efficient computation.
    
    \item \textbf{Tightness and certification.}
    The relaxation is \emph{tight} if the SDP solution $\bX^\star$ satisfies $\rank(\bX^\star) \!=\! 1$, allowing exact recovery of $\bx^\star$ with a global optimality certificate via zero duality gap.
    Tightness can be guaranteed under bounded noise, non-degeneracy, or favorable spectral properties~\cite{dumbgen2024toward}.
    
    \item \textbf{Recovery and rounding.}
    When $\bX^\star$ is higher-rank, low-rank factorization ($\bX^\star \!\approx\! \bY\bY^\top$) followed by manifold-constrained local refinement yields approximate solutions~\cite{tirado2024correspondences}.
    Dual certificates provide a posteriori suboptimality bounds~\cite{garcia2021certifiable,garcia2023certifiable,garcia2023fast,garcia2024fast}.
\end{itemize}

\PAR{Theoretical properties.}
Shor's relaxation enjoys strong duality under mild assumptions (\eg Slater's condition~\cite{boyd2004convex}).
When tight, that is, when the SDP solution $\bX^\star$ satisfies $\rank(\bX^\star) \!=\! 1$, the method returns globally optimal solutions with certificates via zero duality gap.
Tightness guarantees exist for certain 3D vision problems.
For instance, rotation averaging~\cite{eriksson2018rotation}, relative pose estimation~\cite{zhao2020efficient}, and pose graph optimization~\cite{rosen2019se} admit exact recovery when graph connectivity, spectral properties, or measurement quality are favorable.
In noiseless settings (\eg Wahba problem), Shor's relaxation provably recovers the true solution~\cite{yang2019quaternion}.
When the relaxation is not tight, dual certificates can still bound suboptimality, and rounding followed by local refinement often yields near-optimal solutions.

\PAR{Computational complexity.}
Standard IPMs for solving SDPs of dimension $n$ and $m$ constraints require $O(n^3 \!+\! m^2 n^2 \!+\! m^3)$\footnote{$O(n^3)$ for spectral decomposition of dense primal and dual iterates, $O(m^2 n^2)$ for forming the Schur complement system, and $O(m^3)$ for factorizing and solving the Schur complement system.} per iteration and $O(\sqrt{n \!+\! m})$ iterations, with memory $O(n^2 \!+\! m)$.
For typical 3D vision problems where $m \!=\! O(n)$, this yields $O(n^{3.5})$ overall complexity, though larger constraint sets may incur additional costs.
Structure-exploiting techniques (\eg chordal decomposition~\cite{dumbgen2024exploiting}, low-rank factorization~\cite{boumal2016non,burer2003nonlinear}) can improve scalability, though typically at the cost of relaxing strict optimality guarantees.

\PAR{Robust estimation via M-estimation.}
Shor's relaxation extends naturally to robust estimation by embedding M-estimation~\cite{huber19811981,hampel1986} within the SDP formulation.
Representative approaches include~\cite{harenstam2023semidefinite}, that integrates TLS losses, and 
GlobustVP~\cite{liao2025convex}, which employs truncated multi-selection errors.
These formulations maintain convexity while achieving robustness to moderate outlier ratios, providing certificates when tight.

\PAR{Hybrid frameworks.}
To improve scalability, many solvers combine Shor's relaxation with local refinement.

\begin{itemize}
    \item \textbf{Shor's relaxation initialization.}
    A Shor-based global step provides initialization for local refinement~\cite{chen2021hybrid}.
    
    \item \textbf{Shor's relaxation certification.}
    Fast local solvers or sampling-based methods (\eg RANSAC~\cite{fischler1981random}) generate candidate solutions that are certified or corrected via Shor's relaxation~\cite{rosen2019se,yang2020teaser,dellaert2020shonan}.
\end{itemize}

\PAR{Strengths and limitations.}

\begin{itemize}
    \item \textbf{Strengths.}
    Shor's relaxation offers three key advantages:
    \begin{enumerate}
        \item \emph{Convex formulation}: The SDP relaxation is convex and solvable to global optimality via standard solvers.
        \item \emph{Certifiable solutions}: When tight, the method provides certificates of global optimality via zero duality gap.
        \item \emph{Principled recovery}: Dual certificates enable a posteriori verification and suboptimality bounds even when the relaxation is not tight.
     \end{enumerate}

    \item \textbf{Limitations.}
    The primary limitation is \emph{cubic computational complexity}, with $O(n^{3.5})$ scaling and $O(n^2)$ memory limiting applicability to moderate-scale problems.
    When the relaxation is not tight, rounding and local refinement are necessary, potentially sacrificing global optimality guarantees.
    Structure-exploiting algorithms can improve scalability but may compromise tightness or exact certification.
\end{itemize}

\subsection{Moment-SOS Relaxation}
\label{subsec:moment-sos}

\emph{Moment-SOS relaxation}~\cite{lasserre2001global,lasserre2009moments,parrilo2003semidefinite}, also known as the Lasserre hierarchy, is a powerful framework for globally solving polynomial optimization problems (POPs) where objectives and constraints are multivariate polynomials.
The key idea is to replace polynomial nonnegativity conditions with SOS certificates, polynomial decompositions of the form $p(\bx) \!=\! \sum_i \sigma_i^2(\bx)$, yielding a hierarchy of increasingly tight SDP relaxations.
Under mild assumptions, this hierarchy converges monotonically to the global optimum, and low-order relaxations (typically degree $2$--$4$) often suffice for geometric vision problems.

\PAR{Formulation: from POP to SDP.}
Many 3D vision problems can be naturally formulated as a POP:
\begin{equation}
\label{eq:pop}
    \begin{aligned}
        \min_{\bx \in \bbR^n} \quad & f(\bx) \\
        \text{s.t.}           \quad & g_i(\bx) \geq 0, \quad i = 1,\dots,m, \\
                                    & h_j(\bx) = 0, \quad j = 1,\dots,p,
    \end{aligned}
    \tag{POP}
\end{equation}
where $f, g_i, h_j \!\in\! \bbR[\bx]$ are multivariate polynomials.
\eqref{eq:pop} arises naturally from polynomial constraints, \eg orthogonality ($\bR^\top\bR \!=\! \bI$), unit determinant ($\det(\bR) \!=\! 1$), and epipolar relations.
Despite its algebraic simplicity,~\eqref{eq:pop} is NP-hard in general.

Moment-SOS relaxation transforms~\eqref{eq:pop} into a convex SDP by reformulating polynomial nonnegativity as SOS conditions, representing SOS polynomials via positive semidefinite Gram matrices, and constructing a dual moment formulation.

The optimization in~\eqref{eq:pop} seeks the largest $\gamma$ such that $f(\bx) \!-\! \gamma \!\geq\! 0$ on the feasible set.
As deciding global nonnegativity of a polynomial is itself NP-hard, a tractable surrogate is to require that $f(\bx) \!-\! \gamma$ admits a \emph{SOS} decomposition:
\begin{equation*}
\label{eq:pop-sos}
    f(\bx)-\gamma = \sigma_0(\bx)+\sum_{i=1}^m \sigma_i(\bx)\, g_i(\bx) + \sum_{j=1}^p \lambda_j(\bx)h_j(\bx),
\end{equation*}
where each $\sigma_i(\bx)$ is an SOS polynomial and each $\lambda_j(\bx)$ is an arbitrary polynomial.
Each SOS polynomial can be represented via a Gram matrix $\sigma_i(\bx) \!=\! \bz^\top(\bx)\bQ_i\bz(\bx)$ with $\bQ_i \!\succeq\! 0$, where $\bz(\bx)$ is a vector of monomials.
Maximizing $\gamma$ subject to $\bQ_i \!\succeq\! 0$ and linear constraints (from coefficient matching) yields a convex SDP solvable by standard solvers~\cite{tutuncu2003solving,aps2019mosek}.

An equivalent dual perspective is the \emph{moment formulation}.
Let $[\bx]_d \!=\! \{ \bx^{\valpha} \mid \|\valpha\|_1 \!\leq\! d, \valpha \!\in\! \nnint^n \}$ denote the vector of all monomials in $\bx$ up to degree $d$ (the relaxation order).
Given $[\bx]_d$, let $\bM_d \!=\! [\bx]_d [\bx]_d^\top$ be \emph{moment matrix}.
The $d$-th order relaxation relaxes the nonconvex rank-one constraint, requiring instead $\bM_d \!\succeq\! 0$ along with \emph{localizing matrices} $g_i \cdot \bM_{d-\lceil \deg(g_i)/2 \rceil} \!\succeq\! 0$ for each constraint $g_i \!\geq\! 0$.

The final resulting \emph{Lasserre's (Moment-SOS) hierarchy} produces a sequence of SDPs (with increasing $d$) whose optimal values monotonically converge to the global optimum under mild assumptions (Archimedean condition)~\cite{lasserre2001global,lasserre2009moments,nie2014optimality,blekherman2012semidefinite}.
When the relaxation is tight, indicated by a rank-one condition on the moment matrix $\bM_d$, a globally optimal solution can be extracted from low-rank factorization or moment-based extraction.\footnote{%
More generally, tightness can be characterized by the \emph{flatness extension property}, which does not require rank-one moment matrices~\cite{henrion2005detecting,nie2023moment}, \cite[Sec. 5.6]{Yang2024book-sdp}.
\label{fn:flatness}}

\PAR{Modeling and solution ingredients.}
Applying Moment-SOS relaxation to geometric problems involves:

\begin{itemize}
    \item \textbf{Polynomial modeling of constraints.}
    Nonlinear geometric relations such as orthogonality ($\bR^\top \bR \!=\! \bI$), unit determinant ($\det(\bR) \!=\! 1$), and epipolar constraints are naturally expressed as polynomial equations and inequalities.
    
    \item \textbf{Relaxation order and hierarchy.}
    The relaxation order $d$ controls the size and tightness of the SDP: higher $d$ provides tighter bounds but increases computational cost combinatorially.
    Low orders ($d \!=\! 2$--$4$) often suffice for vision problems.
    
    \item \textbf{Tightness and extraction.}
    When the moment matrix $\bM_d$ satisfies a rank condition, the relaxation is tight and a globally optimal solution can be extracted via low-rank factorization or moment-based extraction.\footref{fn:flatness}
\end{itemize}

\cref{fig:moment-sos-hierarchy} illustrates the hierarchy's convergence properties and the fundamental trade-off: higher relaxation orders $d$ yield monotonically tighter lower bounds $\gamma_d$ that converge to the global optimum $f^*$ at a finite order $d^*$, but require solving increasingly large SDPs with moment matrix size $\binom{n+d}{d}$.

\input{fig/moment_sos}

\PAR{Theoretical properties.}
Moment-SOS hierarchy enjoys strong convergence guarantees: under mild assumptions (Archimedean condition), the sequence of SDP relaxations converges monotonically to the global optimum as the relaxation order $d$ increases~\cite{lasserre2001global,nie2014optimality}.
When the relaxation is tight at order $d$, indicated by rank conditions on the moment matrix $\bM_d$, it yields the exact global solution with a certificate of optimality.
Tightness at low orders ($d \!=\! 2$--$4$) has been observed for several geometric vision problems, including absolute pose estimation~\cite{schweighofer2008globally}, fundamental matrix estimation~\cite{bugarin2015rank}, and pose graph optimization~\cite{mangelson2019guaranteed}.
When not tight, dual bounds provide suboptimality certificates, and low-rank rounding yields approximate solutions.

\PAR{Computational complexity.}
The main challenge lies in the combinatorial growth of SDP size with relaxation order $d$.
For $n$ variables and order $d$, the moment matrix dimension $r \!=\! \binom{n+d}{d}$ and the number of linear equality constraints $p \!=\! \binom{n+2d}{2d}$ grow rapidly (\eg $r \!=\! 1001$ and $p \!=\! 43,758$ for $n \!=\! 10$, $d \!=\! 4$), yielding complexity $O(r^{3.5})$ via IPMs with $O(r^2)$ memory.
This combinatorial explosion limits high-order relaxations, though low-order relaxations ($d \!=\! 2$--$4$) often suffice for geometric vision problems.
Similar to Shor’s relaxation, structure-exploiting techniques (\eg sparse-SOS~\cite{mangelson2019guaranteed}, chordal decomposition~\cite{dumbgen2024exploiting}, low-rank factorization~\cite{boumal2016non,burer2003nonlinear}) can improve scalability but remain limited to moderate problem sizes.

\PAR{Robust estimation.}
Moment-SOS relaxations extend naturally to robust estimation through two complementary approaches.
First, M-estimation~\cite{huber19811981,hampel1986} can be embedded within the polynomial formulation to mitigate outlier influence while maintaining SDP convexity~\cite{yang2020one,yang2022certifiably}.
Second, consensus maximization itself can be reformulated as a polynomial program~\cite{yang2022certifiably}, enabling globally optimal inlier set identification through Moment-SOS hierarchy.

\PAR{Hybrid frameworks.}
To improve practical efficiency, Moment-SOS relaxations are often combined with local or sampling-based methods (\eg iterative reweighting~\cite{weiszfeld1937point}, RANSAC~\cite{fischler1981random}).
These solvers generate candidate solutions, which are then certified or refined using low-order Moment-SOS relaxations~\cite{yang2020one,yang2022certifiably}.

\PAR{Strengths and limitations.}

\begin{itemize}
    \item \textbf{Strengths.}
    Moment-SOS relaxation offers three key advantages:
    \begin{enumerate}
        \item \emph{Unified polynomial framework}: Diverse geometric constraints (\eg orthogonality, determinant, epipolar relations) are naturally expressed as polynomial equations.
        \item \emph{Convergent hierarchy}: The Lasserre hierarchy converges monotonically to the global optimum with provable guarantees.
        \item \emph{Low-order sufficiency}: Many vision problems admit tight relaxations at low orders ($d \!=\! 2$--$4$), avoiding the need for sampling or initialization.
    \end{enumerate}

    \item \textbf{Limitations.}
    The primary limitation is \emph{combinatorial SDP growth}: the moment matrix size $\binom{n+d}{d}$ and the number of linear equality constraints $\binom{n+2d}{2d}$ increase rapidly with problem dimension $n$ and relaxation order $d$, limiting scalability to moderate-size problems.
    Tightness at low order is problem-dependent and not guaranteed a priori, requiring trial-and-error or higher-order (more expensive) relaxations.
    Numerical conditioning can degrade with high-degree monomials, particularly for ill-scaled or noisy data.
\end{itemize}

\subsection{Other Relaxation Techniques}
\label{subsec:other_relaxation}

Beyond Shor's relaxation and Moment-SOS methods, a family of specialized convex relaxations exploits problem-specific structure in 3D vision, trading modeling generality for computational efficiency and scalability.
These techniques leverage quasi-convexity, second-order cone programming (SOCP), linear programming (LP), quadratic programming (QP), or convex-hull approximations to obtain globally or near-globally optimal solutions for targeted problem classes.

\PAR{Formulation principles.}
Rather than constructing general-purpose relaxations for arbitrary polynomial or quadratic programs, these methods embed problem-specific structure into tailored convex formulations.
Key strategies include:
(i) quasi-convex reformulations for $L_\infty$-type objectives,
(ii) SOCP, LP, or QP relaxations for norm-based constraints, and 
(iii) convex hull approximations of nonconvex feasible sets.
While sacrificing generality, these specialized relaxations achieve substantially improved scalability and are often the method of choice for large-scale or time-critical applications.

\PAR{Modeling and solution ingredients.}
Representative modeling patterns include:
\begin{itemize}
  \item \textbf{Quasi-convex reformulations.}
  Problems such as triangulation and absolute pose estimation admit $L_\infty$-norm formulations that are quasi-convex, enabling global solutions via bisection over SOCP or LP subproblems~\cite{hartley2004sub,kahl2005multiple,sim2006removing,sim2006recovering,olsson2007efficient,seo2007fast,ke2007quasiconvex,kahl2008multiple,agarwal2008fast,olsson2008polynomial,li2009efficient,olsson2010outlier}.

  \item \textbf{SOCP/QP/LP relaxations.}
  Norm-based objectives and constraints can be approximated by SOCP, QP, or LP models, yielding convex programs that are efficiently solvable and provably exact under favorable conditions or statistical regimes~\cite{ozyesil2015robust,goldstein2016shapefit,papalia2022score}.
  
  \item \textbf{Convex hull relaxations.}
  Nonconvex feasible regions are approximated by their convex hulls, enabling linear or quadratic formulations with controlled approximation quality~\cite{saunderson2015semidefinite,rosen2015convex,reich2017global,zhou20153d}.
\end{itemize}

\PAR{Theoretical properties.}
Optimality guarantees for these methods depend strongly on problem structure and formulation choice.
Quasi-convex reformulations are exact for $L_\infty$-norm objectives, providing global optimality certificates via bisection convergence.
SOCP, QP, and LP relaxations yield tractable convex programs with provable recovery guarantees under favorable geometric or statistical conditions (\eg bounded noise, sufficient measurements).
Convex hull relaxations provide outer approximations whose tightness depends on the geometry of the feasible set.

\PAR{Computational complexity.}
Compared to Shor's relaxation and Moment-SOS hierarchies, these specialized methods are substantially more scalable, often by orders of magnitude.
Quasi-convex formulations reduce global optimization to a sequence of convex subproblems via bisection, typically converging in tens of iterations.
SOCP, QP, and LP solvers handle significantly larger problem instances with lower memory requirements than generic SDP solvers.
Convex-hull approaches often achieve the best scalability, making them suitable for large-scale or real-time applications.

\PAR{Strengths and limitations.}

\begin{itemize}
    \item \textbf{Strengths.}
    These specialized approaches extend certifiable optimization to settings where Shor's or Moment-SOS methods become impractical due to problem scale.
    By exploiting problem-specific structure, they achieve orders-of-magnitude speedups while often retaining provable optimality guarantees or tight approximations.

    \item \textbf{Limitations.}
    The primary limitation is \emph{problem specificity}: each technique targets a narrow class of problems and may fail when structural assumptions are violated.
    Approximation quality can degrade significantly for problems outside the intended scope, and careful problem-specific modeling is necessary to achieve practical performance.
\end{itemize}

%% file: fig/shor.tex
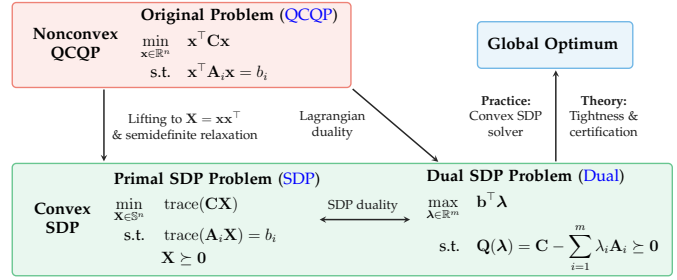
\begin{figure}[t]
\centering

\definecolor{nonconvexColor}{HTML}{E74C3C}
\definecolor{convexColor}{HTML}{27AE60}
\definecolor{resultColor}{HTML}{3498DB}

\resizebox{\columnwidth}{!}{
\begin{tikzpicture}[
    box/.style={
        draw=black!70,
        rectangle,
        rounded corners=3pt,
        minimum width=4.5cm,
        minimum height=1.5cm,
        align=center,
        font=\large,
        line width=0.9pt,
        fill=white
    },
    nonconvexbox/.style={
        box,
        fill=nonconvexColor!10,
        draw=nonconvexColor!70,
        line width=1pt
    },
    convexbox/.style={
        box,
        minimum width=17cm,
        minimum height=2.5cm,
        fill=convexColor!10,
        draw=convexColor!70,
        line width=1pt
    },
    resultbox/.style={
        box,
        minimum width=4cm,
        minimum height=1.2cm,
        fill=resultColor!10,
        draw=resultColor!70,
        line width=1pt
    },
    arrow/.style={
        -stealth,
        line width=1.1pt,
        black!85,
        shorten >=1pt,
        shorten <=1pt
    },
    doublearrow/.style={
        stealth-stealth,
        line width=1.1pt,
        black!85,
        shorten >=1pt,
        shorten <=1pt
    },
    outsidelabel/.style={
        font=\small,
        align=center,
        fill=white,
        inner sep=3pt,
        outer sep=1pt,
        text=black!90
    },
    insidelabel/.style={
        font=\small,
        align=center,
        fill=none,
        inner sep=3pt,
        outer sep=1pt,
        text=black!90
    }
]

\node[nonconvexbox, text width=8.5cm, anchor=west] (original) at (-4, 4.5) {
    \begin{minipage}{8.5cm}
    \centering
    \begin{tabular}{c@{\hspace{0.4em}}c}
        \begin{tabular}{c}
            \textbf{Nonconvex} \\
            \textbf{QCQP}
        \end{tabular}
        &
        \begin{tabular}{l}
            \textbf{Original Problem}~\eqref{eq:qcqp} \\[0.35em]
            $\begin{aligned}
                \min_{\bx \in \bbR^n} \quad & \bx^\top \bC \bx \\
                \text{s.t.} \quad & \bx^\top \bA_i \bx = b_i
            \end{aligned}$
        \end{tabular}
    \end{tabular}
    \end{minipage}
};

\node[resultbox] (global) at (10, 4.5) {
    \textbf{Global Optimum}
};

\node[convexbox, text width=16cm, anchor=west] (sdp) at (-4, 0) {
    \begin{minipage}{16cm}
    \hspace{-0.5cm}
    \begin{tabular}{c@{\hspace{0.4em}}l}
        \begin{tabular}{c}
            \textbf{Convex} \\
            \textbf{SDP}
        \end{tabular}
        &
        \begin{tabular}{l}
            \textbf{Primal SDP Problem}~\eqref{eq:sdp} \\[0.35em]
            $\begin{aligned}
                \min_{\bX \in \bbS^n} \quad & \trace(\bC \bX) \\
                \text{s.t.} \quad & \trace(\bA_i \bX) = b_i \\
                                  & \bX \succeq \mathbf{0}
            \end{aligned}$
        \end{tabular}
    \end{tabular}
    \hfill
    \begin{tabular}{l}
        \textbf{Dual SDP Problem}~\eqref{eq:sdp_dual} \\[0.35em]
        $\begin{aligned}
            \max_{\vlambda \in \bbR^m} \quad & \bbb^\top \vlambda \\
            \text{s.t.} \quad & \bQ(\vlambda) = \bC - \sum_{i=1}^m \lambda_i \bA_i \succeq \mathbf{0}
        \end{aligned}$
    \end{tabular}
    \hspace{-0.5cm}
    \end{minipage}
};

\coordinate (primal_right) at (3.9, 0);
\coordinate (dual_left) at (6.3, 0);

\coordinate (lifting_x) at ($(original.south) + (-2, 0)$);
\draw[arrow] (lifting_x) -- node[outsidelabel, right, xshift=2pt, font=\normalsize] {
    Lifting to $\bX = \bx\bx^\top$ \\
    \& semidefinite relaxation
} (lifting_x |- sdp.north);

\draw[arrow] (original.south east) -- node[outsidelabel, left, xshift=-18pt, font=\normalsize] {
    Lagrangian \\
    duality
} ($(sdp.north) + (2.5, 0)$);

\draw[doublearrow] (primal_right) -- node[insidelabel, above, yshift=1pt, font=\normalsize] {
    SDP duality
} (dual_left);

\coordinate (practice_x) at ($(sdp.north) + (5.5, 0)$);
\draw[arrow] (practice_x) -- node[outsidelabel, left, xshift=-6pt, font=\normalsize] {
    \textbf{Practice:} \\
    Convex SDP \\
    solver
} node[outsidelabel, right, xshift=6pt, font=\normalsize] {
    \textbf{Theory:} \\
    Tightness \& \\
    certification
} (practice_x |- global.south);

\end{tikzpicture}
}
\caption{%
    Shor's relaxation for solving nonconvex QCQPs.
    The original nonconvex problem~\eqref{eq:qcqp} is transformed into a convex SDP via lifting to matrix variable $\bX = \bx\bx^\top$ and semidefinite relaxation.
    The primal SDP~\eqref{eq:sdp} and its dual~\eqref{eq:sdp_dual} are related through SDP duality.
    In practice, the global optimum is computed using convex SDP solvers based on IPMs~\cite{aps2019mosek,tutuncu2003solving,sturm1999using}.
    The theoretical properties including tightness of the relaxation and certification of global optimality are discussed in~\cref{subsec:shor}.
}
\label{fig:shor-relaxation}
\end{figure}

%% file: fig/moment_sos.tex
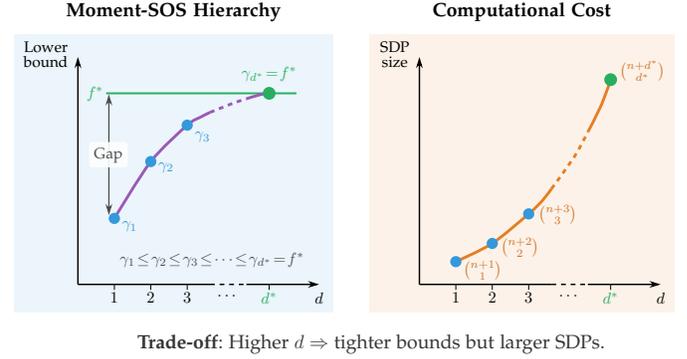
\begin{figure}[t]
\centering

\definecolor{domainColor}{HTML}{3498DB}
\definecolor{relaxColor}{HTML}{E67E22}
\definecolor{optColor}{HTML}{27AE60}
\definecolor{boundColor}{HTML}{9B59B6}

\resizebox{\columnwidth}{!}{
\begin{tikzpicture}[
    >={Stealth[length=2.5mm, width=1.5mm]},
    font=\small,
    every node/.style={font=\small}
]

\begin{scope}[shift={(0,0)}]
    
    \fill[domainColor!10] (-1.4, -0.6) rectangle (5.6, 5.5);
    
    \node[font=\large\bfseries] at (2.1, 6) {Moment-SOS Hierarchy};
    
    \draw[line width=0.9pt] (0, 0) -- (2.9, 0);
    \draw[line width=0.9pt, dashed] (2.9, 0) -- (3.7, 0);
    \draw[->, line width=0.9pt] (3.7, 0) -- (5.3, 0);
    
    \draw[->, line width=0.9pt] (0, 0) -- (0, 5);
    
    \node[below, font=\normalsize] at (5.3, -0.05) {$d$};
    \node[above left, font=\normalsize, align=center] at (-0.1, 4.65) {Lower\\[-2pt]bound};
    
    \def\fstar{4.2}
    \draw[optColor, line width=1.2pt] (0.62, \fstar) -- (4.8, \fstar);
    \node[left, font=\normalsize, optColor] at (0.68, \fstar) {$f^*$};
    
    \draw[boundColor, line width=1.8pt, smooth] plot coordinates {
        (0.8, 1.45) (1.6, 2.7) (2.4, 3.5) (2.7, 3.68) (2.9, 3.78)
    };

    \draw[boundColor, line width=1.8pt, dashed, smooth] plot coordinates {
        (2.9, 3.78) (3.4, 3.98) (3.7, 4.08)
    };
    
    \draw[boundColor, line width=1.8pt, smooth] plot coordinates {
        (3.7, 4.08) (4.0, 4.17) (4.2, 4.2)
    };
    
    \fill[domainColor] (0.8, 1.45) circle (3.5pt);
    \fill[domainColor] (1.6, 2.7) circle (3.5pt);
    \fill[domainColor] (2.4, 3.5) circle (3.5pt);
    \fill[optColor] (4.2, 4.2) circle (4pt);
    
    \node[below, font=\normalsize] at (0.8, -0.05) {$1$};
    \node[below, font=\normalsize] at (1.6, -0.05) {$2$};
    \node[below, font=\normalsize] at (2.4, -0.05) {$3$};
    \node[below, font=\normalsize] at (3.3, -0.05) {$\cdots$};
    \node[below, font=\normalsize, optColor] at (4.2, -0.05) {$d^*$};
    
    \draw[line width=0.5pt] (0.8, 0) -- (0.8, -0.12);
    \draw[line width=0.5pt] (1.6, 0) -- (1.6, -0.12);
    \draw[line width=0.5pt] (2.4, 0) -- (2.4, -0.12);
    \draw[line width=0.5pt] (4.2, 0) -- (4.2, -0.12);
    
    \node[below right, font=\normalsize, domainColor] at (0.85, 1.5) {$\gamma_1$};
    \node[below right, font=\normalsize, domainColor] at (1.65, 2.8) {$\gamma_2$};
    \node[below right, font=\normalsize, domainColor] at (2.45, 3.47) {$\gamma_3$};
    \node[above, font=\normalsize, optColor] at (4.2, 4.35) {$\gamma_{d^*}\!=\!f^*$};
    
    \draw[<->, line width=0.8pt, black!70] (0.68, \fstar-0.02) -- (0.68, 1.52);
    \node[left, font=\normalsize, black!70, fill=white, fill opacity=0.9, text opacity=1, rounded corners=2pt] at (1.1, 2.85) {Gap};
    
    \node[font=\normalsize, align=center, black!70, fill=domainColor!10, inner sep=2pt, rounded corners=2pt] 
        at (2.95, 0.55) {$\gamma_1 \!\leq\! \gamma_2 \!\leq\! \gamma_3 \!\leq\! \cdots \!\leq\! \gamma_{d^*} \!=\! f^*$};
    
\end{scope}

\begin{scope}[shift={(7.5,0)}]
    
    \fill[relaxColor!10] (-1.1, -0.6) rectangle (5.6, 5.5);
    
    \node[font=\large\bfseries] at (2.25, 6) {Computational Cost};
    
    \draw[line width=0.9pt] (0, 0) -- (2.9, 0);
    \draw[line width=0.9pt, dashed] (2.9, 0) -- (3.7, 0);
    \draw[->, line width=0.9pt] (3.7, 0) -- (5.3, 0);
    
    \draw[->, line width=0.9pt] (0, 0) -- (0, 5);
    
    \node[below, font=\normalsize] at (5.3, -0.05) {$d$};
    \node[above left, font=\normalsize, align=center] at (-0.1, 4.65) {SDP\\[-2pt]size};
    
    \draw[relaxColor, line width=1.8pt, smooth] plot coordinates {
        (0.8, 0.5) (1.6, 0.9) (2.4, 1.55) (2.7, 1.84) (2.9, 2.1)
    };
    
    \draw[relaxColor, line width=1.8pt, dashed, smooth] plot coordinates {
        (2.9, 2.1) (3.4, 2.8) (3.7, 3.35)
    };
    
    \draw[relaxColor, line width=1.8pt, smooth] plot coordinates {
        (3.7, 3.35) (4.0, 3.95) (4.2, 4.5)
    };
    
    \fill[domainColor] (0.8, 0.5) circle (3.5pt);
    \fill[domainColor] (1.6, 0.9) circle (3.5pt);
    \fill[domainColor] (2.4, 1.55) circle (3.5pt);
    \fill[optColor] (4.2, 4.5) circle (4pt);
    
    \node[below, font=\normalsize] at (0.8, -0.05) {$1$};
    \node[below, font=\normalsize] at (1.6, -0.05) {$2$};
    \node[below, font=\normalsize] at (2.4, -0.05) {$3$};
    \node[below, font=\normalsize] at (3.3, -0.05) {$\cdots$};
    \node[below, font=\normalsize, optColor] at (4.2, -0.05) {$d^*$};
    
    \draw[line width=0.5pt] (0.8, 0) -- (0.8, -0.12);
    \draw[line width=0.5pt] (1.6, 0) -- (1.6, -0.12);
    \draw[line width=0.5pt] (2.4, 0) -- (2.4, -0.12);
    \draw[line width=0.5pt] (4.2, 0) -- (4.2, -0.12);
    
    \node[below right, font=\normalsize, relaxColor!90!black] at (0.85, 0.65) {$\tbinom{n+1}{1}$};
    \node[below right, font=\normalsize, relaxColor!90!black] at (1.67, 1.15) {$\tbinom{n+2}{2}$};
    \node[above right, font=\normalsize, relaxColor!90!black] at (2.5, 1.2) {$\tbinom{n+3}{3}$};
    \node[right, font=\normalsize, relaxColor!90!black] at (4.28, 4.7) {$\tbinom{n+d^*}{d^*}$};
    
\end{scope}

\node[font=\large, align=center, black!80] at (6.5, -1.3) {
    \textbf{Trade-off}: Higher $d$ $\Rightarrow$ tighter bounds but larger SDPs. 
};

\end{tikzpicture}
}
\caption{%
    Moment-SOS (Lasserre) hierarchy for polynomial optimization.
    \textbf{Left}: Lower bounds $\gamma_d$ from successive SDP relaxations converge monotonically to the \textcolor{optColor}{global optimum} $f^*$, with tightness achieved at finite order $d^*$.
    \textbf{Right}: The moment matrix dimension $\tbinom{n+d}{d}$ grows combinatorially with relaxation order $d$, illustrating the computational cost of tightening the bounds.
    For many 3D vision problems, low-order relaxations ($d \!=\! 2$--$4$) suffice for tightness.
}
\label{fig:moment-sos-hierarchy}
\end{figure}

%% file: sec/5_gnc.tex
\section{Global Solver 3: Graduated Non-Convexity (GNC)}
\label{sec:gnc}

\emph{Graduated Non-Convexity (GNC)}~\cite{black1996unification,rangarajan1990generalized,mobahi2015link,rose2002deterministic} is a deterministic continuation framework for global optimization that addresses nonconvex problems through \emph{homotopy deformation}.
Rather than exhaustively searching (as in BnB) or lifting to higher dimensions (as in convex relaxation), GNC embeds the original hard nonconvex objective into a parametrized family of progressively nonconvex surrogates controlled by a continuation parameter $\mu$.
Starting from a convex (or nearly convex) surrogate at large $\mu$, GNC gradually decreases $\mu$ and tracks a solution path through a sequence of easier subproblems, whose minimizers smoothly evolve toward a solution of the original objective.
Under appropriate conditions on the homotopy schedule and problem structure, this continuation strategy often converges to near-global minima \emph{empirically}, though without \emph{a priori} global optimality guarantees.
In addition, by naturally suppressing outliers during the continuation process, GNC has become a practical and widely-used tool for robust estimation in geometric vision.

\PAR{Modeling and solution ingredients.}
Applying GNC to geometric problems requires three modeling components:

\begin{itemize}
    \item \textbf{Robust loss function and parametrization.}
    The choice of robust penalty $\rho(r,\mu)$ governs the convexity-to-nonconvexity transition and the method's resilience to outliers.
    Common choices include GM~\cite{geman1985}, Welsch~\cite{dennis1978techniques,leclerc1989constructing}, and TLS~\cite{yang2020graduated}.
    Each admits a continuation parameter $\mu$ that controls the transition from a convex surrogate to the desired nonconvex robust loss, though the specific parametrization and schedule direction vary across loss functions.
    
    For instance, the GM surrogate is:
    \begin{equation*}
        \rho(r,\mu) = \frac{\mu r^2}{\mu\sigma^2 + r^2}, \quad \mu > 0,
    \end{equation*}
    where $\sigma$ controls robustness.
    As $\mu \!\to\!\infty$, $\rho(r,\mu) \!\to\! r^2/\sigma^2$ (convex quadratic), and as $\mu \!\to\! 1$, it recovers the nonconvex GM loss $\rho(r) \!=\! r^2/(\sigma^2+r^2)$.
    
    \item \textbf{Continuation schedule and iterative reweighting.}
    At each fixed $\mu_k$, the problem reduces to an iteratively reweighted least-squares (IRLS) form:
    \begin{equation}
    \label{eq:gnc-irls}
        \min_{\bx} \sum_{i=1}^N w_i(\mu_k)\, r_i^2(\bx), \quad
        w_i(\mu_k)=\frac{\psi(r_i,\mu_k)}{2r_i},
        \tag{IRLS}
    \end{equation}
    where $\psi(r,\mu) \!=\! \partial\rho/\partial r$ is the influence function.
    The weights $w_i(\mu_k)$ progressively down-weight large residuals as $\mu$ decreases, naturally implementing M-estimation~\cite{huber19811981} without explicit inlier and outlier classification.
    A typical cooling schedule is geometric: $\mu_{k+1} \!=\! \alpha\mu_k$ with $\alpha \!\in\! (0,1)$.
    Beyond fixed schedules, adaptive strategies can dynamically adjust $\mu$ based on local convexity~\cite{sidhartha2023adaptive,chitturi2024adaptive} or learned through reinforcement learning~\cite{mai2026neural} for improved reliability and efficiency.
    
    \item \textbf{Initialization and termination.}
    GNC is initialized with a convex surrogate at large $\mu_0$.
    The continuation loop terminates when $\mu_k$ reaches a target $\mu_{\min}$ (related to expected inlier noise) or when the estimate $\hat{\bx}$ stabilizes.
    Proper initialization and schedule tuning are critical for convergence to global solutions.
\end{itemize}
A generic GNC algorithm proceeds as follows:
\begin{enumerate}
    \item \textbf{Initialization.} Set $\mu_0$ large (nearly convex) and solve $\min_{\bx}\sum_i r_i^2(\bx)$ to obtain $\bx^{(0)}$.
    
    \item \textbf{Continuation loop ($k \!=\! 1,2,\dots$).}
    \begin{enumerate}
        \item Solve the IRLS problem~\eqref{eq:gnc-irls} at $\mu_k$, warm-started from $\bx^{(k-1)}$, to obtain $\bx^{(k)}$.
        
        \item Update the continuation parameter: $\mu_{k+1} \!=\! \alpha\mu_k$ (or adaptively).
    \end{enumerate}
    
    \item \textbf{Termination.} Stop when $\mu_k \!\le\! \mu_{\min}$ or $\|\bx^{(k)} \!-\! \bx^{(k-1)}\| \!<\! \epsilon$.
\end{enumerate}

\input{fig/gnc}

\cref{fig:gnc} illustrates the core mechanism of GNC: as the continuation parameter $\mu$ decreases, the surrogate loss transitions from convex to nonconvex (left panel), while the optimization tracks a solution path through progressively more challenging landscapes (right panel), leveraging warm-starting to escape poor local minima and converge to near-global solutions.

\PAR{Theoretical properties.}
Unlike BnB or convex relaxations, GNC does not provide \emph{a priori} global optimality guarantees or certificates.
Instead, it operates on the principle of \emph{deterministic annealing}~\cite{rose2002deterministic} and inherits convergence properties from continuation methods.
Under smoothness and local convexity assumptions on $\rho(\cdot,\mu)$, each subproblem~\eqref{eq:gnc-irls} converges to a local minimum, and the full continuation is guaranteed to reach a stationary point of the final objective under mild regularity conditions~\cite{black1996unification,mobahi2015link}.
The key question is whether this stationary point is globally optimal.
In low-noise regimes or when outliers are well-separated, GNC often converges to near-global minima empirically, especially when initialized from a good convex approximation~\cite{yang2020graduated}.
The progressive down-weighting mechanism $w_i(\mu)$ provides inherent robustness to contaminated data. Unlike consensus maximization that enumerates inlier sets or M-estimation in convex relaxations, GNC achieves outlier suppression deterministically through the continuation path itself, handling outlier ratios up to $30$--$50\%$ reliably in practice.
However, GNC may fail under extreme outlier ratios ($>50\%$), adversarial configurations where outliers cluster near the inlier distribution, or poor schedule tuning where $\mu$ decreases too rapidly.
Despite its lack of optimality guarantees, GNC's deterministic nature, empirical reliability, and computational efficiency make it a practical choice for large-scale robust estimation.

\PAR{Computational complexity.}
GNC's computational cost is dominated by solving the weighted least-squares subproblems~\eqref{eq:gnc-irls} at each continuation step.
For $N$ measurements and $n$ parameters, each IRLS iteration costs $O(Nn^2)$ (Gauss-Newton) or $O(Nn+n^3)$ (direct factorization), with typically $O(\log(1/\mu_{\min}))$ continuation steps under geometric schedules.
Unlike BnB (exponential in dimension) or SDP (cubic in lifted dimension), GNC scales linearly with the number of measurements, enabling large-scale application.

\PAR{Variants and extensions.}
The GNC framework is a general template compatible with various robust penalties and adaptive strategies.
\begin{itemize}
    \item \textbf{Penalty variants.}
    GNC-GM~\cite{yang2020graduated} provides smooth transitions with strong theoretical properties, while GNC-TLS~\cite{yang2020graduated,antonante2021outlier} offers sharper outlier rejection at the cost of less smooth homotopy paths.
    
    \item \textbf{Adaptive annealing.} 
    Adaptive strategies~\cite{sidhartha2023adaptive,chitturi2024adaptive} dynamically adjust $\mu$ based on local convexity (Hessian definiteness~\cite{sidhartha2023adaptive}, Laplacian structure~\cite{chitturi2024adaptive}) to improve robustness and convergence reliability.
    
    \item \textbf{Minimally tuned GNC (GNC-MinT).}
    The minimally tuned version of GNC~\cite{antonante2021outlier} introduces an inlier-threshold-free termination criterion based on residual distribution analysis, reducing dependence on prior knowledge of noise levels or inlier thresholds.

    \item \textbf{Hybrid strategies.}
    GNC can be combined with certifiable methods for enhanced reliability: GNC provides efficient near-global solutions that can be certified or refined using convex relaxations~\cite{yang2020teaser,yang2022certifiably}.
\end{itemize}

\PAR{Strengths and limitations.}

\begin{itemize}
    \item \textbf{Strengths.}
    GNC offers three key advantages:
    \begin{enumerate}
        \item \emph{Deterministic robustness}: Unlike sampling-based methods (\eg RANSAC~\cite{fischler1981random}), GNC achieves outlier suppression deterministically through progressive down-weighting, handling $30$--$50\%$ outlier ratios reliably.
        \item \emph{Scalability}: GNC scales linearly with the number of measurements, enabling large-scale applications where BnB or convex relaxations become computationally prohibitive.
        \item \emph{Broad applicability}: Its conceptual simplicity and compatibility with various robust losses make it a foundational tool across diverse 3D vision tasks.
    \end{enumerate}

    \item \textbf{Limitations.}
    The major limitation of GNC is the lack of \emph{a priori} global optimality guarantees.
    While it often converges to near-global minima in practice, failure modes exist under extreme outlier ratios ($>50\%$), adversarial configurations, or poor schedule tuning.
    Performance depends on a proper initial convex solver and parameter tuning ($\mu_0$, $\alpha$, $\mu_{\min}$), though recent adaptive variants reduce this sensitivity.
    Unlike BnB or convex relaxations, GNC cannot certify global optimality without additional verification steps.
\end{itemize}

%% file: fig/gnc.tex
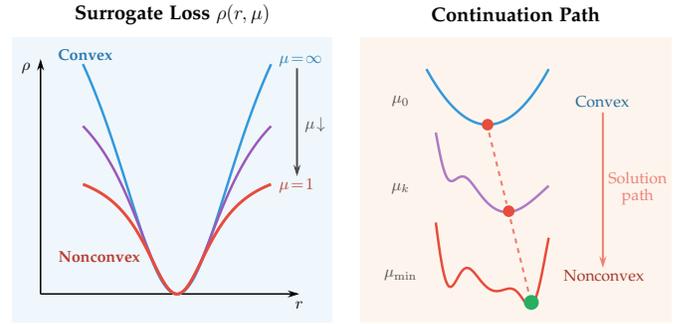
\begin{figure}[t]
\centering

\definecolor{domainColor}{HTML}{3498DB}
\definecolor{relaxColor}{HTML}{E67E22}
\definecolor{optColor}{HTML}{27AE60}
\definecolor{boundColor}{HTML}{9B59B6}
\definecolor{pathColor}{HTML}{E74C3C}

\definecolor{convexColor}{HTML}{3498DB}
\definecolor{intermediateColor}{HTML}{9B59B6}
\definecolor{robustColor}{HTML}{E74C3C}

\resizebox{\columnwidth}{!}{
\begin{tikzpicture}[
    >={Stealth[length=2.5mm, width=1.5mm]},
    font=\small,
    every node/.style={font=\small}
]

\begin{scope}[shift={(0,0)}]
    
    \fill[domainColor!8] (-1.1, -0.6) rectangle (5.7, 5.45);
    
    \node[font=\large\bfseries] at (2.3, 5.95) {Surrogate Loss $\rho(r,\mu)$};
    
    \draw[->, line width=0.9pt] (-0.5, 0) -- (5.0, 0);
    \draw[->, line width=0.9pt] (-0.5, 0) -- (-0.5, 5.0);
    
    \node[below, font=\normalsize] at (5, -0.05) {$r$};
    \node[left, font=\normalsize] at (-0.6, 4.8) {$\rho$};
    
    \draw[convexColor, line width=1.5pt, smooth, samples=50, domain=0.4:4.4]  
        plot (\x, {4*(\x-2.4)*(\x-2.4)/(4*0.8+(\x-2.4)*(\x-2.4))*2.2});
    
    \draw[intermediateColor, line width=1.5pt, smooth, samples=100, domain=0.4:4.4] 
        plot (\x, {2*(\x-2.4)*(\x-2.4)/(2*0.8+(\x-2.4)*(\x-2.4))*2.5});
    
    \draw[robustColor, line width=1.8pt, smooth, samples=100, domain=0.4:4.4] 
        plot (\x, {(\x-2.4)*(\x-2.4)/(0.8+(\x-2.4)*(\x-2.4))*2.8});
    
    \draw[->, line width=1.2pt, black!70] (4.95, 4.8) -- (4.95, 2.5);
    \node[right, font=\normalsize, black!70] at (5, 3.6) {$\mu\!\downarrow$};
    
    \node[right, font=\normalsize, convexColor!80!black] at (4.45, 4.95){$\mu\!=\!\infty$};
    \node[right, font=\normalsize, robustColor!80!black] at (4.45, 2.3) {$\mu\!=\!1$};
    
    \node[font=\normalsize, convexColor!80!black, align=center] 
        at (0.45, 5.1) {\textbf{Convex}};
    \node[font=\normalsize, robustColor!80!black, align=center] 
        at (0.75, 0.8) {\textbf{Nonconvex}};
    
\end{scope}

\begin{scope}[shift={(6.6,0)}]
    
    \fill[relaxColor!8] (-0.3, -0.6) rectangle (6.3, 5.45);
    
    \node[font=\large\bfseries] at (3, 5.95) {Continuation Path};
    
    \begin{scope}[shift={(0, 3.2)}]
        \draw[convexColor, line width=1.5pt, smooth, samples=50, domain=1.1:3.7]
            plot (\x, {0.7*(\x-2.4)*(\x-2.4) + 0.4});
        
        \fill[pathColor] (2.4, 0.4) circle (3.5pt);
        
        \node[font=\normalsize, black!70] at (0.55, 0.9) {$\mu_0$};
        
        \node[right, font=\normalsize, convexColor!70!black] at (4.15, 0.9) {Convex};
    \end{scope}
    
    \begin{scope}[shift={(0, 1.6)}]
        \draw[intermediateColor!80, line width=1.5pt, smooth, samples=80, domain=1.3:3.7]
            plot (\x, {0.5*(\x-3.3)*(\x-3.3) + 0.4*sin(\x*110) - 1.1*exp(-(\x-1.5)*(\x-1.5)/0.1) + 0.33});
        
        \fill[pathColor] (2.85, 0.16) circle (3.5pt);
        
        \node[font=\normalsize, black!70] at (0.55, 0.65) {$\mu_k$};
    \end{scope}
    
    \begin{scope}[shift={(0, 0)}]
        \draw[robustColor, line width=1.5pt, smooth, samples=100, domain=1.3:3.7]
            plot (\x, {0.5*(\x-3)*(\x-3) + 0.95*sin(\x*110) - 1.95*exp(-(\x-1.5)*(\x-1.5)/0.12) - 1.35*exp(-(\x-3.4)*(\x-3.4)/0.12) + 0.9});
        
        \fill[optColor] (3.33, -0.18) circle (4.5pt);
        
        \node[font=\normalsize, black!70] at (0.55, 0.4) {$\mu_{\min}$};
        
        \node[right, font=\normalsize, robustColor!70!black] at (3.9, 0.4) {Nonconvex};
    \end{scope}
    
    \draw[pathColor!70, line width=1.3pt, dashed] (2.43, 3.43) -- (2.8042, 1.92);
    \draw[pathColor!70, line width=1.3pt, dashed] (2.8865, 1.588) -- (3.27, 0.04);
    
    \draw[->, line width=1.2pt, pathColor!70] (4.85, 3.86) -- (4.85, 0.55);
    \node[right, font=\normalsize, pathColor!70, align=center] at (4.85, 2.25) {Solution\\[-1pt]path};z
    
\end{scope}

\end{tikzpicture}
}
\caption{%
    Graduated Non-Convexity (GNC) for robust optimization.
    \textbf{Left}: The surrogate loss $\rho(r,\mu)$ transitions from \textcolor{convexColor}{convex} (large $\mu$) to \textcolor{robustColor}{nonconvex} (small $\mu$) via homotopy, enabling progressive outlier down-weighting.
    \textbf{Right}: By solving a sequence of subproblems with decreasing $\mu$ (top to bottom), GNC tracks a \textcolor{pathColor!70}{solution path} that starts from the global minimum of the convex surrogate and converges to a \textcolor{optColor}{near-global solution} of the target nonconvex objective through warm-starting.
}
\label{fig:gnc}
\end{figure}

%% file: sec/6_comparison.tex
\section{Comparative Analysis}
\label{sec:compare}

\input{tab/overview_comparison}

The global solvers reviewed in~\cref{sec:bnb,sec:relaxation,sec:gnc} offer distinct trade-offs in optimality guarantees, computational complexity and scalability, outlier robustness, and practical deployment.
\cref{tab:comparison} summarizes these trade-offs across the major solver families, while subsequent discussion examines each dimension in detail and provides guidance for method selection in different application scenarios.

\subsection{Global Optimality Guarantees}
\label{subsec:optimality}

BnB provides the strongest theoretical guarantee: \emph{guaranteed} global optimality with certificates upon convergence, at the cost of exponential worst-case complexity.
Shor's relaxation and Moment-SOS methods offer \emph{certifiable} solutions when the relaxation is tight, providing duality-based certificates and suboptimality bounds even when not exact.
The tightness of these relaxations depends on the problem structure and relaxation order, with Moment-SOS offering stronger guarantees through hierarchical refinement but at increased computational cost.
Other specialized relaxations provide global or near-global guarantees, often exact for particular problem classes (\eg quasi-convex formulations for $L_\infty$ objectives).
GNC provides \emph{no a priori} optimality guarantees or certificates, while it often achieves near-global solutions empirically under favorable conditions.

\subsection{Computational Complexity and Scalability}
\label{subsec:complexity}

The methods exhibit a clear trade-off between optimality guarantees and computational efficiency.
BnB's exponential complexity in problem dimension limits its applicability to small-scale problems (typically $\!<\! 20$ measurements), though advanced pruning and bound tightening can extend this range.
Shor's relaxation scales to medium-scale problems (hundreds of measurements) with $O(n^{3.5})$ complexity per SDP solve (assuming $m \!=\! O(n)$ constraints), while Moment-SOS methods are more restrictive, scaling as $O(r^{3.5})$ where $r \!=\! \binom{n \!+\! d}{d}$ grows rapidly with relaxation order $d$, along with $O\big(\binom{n+2d}{2d}\big)$ linear equality constraints.
In practice, Moment-SOS is effective for small- to medium-scale problems (typically $\!<\!100$ measurements) at low orders ($d \!=\! 2$--$4$), though sparsity exploitation can extend this range.
Other specialized relaxations offer superior scalability by exploiting problem structure, with SOCP, QP, LP, and convex-hull methods handling thousands of measurements.
GNC achieves the best scalability, with $O(Kn^2)$ complexity per iteration, enabling large-scale and real-time applications.

\subsection{Robustness to Outliers}
\label{subsec:robustness}

Outlier handling capabilities vary significantly across methods.
BnB naturally integrates with consensus maximization, providing robust solutions at the cost of discrete search over inlier and outlier assignments.
Shor's and Moment-SOS relaxations incorporate robustness through M-estimation or consensus maximization embedded within the convex formulation, though this requires explicit robust modeling rather than inherent robustness mechanisms.
The convex relaxation framework provides principled bounds for computing robust estimates.
GNC offers inherent robustness through progressive down-weighting, reliably handling $30$--$50\%$ outlier ratios without explicit inlier and outlier classification.
This deterministic robustness mechanism, combined with excellent scalability, makes GNC particularly attractive for large-scale robust estimation.
Other specialized relaxations typically have limited outlier handling capabilities, though some variants incorporate robust formulations.

\subsection{Method Selection and Deployment}
\label{subsec:selection}

Method selection depends on problem characteristics and application requirements.
For small-scale problems where certified global optimality is critical, BnB or tight convex relaxations are preferable despite higher computational cost.
Shor's relaxation offers a good balance for medium-scale problems, providing certificates when tight while remaining computationally tractable.
Moment-SOS is best suited for small problems where the hierarchical structure can be exploited at low orders.
For large-scale or time-critical applications, GNC or specialized relaxations are the methods of choice, trading optimality certificates for scalability and efficiency.
In practice, hybrid approaches combining multiple methods can leverage their complementary strengths, such as that GNC for efficient initialization, followed by convex relaxations for certification when needed.

%% file: tab/overview_comparison.tex
\begin{table*}[t]
    \centering
    \caption{
        Comparative analysis of global solvers for 3D vision.
        Each method offers distinct trade-offs in optimality guarantees, 
        computational complexity, outlier robustness, and scalability.
    }
    \label{tab:comparison}
    \resizebox{\textwidth}{!}{%
        \begin{tabular}{l|cccccc}
            \toprule
            \textbf{Method} & 
            \textbf{Optimality} & 
            \textbf{Outliers} & 
            \textbf{Per-Solve Complexity} & 
            \textbf{Memory} & 
            \textbf{Scalability} & 
            \textbf{Certificates} \\
            \midrule
            Branch-and-Bound & 
            Global & 
            CM & 
            Exponential in $n$ & 
            $O(2^n)$ & 
            Poor ($N \!<\! 20$) & 
            Yes \\
            \midrule
            Shor's Relaxation & 
            Global (if tight) & 
            M-est & 
            $O(n^{3.5})$ & 
            $O(n^2)$ & 
            Medium ($100 \!<\! N \!<\! 1000$) & 
            Yes (if tight) \\
            \midrule
            Moment-SOS Relaxation & 
            Global (if tight) & 
            CM, M-est & 
            $O(r^{3.5})$, $r \!=\! \binom{n\!+\!d}{d}$ & 
            $O(r^2)$ & 
            Low to medium ($N \!<\! 100$) & 
            Yes (if tight) \\
            \midrule
            Other Relaxations & 
            Global or Near-global & 
            Problem-specific & 
            $O(n^3)$ to $O(n)$ & 
            $O(n^2)$ to $O(n)$ & 
            High ($N \!>\! 1000$) & 
            Problem-specific \\
            \midrule
            Graduated Non-Convexity & 
            Near-global & 
            Built-in ($30$--$50\%$ ratio) & 
            $O(Kn^2)$ per iteration & 
            $O(Kn)$ & 
            Excellent ($N \!>\! 1000$) & 
            No \\
            \bottomrule
        \end{tabular}
    }
    \\[2pt]
    {\scriptsize\raggedright
        $n$: dimension of model parameter space;
        $d$: relaxation order;
        $K$: number of GNC iterations (typically $10$--$50$);
        $N$: number of measurements;
        CM: consensus maximization;
        M-est: M-estimation.
    \par}
\end{table*}

%% file: sec/7_task.tex
\section{Tasks and Applications}
\label{sec:app}

\input{tab/task_formulation}

Building on the theoretical foundations established in~\cref{sec:bnb,sec:relaxation,sec:gnc}, we now examine how global solvers perform on concrete problems in 3D vision.
Task structure varies substantially across applications, from manifold constraints on $\mathrm{SO}(3)$ and $\mathrm{SE}(3)$, to algebraic relations in projection geometry, to graph sparsity in multi-view reconstruction, and varying prevalence of outliers, leading the same optimization principles to exhibit markedly different performance characteristics.
This section therefore analyzes task-specific structures and representative approaches, evaluating methods along three practical axes:
\emph{Optimality} (certifiability and exact-recovery conditions),
\emph{Robustness} (tolerance to noise and outliers), and
\emph{Scalability} (computational complexity and large-scale feasibility).

We organize the coverage by problem domain.
\cref{subsec:wahba,subsec:pnp,subsec:vp} address single-view geometry;
\cref{subsec:relpose,subsec:registration} covers two-view geometry;
\cref{subsec:rot-avg,subsec:trans-avg,subsec:triangulation} discuss multi-view geometry;
\cref{subsec:pgo,subsec:bundle-adjustment} present large-scale multi-view problems.
\cref{fig:tasks-overview} provides a visual overview of these tasks, illustrating the problem setups, geometric relationships, and variables to be estimated in each case.
For each task, we follow a consistent presentation template:
(1)~problem definition and mathematical formulation,
(2)~properties and local optimization methods,
(3)~closed-form or analytical solutions when available,
(4)~BnB approaches,
(5)~Shor's relaxations,
(6)~Moment-SOS relaxations,
(7)~alternative convex formulations or global methods, and
(8)~GNC methods.
This structure enables systematic comparison of how different global solvers address the specific challenges of each geometric estimation problem, revealing where global optimality is achievable (through tight relaxations or exact recovery conditions), where robustness mechanisms are essential (via consensus maximization or robust cost functions), and how scalability is attained (through problem structure exploitation, efficient solvers, or approximation strategies).

\cref{tab:task-formulation} presents the problem formulations for each task, while \cref{fig:solver-task-index} provides a comprehensive index of representative methods organized by solver family.

\input{fig/tasks/tasks_overview}
\input{fig/solver_task_index}

\input{sec/task_subsec/1_wahba}
\input{sec/task_subsec/2_vp}
\input{sec/task_subsec/3_pnp}
\input{sec/task_subsec/4_relpose}
\input{sec/task_subsec/5_reg}
\input{sec/task_subsec/6_ra}
\input{sec/task_subsec/7_ta}
\input{sec/task_subsec/8_tri}
\input{sec/task_subsec/9_pgo}
\input{sec/task_subsec/10_ba}

%% file: tab/task_formulation.tex
\begin{table*}[t]
\centering
\scriptsize
\setlength{\tabcolsep}{20pt}
\caption{%
    Problem formulations for 3D vision tasks.
    Each task is characterized by its unknowns (model parameters), inputs (measurements), and residual function (minimized in least squares).
    The residual $r(\cdot)$ corresponds to the generic formulation $\min_{\bx} \sum_{i=1}^N r^2(\bx, y_i)$ from~\eqref{eq:generic-ls}.
}
\label{tab:task-formulation}
\resizebox{\textwidth}{!}{%
\begin{tabular}{l|c|c|c}
\toprule
\textbf{Task} & 
\textbf{Unknowns} & 
\textbf{Inputs} & 
\textbf{Residual Function $r(\cdot)$} \\
\midrule
Wahba Problem & 
$\bR \!\in\! \mathrm{SO}(3)$ & 
$\{\ba_i, \bbb_i\}$ & 
$ \bbb_i - \bR \ba_i $ \\
\midrule
Vanishing Point Estimation & 
$\{\bv_i \!\in\! \bbR^3\}$ & 
$\{\ell_j\}$ & 
$ \bn_j^\top \bd_i $ \\
\midrule
Absolute Pose Estimation & 
$(\bR,\bt) \!\in\! \mathrm{SE}(3)$ & 
$\{\bX_i, \bx_i\}$ &
$\pi(\bR \bX_i+\bt) - \bx_i$ \\
\midrule
Relative Pose Estimation & 
$(\bR,\bt) \!\in\! \mathrm{SE}(3)$ & 
$\{\bx_{0i}, \bx_{1i}\}$ & 
$\bx_{1i}^\top \bF \bx_{0i}$ / $\hat{\bx}_{1i}^\top \bE \hat{\bx}_{0i}$ \\
\midrule
3D Registration & 
$(s,\bR,\bt) \!\in\! \pR \!\times\! \mathrm{SE}(3)$ & 
$\{\bp_i, \bq_i\}$ & 
$\bq_i - (s\bR \bp_i + \bt)$ \\
\midrule
Rotation Averaging & 
$\{\bR_i \!\in\! \mathrm{SO}(3)\}$ &
$\{\bR_{ij}\}$ & 
$d_\bR(\bR_{ij}, \bR_i^{-1} \bR_j)$ \\
\midrule
Translation Averaging & 
$\{\bt_i \!\in\! \bbR^3\}$ & 
$\{\bt_{ij}\}$ & 
$(\bt_j - \bt_i)/\|\bt_j - \bt_i\|_2 - \bt_{ij}$ \\
\midrule
Triangulation & 
$\bX \!\in\! \bbR^3$ & 
$\{\bx_i\}, \{\bK_i\}, \{\bR_i, \bt_i\}$ & 
$\pi\big(\bK_i (\bR_i \bX + \bt_i)\big) - \bx_i$ \\
\midrule
Pose Graph Optimization & 
$\left\{ (\bR_i,\bt_i) \!\in\! \mathrm{SE}(3) \right\}$ & 
$\{\bR_{ij}, \bt_{ij}\}$ & 
$d_\bR(\bR_{ij}, \bR_i^{-1} \bR_j)$; $\bt_j - (\bR_i \bt_{ij} + \bt_i)$ \\
\midrule
Bundle Adjustment & 
$\left\{ (\bR_i,\bt_i) \!\in\! \mathrm{SE}(3), \bX_j \!\in\! \bbR^3 \right\}$ & 
$\{\bx_{ij}\}, \{\bK_i\}$ & 
$\pi\big(\bK_i (\bR_i \bX_j + \bt_i)\big) - \bx_{ij}$ \\
\bottomrule
\end{tabular}
}
\\[2pt]
{\scriptsize\raggedright
    $\pi$: perspective projection; 
    $\bF$: fundamental matrix; 
    $\bE$: essential matrix; 
    $\hat{\bx}$: normalized coordinates;
    $d_\bR$: rotation distance;
    $\bd_i \!=\! \bK^{-1}\bv_i$.
\par}

\end{table*}

%% file: fig/tasks/tasks_overview.tex
\begin{figure*}[t]
\centering

\definecolor{domainColor}{HTML}{3498DB}      
\definecolor{relaxColor}{HTML}{E67E22}       
\definecolor{optColor}{HTML}{27AE60}         
\definecolor{boundColor}{HTML}{9B59B6}       
\definecolor{pathColor}{HTML}{E74C3C}        
\definecolor{observColor}{HTML}{F39C12}      
\definecolor{axisRed}{RGB}{230,50,50}
\definecolor{axisGreen}{RGB}{50,180,50}
\definecolor{axisBlue}{RGB}{50,50,230}

\resizebox{\textwidth}{!}{
\begin{tikzpicture}[
    >={Stealth[length=2.5mm, width=1.5mm]},
    font=\scriptsize,
    every node/.style={font=\scriptsize}
]

\pgfmathsetmacro{\cellW}{4.2}
\pgfmathsetmacro{\cellH}{5.5}
\pgfmathsetmacro{\gapX}{0.1}
\pgfmathsetmacro{\gapY}{0.1}
\pgfmathsetmacro{\stepX}{\cellW+\gapX}
\pgfmathsetmacro{\stepY}{\cellH+\gapY}

\input{fig/tasks/wahba}

\input{fig/tasks/vp}

\input{fig/tasks/pnp}

\input{fig/tasks/relpose}

\input{fig/tasks/reg}

\pgfmathsetmacro{\rowOffset}{-\stepY}

\input{fig/tasks/ra}

\input{fig/tasks/ta}

\input{fig/tasks/tri}

\input{fig/tasks/pgo}

\input{fig/tasks/ba}

\end{tikzpicture}
}

\caption{%
    Overview of geometric estimation tasks in this survey.
    \textbf{(a)}~Wahba problem: estimate rotation $\bR \!\in\! \mathrm{SO}(3)$ from vector correspondences $\{\ba_i, \bbb_i\}$;
    \textbf{(b)}~Vanishing point estimation: detect convergence points $\{\bv_i\}$ of parallel line families $\{\ell_j\}_{1,2,3}$;
    \textbf{(c)}~Absolute pose estimation: recover camera pose $(\bR, \bt)$ from 3D-2D correspondences $\{\bX_i, \bx_i\}$;
    \textbf{(d)}~Relative pose estimation: compute essential/fundamental matrix from 2D-2D correspondences $\{\bx_{0i}, \bx_{1i}\}$;
    \textbf{(e)}~3D registration: align point clouds $\{\bp_i, \bq_i\}$ via rigid transformation;
    \textbf{(f)}~Rotation averaging: recover absolute orientations $\{\bR_i\}$ from pairwise rotation measurements $\{\bR_{ij}\}$;
    \textbf{(g)}~Translation averaging: estimate absolute positions $\{\bt_i\}$ from pairwise translation direction measurements $\{\bt_{ij}\}$;
    \textbf{(h)}~Triangulation: reconstruct 3D point $\bX$ from multi-view observations $\{\bx_i\}$;
    \textbf{(i)}~Pose graph optimization: refine camera trajectory $\{\bT_i\}$ with loop closures;
    \textbf{(j)}~Bundle adjustment: jointly optimize camera poses $\{\bR_i, \bt_i\}$ and structure $\{\bX_j\}$.
}
\label{fig:tasks-overview}
\end{figure*}
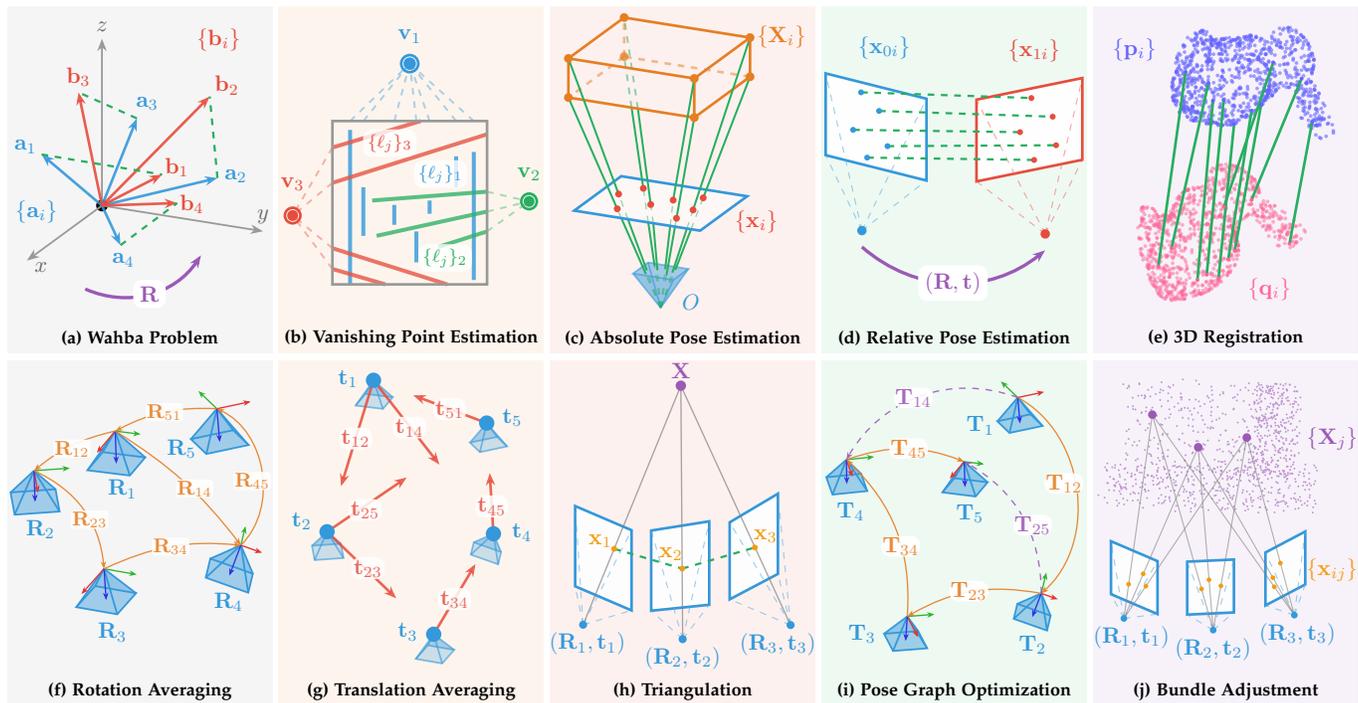

%% file: fig/tasks/wahba.tex
\begin{scope}[shift={(0, 0)}]
    \fill[gray!8] (0,0) rectangle (\cellW, \cellH);

    \begin{scope}[shift={(1.5, 2.35)}, scale=0.88]
        \tdplotsetmaincoords{70}{115}
        \begin{scope}[tdplot_main_coords]

            \coordinate (O) at (0,0,0);
            \fill[black] (O) circle (3pt);
            
            \draw[-{Stealth[length=2mm]}, thick, black!40] (O) -- (3.2,0,0) node[anchor=west, font=\normalsize, black!60, shift={(0,-0.1cm)}] {$x$};
            \draw[-{Stealth[length=2mm]}, thick, black!40] (O) -- (0,3.2,0) node[anchor=south, font=\normalsize, black!60] {$y$};
            \draw[-{Stealth[length=2mm]}, thick, black!40] (O) -- (0,0,3.2) node[anchor=south, font=\normalsize, black!60] {$z$};
            
            \coordinate (a1) at (3, 0.2, 2);
            \coordinate (a2) at (-0.2, 2.2, 0.8);
            \coordinate (a3) at (1.3, 1.3, 2.3);
            \coordinate (a4) at (2, 1.3, 0.1);
            
            \draw[-{Stealth[length=2mm]}, domainColor, line width=1.2pt, opacity=0.9] (O) -- (a1);
            \draw[-{Stealth[length=2mm]}, domainColor, line width=1.2pt, opacity=0.9] (O) -- (a2);
            \draw[-{Stealth[length=2mm]}, domainColor, line width=1.2pt, opacity=0.9] (O) -- (a3);
            \draw[-{Stealth[length=2mm]}, domainColor, line width=1.2pt, opacity=0.9] (O) -- (a4);
            
            \node[domainColor, font=\normalsize] at ($(a1)+(0.25,-0.2,0.2)$) {$\mathbf{a}_1$};
            \node[domainColor, font=\normalsize] at ($(a2)+(-0.25,0.25,0)$) {$\mathbf{a}_2$};
            \node[domainColor, font=\normalsize] at ($(a3)+(0.2,0.3,0.4)$) {$\mathbf{a}_3$};
            \node[domainColor, font=\normalsize] at ($(a4)+(0.3,0.2,-0.15)$) {$\mathbf{a}_4$};
            
            \coordinate (b1) at (1.3, 1.8, 1.3);
            \coordinate (b2) at (-0.3, 2.0, 2.3);
            \coordinate (b3) at (1, 0, 2.5);
            \coordinate (b4) at (1.3, 2.1, 0.8);
            
            \draw[-{Stealth[length=2mm]}, pathColor, line width=1.2pt, opacity=0.9] (O) -- (b1);
            \draw[-{Stealth[length=2mm]}, pathColor, line width=1.2pt, opacity=0.9] (O) -- (b2);
            \draw[-{Stealth[length=2mm]}, pathColor, line width=1.2pt, opacity=0.9] (O) -- (b3);
            \draw[-{Stealth[length=2mm]}, pathColor, line width=1.2pt, opacity=0.9] (O) -- (b4);
            
            \node[pathColor, font=\normalsize] at ($(b1)+(-0.1,0.25,0.15)$) {$\mathbf{b}_1$};
            \node[pathColor, font=\normalsize] at ($(b2)+(-0.3,0.2,0.15)$) {$\mathbf{b}_2$};
            \node[pathColor, font=\normalsize] at ($(b3)+(0,0,0.3)$) {$\mathbf{b}_3$};
            \node[pathColor, font=\normalsize] at ($(b4)+(-0.2,0.2,-0.05)$) {$\mathbf{b}_4$};
            
            \draw[optColor, dashed, line width=1pt, opacity=1] (a1) -- (b1);
            \draw[optColor, dashed, line width=1pt, opacity=1] (a2) -- (b2);
            \draw[optColor, dashed, line width=1pt, opacity=1] (a3) -- (b3);
            \draw[optColor, dashed, line width=1pt, opacity=1] (a4) -- (b4);
            
            \node[domainColor, font=\normalsize\bfseries, anchor=south west] at (2.5, -0.7, 0.25) {
                $\{\mathbf{a}_i\}$
            };
            \node[pathColor, font=\normalsize\bfseries, anchor=south west] at (-0.5, 1.5, 2.9) {
                $\{\mathbf{b}_i\}$
            };
            
            \draw[-{Stealth[length=2mm]}, boundColor, line width=1.5pt] 
                (0.5, -0.1, -1.45) to[bend right=40] 
                node[midway, font=\normalsize\bfseries, fill=white, inner sep=3pt, rounded corners] {$\mathbf{R}$} 
                (0.5, 2.2, -0.45);

        \end{scope}
    \end{scope}
    
    \node[font=\footnotesize\bfseries] at (\cellW/2, 0.25) {(a) Wahba Problem};
    
\end{scope}

%% file: fig/tasks/vp.tex
\begin{scope}[shift={(\stepX, 0)}]
    \fill[relaxColor!8] (0,0) rectangle (\cellW, \cellH);

    \begin{scope}[shift={(0.85, 1.1)}, scale=0.7]
        \tdplotsetmaincoords{0}{0}
        \begin{scope}[tdplot_main_coords]
            
            \coordinate (vp1) at (1.75, 5);
            \coordinate (vp2) at (4.45, 1.88);
            \coordinate (vp3) at (-0.9, 1.56);
            
            \draw[domainColor, line width=1.8pt, opacity=0.85] (0.4, 0) -- (0.4, 3.5);
            \draw[domainColor, line width=1.8pt, opacity=0.85] (0.7, 1.1) -- (0.7, 2.2);
            \draw[domainColor, line width=1.8pt, opacity=0.85] (1.4, 1.35) -- (1.4, 1.8);
            \draw[domainColor, line width=1.8pt, opacity=0.85] (1.9, 0.5) -- (1.9, 1.2);
            \draw[domainColor, line width=1.8pt, opacity=0.85] (2.2, 1.6) -- (2.2, 1.9);
            \draw[domainColor, line width=1.8pt, opacity=0.85] (2.8, 2.2) -- (2.8, 2.9);
            \draw[domainColor, line width=1.8pt, opacity=0.85] (3.2, 0.1) -- (3.2, 3);
            
            \draw[domainColor, line width=0.9pt, opacity=0.4, dashed] (0.4, 3.7) -- (vp1);
            \draw[domainColor, line width=0.9pt, opacity=0.4, dashed] (0.7, 3.7) -- (vp1);
            \draw[domainColor, line width=0.9pt, opacity=0.4, dashed] (1.4, 3.7) -- (vp1);
            \draw[domainColor, line width=0.9pt, opacity=0.4, dashed] (1.9, 3.7) -- (vp1);
            \draw[domainColor, line width=0.9pt, opacity=0.4, dashed] (2.2, 3.7) -- (vp1);
            \draw[domainColor, line width=0.9pt, opacity=0.4, dashed] (2.8, 3.7) -- (vp1);
            \draw[domainColor, line width=0.9pt, opacity=0.4, dashed] (3.2, 3.7) -- (vp1);
            
            \draw[optColor, line width=1.8pt, opacity=0.8] (2.4, 0.85) -- (3.5, 1.2);
            \draw[optColor, line width=1.8pt, opacity=0.8] (1.0, 1.1) -- (3.5, 1.65);
            \draw[optColor, line width=1.8pt, opacity=0.8] (0.9, 1.9) -- (3.5, 2.1);
            
            \draw[optColor, line width=0.9pt, opacity=0.4, dashed] (3.5, 1.2) -- (vp2);
            \draw[optColor, line width=0.9pt, opacity=0.4, dashed] (3.5, 1.65) -- (vp2);
            \draw[optColor, line width=0.9pt, opacity=0.4, dashed] (3.5, 2.1) -- (vp2);
            
            \draw[pathColor, line width=1.8pt, opacity=0.8] (0, 3.1) -- (2.0, 3.7);
            \draw[pathColor, line width=1.8pt, opacity=0.8] (0, 2.3) -- (3.5, 3.4);
            \draw[pathColor, line width=1.8pt, opacity=0.8] (0, 0.82) -- (2.6, 0);
            \draw[pathColor, line width=1.8pt, opacity=0.8] (0, 0.35) -- (1.2, 0);
            
            \draw[pathColor, line width=0.9pt, opacity=0.4, dashed] (0, 3.1) -- (vp3);
            \draw[pathColor, line width=0.9pt, opacity=0.4, dashed] (0, 2.3) -- (vp3);
            \draw[pathColor, line width=0.9pt, opacity=0.4, dashed] (0, 0.82) -- (vp3);
            \draw[pathColor, line width=0.9pt, opacity=0.4, dashed] (0, 0.35) -- (vp3);
            
            \fill[domainColor] (vp1) circle (4pt);
            \draw[domainColor, very thick] (vp1) circle (5.5pt);
            \node[domainColor, font=\normalsize\bfseries, above=6pt] at (vp1) {$\bv_1$};
            
            \fill[optColor] (vp2) circle (3.5pt);
            \draw[optColor, very thick] (vp2) circle (5pt);
            \node[optColor, font=\normalsize\bfseries, above=6pt] at (vp2) {$\bv_2$};
            
            \fill[pathColor] (vp3) circle (3.5pt);
            \draw[pathColor, very thick] (vp3) circle (5pt);
            \node[pathColor, font=\normalsize\bfseries, above=8pt] at (vp3) {$\bv_3$};
            
            \node[domainColor, font=\footnotesize, fill=white, fill opacity=0.8, 
                  text opacity=1, inner sep=2pt, rounded corners] at (2.45, 2.5) {$\{\ell_j\}_1$};
            \node[optColor, font=\footnotesize, fill=white, fill opacity=0.8,
                  text opacity=1, inner sep=2pt, rounded corners] at (2.55, 0.6) {$\{\ell_j\}_2$};
            \node[pathColor, font=\footnotesize, fill=white, fill opacity=0.8,
                  text opacity=1, inner sep=2pt, rounded corners] at (1.3, 3.2) {$\{\ell_j\}_3$};

            \coordinate (imgBL) at (0, 0);
            \coordinate (imgTR) at (3.5, 3.7);
            
            \fill[white, opacity=0.05] (imgBL) rectangle (imgTR);
            \draw[black!40, line width=1.2pt] (imgBL) rectangle (imgTR);

        \end{scope}
    \end{scope}
    
    \node[font=\footnotesize\bfseries] at (\cellW/2, 0.25) {(b) Vanishing Point Estimation};
    
\end{scope}

%% file: fig/tasks/pnp.tex
\begin{scope}[shift={(2*\stepX, 0)}]
    \fill[pathColor!8] (0,0) rectangle (\cellW, \cellH);
    
    \begin{scope}[shift={(1.75, 0.73)}, scale=0.44]
        \tdplotsetmaincoords{70}{25}
        \begin{scope}[tdplot_main_coords]
        
            \coordinate (O) at (0, 0, 0);
            
            \def\cs{0.8}
            \def\ch{1.4}
            \coordinate (cFL) at (-\cs, -\cs*0.8, \ch);
            \coordinate (cFR) at (\cs, -\cs*0.8, \ch);
            \coordinate (cBL) at (-\cs, \cs*0.8, \ch);
            \coordinate (cBR) at (\cs, \cs*0.8, \ch);
            
            \fill[domainColor,opacity=0.3] (O) -- (cBL) -- (cBR) -- cycle;
            \fill[domainColor,opacity=0.3] (O) -- (cBL) -- (cFL) -- cycle;
            \fill[domainColor,opacity=0.3] (O) -- (cFR) -- (cBR) -- cycle;
            \fill[domainColor,opacity=0.3] (O) -- (cFL) -- (cFR) -- cycle;
            \fill[domainColor,opacity=0.2] (cFL) -- (cFR) -- (cBR) -- (cBL) -- cycle;
            
            \draw[domainColor,thick,opacity=0.6] (O) -- (cFL);
            \draw[domainColor,thick,opacity=0.6] (O) -- (cFR);
            \draw[domainColor,thick,opacity=0.6] (O) -- (cBL);
            \draw[domainColor,thick,opacity=0.6] (O) -- (cBR);
            \draw[domainColor,thick,opacity=0.6] (cFL) -- (cFR) -- (cBR) -- (cBL) -- cycle;
            
            \def\imgZ{4.0}
            \def\imgW{2.5}
            \def\imgH{2.0}
            
            \coordinate (imgFL) at (-\imgW, -\imgH, \imgZ);
            \coordinate (imgFR) at (\imgW, -\imgH, \imgZ);
            \coordinate (imgBL) at (-\imgW, \imgH, \imgZ);
            \coordinate (imgBR) at (\imgW, \imgH, \imgZ);
            
            \def\objZ{11}
            \def\objS{2.5}
            
            \coordinate (Q1) at (-\objS, -\objS, \objZ-\objS);
            \coordinate (Q2) at (\objS, -\objS, \objZ-\objS);
            \coordinate (Q3) at (\objS, \objS-0.25, \objZ-\objS);
            \coordinate (Q4) at (-\objS, \objS-0.25, \objZ-\objS);
            \coordinate (Q5) at (-\objS, -\objS, \objZ+\objS-3.5);
            \coordinate (Q6) at (\objS, -\objS, \objZ+\objS-3.5);
            \coordinate (Q7) at (\objS, \objS-0.25, \objZ+\objS-3.5);
            \coordinate (Q8) at (-\objS, \objS-0.25, \objZ+\objS-3.5);
            
            \pgfmathsetmacro{\pscaleBot}{\imgZ/(\objZ-\objS)}
            \pgfmathsetmacro{\pscaleTop}{\imgZ/(\objZ+\objS)}
            
            \coordinate (p1) at ({-\objS*\pscaleBot}, {-\objS*\pscaleBot}, \imgZ);
            \coordinate (p2) at ({\objS*\pscaleBot}, {-\objS*\pscaleBot}, \imgZ);
            \coordinate (p3) at ({\objS*\pscaleBot-0.037}, {\objS*\pscaleBot}, \imgZ);
            \coordinate (p4) at ({-\objS*\pscaleBot-0.062}, {\objS*\pscaleBot}, \imgZ);
            \coordinate (p5) at ({-\objS*\pscaleTop-0.585}, {-\objS*\pscaleTop}, \imgZ+0.4);
            \coordinate (p6) at ({\objS*\pscaleTop+0.1}, {-\objS*\pscaleTop}, \imgZ+0.5);
            \coordinate (p7) at ({\objS*\pscaleTop+0.2}, {\objS*\pscaleTop}, \imgZ-0.35);
            \coordinate (p8) at ({-\objS*\pscaleTop}, {\objS*\pscaleTop}, \imgZ-0.4);
            
            \coordinate (b1) at (-0.631, -\imgH, 4.0);
            \coordinate (b2) at (1.502, -\imgH, 4.0);
            \coordinate (b3) at (2.1393, -\imgH, 4.0);
            \coordinate (b4) at (0.4395, -\imgH, 4.0);
            \coordinate (b5) at (-0.3702, -\imgH, 4.0);
            \coordinate (b6) at (1.31, -\imgH, 4.0);
            \coordinate (b7) at (1.974, -\imgH, 4.0);
            \coordinate (b8) at (0.6387, -\imgH, 4.0);
            
            \coordinate (Cb1) at (-0.5865, 0, 1.2);
            \draw[optColor, line width=0.9pt, dashed, opacity=0.5] (O) -- (Cb1);
            \draw[optColor, line width=0.9pt] (Cb1) -- (b1);
            
            \coordinate (Cb2) at (0.228, 0, 1.28);
            \draw[optColor, line width=0.6pt, dashed, opacity=0.5] (O) -- (Cb2);
            \draw[optColor, line width=0.9pt] (Cb2) -- (b2);
            
            \coordinate (Cb3) at (0.585, 0, 1.55);
            \draw[optColor, line width=0.6pt, dashed, opacity=0.5] (O) -- (Cb3);
            \draw[optColor, line width=0.9pt] (Cb3) -- (b3);
            
            \coordinate (Cb4) at (-0.24, 0, 1.55);
            \draw[optColor, line width=0.6pt, dashed, opacity=0.5] (O) -- (Cb4);
            \draw[optColor, line width=0.9pt] (Cb4) -- (b4);
            
            \coordinate (Cb5) at (-0.5501, 0, 1.35);
            \draw[optColor, line width=0.6pt, dashed, opacity=0.5] (O) -- (Cb5);
            \draw[optColor, line width=0.9pt] (Cb5) -- (b5);
            
            \coordinate (Cb6) at (0.178, 0, 1.51);
            \draw[optColor, line width=0.6pt, dashed, opacity=0.5] (O) -- (Cb6);
            \draw[optColor, line width=0.9pt] (Cb6) -- (b6);
            
            \coordinate (Cb7) at (0.4622, 0, 1.42);
            \draw[optColor, line width=0.6pt, dashed, opacity=0.5] (O) -- (Cb7);
            \draw[optColor, line width=0.9pt] (Cb7) -- (b7);
            
            \coordinate (Cb8) at (-0.1305, 0, 1.42);
            \draw[optColor, line width=0.6pt, dashed, opacity=0.5] (O) -- (Cb8);
            \draw[optColor, line width=0.9pt] (Cb8) -- (b8);
            
            \draw[optColor, line width=0.6pt, dashed] (b1) -- (p1);
            \draw[optColor, line width=0.6pt, dashed] (b2) -- (p2);
            \draw[optColor, line width=0.6pt, dashed] (b3) -- (p3);
            \draw[optColor, line width=0.6pt, dashed] (b4) -- (p4);
            \draw[optColor, line width=0.6pt, dashed] (b5) -- (p5);
            \draw[optColor, line width=0.6pt, dashed] (b6) -- (p6);
            \draw[optColor, line width=0.6pt, dashed] (b7) -- (p7);
            \draw[optColor, line width=0.6pt, dashed] (b8) -- (p8);
            
            \draw[domainColor, line width=1.2pt, fill=white, fill opacity=0.5] 
                (imgFL) -- (imgFR) -- (imgBR) -- (imgBL) -- cycle;
             
            \draw[optColor, line width=0.9pt] (p1) -- (Q1);
            \draw[optColor, line width=0.9pt] (p2) -- (Q2);
            \draw[optColor, line width=0.9pt] (p3) -- (Q3);
            
            \coordinate (Qb4) at (-\objS+1.4263, \objS-2.681, \objZ-\objS-0.92);
            \draw[optColor, line width=0.9pt] (p4) -- (Qb4);
            \draw[optColor, line width=0.9pt, dashed, opacity=0.5] (Qb4) -- (Q4);
            
            \coordinate (Qb8) at (\objS, -\objS-4.805, \objZ-\objS+1.93);
            \draw[optColor, line width=0.9pt] (p8) -- (Qb8);
            \draw[optColor, line width=0.9pt, dashed, opacity=0.5] (Qb8) -- (Q8);
            
            \draw[relaxColor, line width=1.2pt] (Q1) -- (Q2) -- (Q3);
            \draw[relaxColor, line width=1.2pt, dashed, opacity=0.5] (Q3) -- (Q4);
            \draw[relaxColor, line width=1.2pt, dashed, opacity=0.5] (Q4) -- (Q1);
            \draw[relaxColor, line width=1.2pt] (Q5) -- (Q6) -- (Q7) -- (Q8) -- cycle;
            \draw[relaxColor, line width=1.2pt] (Q1) -- (Q5);
            \draw[relaxColor, line width=1.2pt] (Q2) -- (Q6);
            \draw[relaxColor, line width=1.2pt] (Q3) -- (Q7);
            \draw[relaxColor, line width=1.2pt, dashed, opacity=0.5] (Q4) -- (Q8);
            
            \draw[optColor, line width=0.9pt] (p5) -- (Q5);
            \draw[optColor, line width=0.9pt] (p6) -- (Q6);
            \draw[optColor, line width=0.9pt] (p7) -- (Q7);
            
            \foreach \p in {p1, p2, p3, p4, p5, p6, p7, p8} {
                \fill[pathColor] (\p) circle (3.5pt);
            }
            
            \foreach \q in {Q1, Q2, Q3, Q5, Q6, Q7, Q8} {
                \fill[relaxColor] (\q) circle (4.5pt);
            }

            \fill[relaxColor, opacity=0.75] (Q4) circle (4.5pt);
            
            \node[domainColor, font=\normalsize\bfseries] at (1.2, 0, 0.4) {$O$};
            \node[pathColor, font=\normalsize\bfseries, align=center] at (3.8, 0.0, \imgZ-0.1) {$\{\bx_i\}$};
            \node[relaxColor, font=\normalsize\bfseries, align=center] at (4.8, 0, \objZ+0.15) {$\{\bX_i\}$};
            
        \end{scope}
    \end{scope}
    
    \node[font=\footnotesize\bfseries] at (\cellW/2, 0.25) {(c) Absolute Pose Estimation};
    
\end{scope}

%% file: fig/tasks/relpose.tex
\begin{scope}[shift={(3*\stepX, 0)}]
    \fill[optColor!8] (0,0) rectangle (\cellW, \cellH);
    
    \begin{scope}[shift={(2.36, 3.55)}, scale=0.53]
        \tdplotsetmaincoords{78}{5}
        \begin{scope}[tdplot_main_coords]

            \coordinate (O1) at (-3, -3, -2.5);
            \coordinate (c1A) at (-4, -4, 0);
            \coordinate (c1B) at (-4, -4, 2.5);
            \coordinate (c1C) at (-1.5, 2, 0.5);
            \coordinate (c1D) at (-1.5, 2, -2);
            
            \draw[domainColor, line width=1.2pt, fill=white, fill opacity=0.92] 
                (c1A) -- (c1B) -- (c1C) -- (c1D) -- cycle;
            \draw[domainColor!80, line width=0.5pt, opacity=0.6, dashed] (O1) -- (c1A);
            \draw[domainColor!80, line width=0.5pt, opacity=0.6, dashed] (O1) -- (c1B);
            \draw[domainColor!80, line width=0.5pt, opacity=0.6, dashed] (O1) -- (c1C);
            \draw[domainColor!80, line width=0.5pt, opacity=0.6, dashed] (O1) -- (c1D);
            
            \coordinate (p11) at (-3, -1.2, -0.65);
            \coordinate (p12) at (-3.5, -0.3, 0);
            \coordinate (p13) at (-2.8, 0.6, 0.4);
            \coordinate (p14) at (-3.2, -1.0, 1.3);
            \coordinate (p15) at (-2.6, 1.2, -0.8);
            
            \node[domainColor, font=\normalsize\bfseries] at (-2.5, -1.1, 2.7) {$\{\bx_{0i}\}$};
            
            \coordinate (O2) at (2.5, -3, -2.5);
            \coordinate (c2A) at (0, 2, -2);
            \coordinate (c2B) at (0, 2, 0.5);
            \coordinate (c2C) at (3.7, -4, 2.5);
            \coordinate (c2D) at (3.7, -4, 0);
            
            \draw[pathColor, line width=1.2pt, fill=white, fill opacity=0.92] 
                (c2A) -- (c2B) -- (c2C) -- (c2D) -- cycle;
            \draw[pathColor!80, line width=0.5pt, opacity=0.6, dashed] (O2) -- (c2A);
            \draw[pathColor!80, line width=0.5pt, opacity=0.6, dashed] (O2) -- (c2B);
            \draw[pathColor!80, line width=0.5pt, opacity=0.6, dashed] (O2) -- (c2C);
            \draw[pathColor!80, line width=0.5pt, opacity=0.6, dashed] (O2) -- (c2D);
            
            \coordinate (p21) at (2, -1.1, -0.65);
            \coordinate (p22) at (1.5, -0.2, 0);
            \coordinate (p23) at (2.4, 0.7, 0.3);
            \coordinate (p24) at (1.9, -0.9, 1.2);
            \coordinate (p25) at (2.45, 1.3, -0.7);
            
            \node[pathColor, font=\normalsize\bfseries] at (2, -1.1, 2.7) {$\{\bx_{1i}\}$};
            
            \draw[optColor, line width=1pt, dashed] (p11) -- (p21);
            \draw[optColor, line width=1pt, dashed] (p12) -- (p22);
            \draw[optColor, line width=1pt, dashed] (p13) -- (p23);
            \draw[optColor, line width=1pt, dashed] (p14) -- (p24);
            \draw[optColor, line width=1pt, dashed] (p15) -- (p25);

            \fill[domainColor] (p11) circle (2.5pt);
            \fill[domainColor] (p12) circle (2.5pt);
            \fill[domainColor] (p13) circle (2.5pt);
            \fill[domainColor] (p14) circle (2.5pt);
            \fill[domainColor] (p15) circle (2.5pt);

            \fill[pathColor] (p21) circle (2.5pt);
            \fill[pathColor] (p22) circle (2.5pt);
            \fill[pathColor] (p23) circle (2.5pt);
            \fill[pathColor] (p24) circle (2.5pt);
            \fill[pathColor] (p25) circle (2.5pt);

            \fill[domainColor] (O1) circle (4pt);
            \fill[pathColor] (O2) circle (4pt);

            \draw[-{Stealth[length=2mm]}, boundColor, line width=1.5pt] 
                (-3.0, -3.0, -3) to[bend right=40] 
                node[midway, font=\normalsize\bfseries, fill=white, inner sep=3pt, rounded corners] {$(\bR, \bt)$} 
                (2.5, -3.0, -3);

        \end{scope}
    \end{scope}
    
    \node[font=\footnotesize\bfseries] at (\cellW/2, 0.25) {(d) Relative Pose Estimation};
    
\end{scope}

%% file: fig/tasks/reg.tex
\begin{scope}[shift={(4*\stepX, 0)}]
    \fill[boundColor!8] (0,0) rectangle (\cellW, \cellH);
    
    \begin{scope}[shift={(2.36, 3.05)}, scale=2.8]
        \tdplotsetmaincoords{70}{110}
        \begin{scope}[tdplot_main_coords]

            \definecolor{sourcecolor}{RGB}{100,100,255}
            \definecolor{targetcolor}{RGB}{255,100,150}
            \definecolor{matchcolor}{RGB}{100,200,100}
            \definecolor{errorcolor}{RGB}{200,50,50}
            
            \input{fig/tasks/bunny_registration}
            
            \foreach \i in {0,...,999}{
                \pgfmathtruncatemacro{\idx}{\i}
                \fill[sourcecolor,opacity=0.6] (s\idx) ++ (0, 0, 0.5) circle (0.3pt);
            }
            
            \foreach \i in {0,...,999}{
                \pgfmathtruncatemacro{\idx}{\i}
                \fill[targetcolor,opacity=0.5] (t\idx) circle (0.3pt);
            }

            \def\correspondenceIndices{1,102,220,325,426,519,603,704,837,908}
            \foreach \idx in \correspondenceIndices {
                \draw[optColor,line width=1.2pt,opacity=1.0] (s\idx) ++ (0, 0, 0.5) -- (t\idx);
                \fill[sourcecolor] (s\idx) ++ (0, 0, 0.5) circle (0.3pt);
                \fill[targetcolor] (t\idx) circle (0.3pt);
            }
            
            \node[sourcecolor,font=\normalsize\bfseries,anchor=east] at (-0.3,-0.58,0.5) {$\{\bp_i\}$};
            \node[targetcolor,font=\normalsize\bfseries,anchor=west] at (2.5,0.9,0.2) {$\{\bq_i\}$};

        \end{scope}
    \end{scope}
    
    \node[font=\footnotesize\bfseries] at (\cellW/2, 0.25) {(e) 3D Registration};
    
\end{scope}

%% file: fig/tasks/ra.tex
\begin{scope}[shift={(0, \rowOffset)}]
    \fill[gray!8] (0,0) rectangle (\cellW, \cellH);
    
    \begin{scope}[shift={(1.71, 3.3)}, scale=1.43]
        \tdplotsetmaincoords{70}{120}
        \begin{scope}[tdplot_main_coords]
            
            \begin{scope}[shift={(0,0,0.8)}]
            \tdplotsetrotatedcoords{10}{5}{0}
            \begin{scope}[tdplot_rotated_coords]
                \coordinate (apex1) at (0,0,0);
                \coordinate (bl1) at (-0.25,-0.25,-0.4);
                \coordinate (br1) at (0.25,-0.25,-0.4);
                \coordinate (tr1) at (0.25,0.25,-0.4);
                \coordinate (tl1) at (-0.25,0.25,-0.4);
                
                \fill[domainColor,opacity=0.3] (apex1) -- (bl1) -- (br1) -- cycle;
                \fill[domainColor,opacity=0.3] (apex1) -- (br1) -- (tr1) -- cycle;
                \fill[domainColor,opacity=0.3] (apex1) -- (tr1) -- (tl1) -- cycle;
                \fill[domainColor,opacity=0.3] (apex1) -- (tl1) -- (bl1) -- cycle;
                \fill[domainColor,opacity=0.2] (bl1) -- (br1) -- (tr1) -- (tl1) -- cycle;
                
                \draw[domainColor,thick] (apex1) -- (bl1);
                \draw[domainColor,thick] (apex1) -- (br1);
                \draw[domainColor,thick] (apex1) -- (tr1);
                \draw[domainColor,thick] (apex1) -- (tl1);
                \draw[domainColor,thick] (bl1) -- (br1) -- (tr1) -- (tl1) -- cycle;
                
                \draw[-{Stealth[length=1.2mm]},semithick,axisRed] (apex1) -- (0.6,0,0);
                \draw[-{Stealth[length=1.2mm]},semithick,axisGreen] (apex1) -- (0,0.3,0);
                \draw[-{Stealth[length=1.2mm]},semithick,axisBlue] (apex1) -- (0,0,-0.4);
                
                \node[domainColor,font=\normalsize] at (0.1,0.1,-0.65) {$\bR_1$};
            \end{scope}
            \end{scope}
            \coordinate (c1) at (0,0,0.8);
            
            \begin{scope}[shift={(1.8,0,0.9)}]
            \tdplotsetrotatedcoords{20}{10}{15}
            \begin{scope}[tdplot_rotated_coords]
                \coordinate (apex2) at (0,0,0);
                \coordinate (bl2) at (-0.25,-0.25,-0.4);
                \coordinate (br2) at (0.25,-0.25,-0.4);
                \coordinate (tr2) at (0.25,0.25,-0.4);
                \coordinate (tl2) at (-0.25,0.25,-0.4);
                
                \fill[domainColor,opacity=0.3] (apex2) -- (bl2) -- (br2) -- cycle;
                \fill[domainColor,opacity=0.3] (apex2) -- (br2) -- (tr2) -- cycle;
                \fill[domainColor,opacity=0.3] (apex2) -- (tr2) -- (tl2) -- cycle;
                \fill[domainColor,opacity=0.3] (apex2) -- (tl2) -- (bl2) -- cycle;
                \fill[domainColor,opacity=0.2] (bl2) -- (br2) -- (tr2) -- (tl2) -- cycle;
                
                \draw[domainColor,thick] (apex2) -- (bl2);
                \draw[domainColor,thick] (apex2) -- (br2);
                \draw[domainColor,thick] (apex2) -- (tr2);
                \draw[domainColor,thick] (apex2) -- (tl2);
                \draw[domainColor,thick] (bl2) -- (br2) -- (tr2) -- (tl2) -- cycle;
                
                \draw[-{Stealth[length=1.2mm]},semithick,axisRed] (apex2) -- (0.5,0,0);
                \draw[-{Stealth[length=1.2mm]},semithick,axisGreen] (apex2) -- (0,0.4,0);
                \draw[-{Stealth[length=1.2mm]},semithick,axisBlue] (apex2) -- (0,0,-0.4);
                
                \node[domainColor,font=\normalsize] at (0,0.05,-0.75) {$\bR_2$};
            \end{scope}
            \end{scope}
            \coordinate (c2) at (1.8,0,0.9);
            
            \begin{scope}[shift={(2.5,1.3,0.2)}]
            \tdplotsetrotatedcoords{10}{10}{-5}
            \begin{scope}[tdplot_rotated_coords]
                \coordinate (apex3) at (0,0,0);
                \coordinate (bl3) at (-0.25,-0.25,-0.4);
                \coordinate (br3) at (0.25,-0.25,-0.4);
                \coordinate (tr3) at (0.25,0.25,-0.4);
                \coordinate (tl3) at (-0.25,0.25,-0.4);
                
                \fill[domainColor,opacity=0.3] (apex3) -- (bl3) -- (br3) -- cycle;
                \fill[domainColor,opacity=0.3] (apex3) -- (br3) -- (tr3) -- cycle;
                \fill[domainColor,opacity=0.3] (apex3) -- (tr3) -- (tl3) -- cycle;
                \fill[domainColor,opacity=0.3] (apex3) -- (tl3) -- (bl3) -- cycle;
                \fill[domainColor,opacity=0.2] (bl3) -- (br3) -- (tr3) -- (tl3) -- cycle;
                
                \draw[domainColor,thick] (apex3) -- (bl3);
                \draw[domainColor,thick] (apex3) -- (br3);
                \draw[domainColor,thick] (apex3) -- (tr3);
                \draw[domainColor,thick] (apex3) -- (tl3);
                \draw[domainColor,thick] (bl3) -- (br3) -- (tr3) -- (tl3) -- cycle;
                
                \draw[-{Stealth[length=1.2mm]},semithick,axisRed] (apex3) -- (0.6,0,0);
                \draw[-{Stealth[length=1.2mm]},semithick,axisGreen] (apex3) -- (0,0.4,0);
                \draw[-{Stealth[length=1.2mm]},semithick,axisBlue] (apex3) -- (0,0,-0.4);
                
                \node[domainColor,font=\normalsize] at (-0.4,-0.15,-1.05) {$\bR_3$};
            \end{scope}
            \end{scope}
            \coordinate (c3) at (2.5,1.3,0.2);
            
            \begin{scope}[shift={(1.2,2.3,0.25)}]
            \tdplotsetrotatedcoords{100}{20}{15}
            \begin{scope}[tdplot_rotated_coords]
                \coordinate (apex4) at (0,0,0);
                \coordinate (bl4) at (-0.25,-0.25,-0.4);
                \coordinate (br4) at (0.25,-0.25,-0.4);
                \coordinate (tr4) at (0.25,0.25,-0.4);
                \coordinate (tl4) at (-0.25,0.25,-0.4);
                
                \fill[domainColor,opacity=0.3] (apex4) -- (bl4) -- (br4) -- cycle;
                \fill[domainColor,opacity=0.3] (apex4) -- (br4) -- (tr4) -- cycle;
                \fill[domainColor,opacity=0.3] (apex4) -- (tr4) -- (tl4) -- cycle;
                \fill[domainColor,opacity=0.3] (apex4) -- (tl4) -- (bl4) -- cycle;
                \fill[domainColor,opacity=0.2] (bl4) -- (br4) -- (tr4) -- (tl4) -- cycle;
                
                \draw[domainColor,thick] (apex4) -- (bl4);
                \draw[domainColor,thick] (apex4) -- (br4);
                \draw[domainColor,thick] (apex4) -- (tr4);
                \draw[domainColor,thick] (apex4) -- (tl4);
                \draw[domainColor,thick] (bl4) -- (br4) -- (tr4) -- (tl4) -- cycle;
                
                \draw[-{Stealth[length=1.2mm]},semithick,axisRed] (apex4) -- (0.25,0,0);
                \draw[-{Stealth[length=1.2mm]},semithick,axisGreen] (apex4) -- (0,0.5,0);
                \draw[-{Stealth[length=1.2mm]},semithick,axisBlue] (apex4) -- (0,0,-0.4);
                
                \node[domainColor,font=\normalsize] at (0.1,0.1,-0.75) {$\bR_4$};
            \end{scope}
            \end{scope}
            \coordinate (c4) at (1.2,2.3,0.25);
            
            \begin{scope}[shift={(0,1.3,1.3)}]
            \tdplotsetrotatedcoords{145}{-5}{-5}
            \begin{scope}[tdplot_rotated_coords]
                \coordinate (apex5) at (0,0,0);
                \coordinate (bl5) at (-0.25,-0.25,-0.4);
                \coordinate (br5) at (0.25,-0.25,-0.4);
                \coordinate (tr5) at (0.25,0.25,-0.4);
                \coordinate (tl5) at (-0.25,0.25,-0.4);
                
                \fill[domainColor,opacity=0.3] (apex5) -- (bl5) -- (br5) -- cycle;
                \fill[domainColor,opacity=0.3] (apex5) -- (br5) -- (tr5) -- cycle;
                \fill[domainColor,opacity=0.3] (apex5) -- (tr5) -- (tl5) -- cycle;
                \fill[domainColor,opacity=0.3] (apex5) -- (tl5) -- (bl5) -- cycle;
                \fill[domainColor,opacity=0.2] (bl5) -- (br5) -- (tr5) -- (tl5) -- cycle;
                
                \draw[domainColor,thick] (apex5) -- (bl5);
                \draw[domainColor,thick] (apex5) -- (br5);
                \draw[domainColor,thick] (apex5) -- (tr5);
                \draw[domainColor,thick] (apex5) -- (tl5);
                \draw[domainColor,thick] (bl5) -- (br5) -- (tr5) -- (tl5) -- cycle;
                
                \draw[-{Stealth[length=1.2mm]},semithick,axisRed] (apex5) -- (0.4,0,0);
                \draw[-{Stealth[length=1.2mm]},semithick,axisGreen] (apex5) -- (0,0.6,0);
                \draw[-{Stealth[length=1.2mm]},semithick,axisBlue] (apex5) -- (0,0,-0.4);
                
                \node[domainColor,font=\normalsize] at (-0.45,0,-0.38) {$\bR_5$};
            \end{scope}
            \end{scope}
            \coordinate (c5) at (0,1.3,1.3);
            
            \begin{scope}
                \draw[relaxColor, semithick, opacity=0.9, -{Stealth[length=1.2mm]}] 
                    (c1) to[bend right=13] node[pos=0.5, font=\small, fill=white, rounded corners=1.5pt, inner sep=0.5pt, shift={(0cm, -0.05cm)}] {$\bR_{12}$} (c2);
                \draw[relaxColor, semithick, opacity=0.9, -{Stealth[length=1.2mm]}] 
                    (c2) to[bend left=30] node[pos=0.6, font=\small, fill=white, rounded corners=1.5pt, inner sep=0.5pt] {$\bR_{23}$} (c3);
                \draw[relaxColor, semithick, opacity=0.9, -{Stealth[length=1.2mm]}] 
                    (c3) to[bend left=15] node[pos=0.5, font=\small, fill=white, rounded corners=1.5pt, inner sep=0.5pt] {$\bR_{34}$} (c4);
                \draw[relaxColor, semithick, opacity=0.9, -{Stealth[length=1.2mm]}] 
                    (c4) to[bend right=55] node[pos=0.5, below, font=\small, fill=white, rounded corners=1.5pt, inner sep=0.5pt, shift={(-0.15cm, 0cm)}] {$\bR_{45}$} (c5);
                \draw[relaxColor, semithick, opacity=0.9, -{Stealth[length=1.2mm]}] 
                    (c5) to[bend right=15] node[pos=0.5, font=\small, fill=white, rounded corners=1.5pt, inner sep=0.5pt] {$\bR_{51}$} (c1);
                
                \draw[relaxColor, semithick, opacity=0.9, -{Stealth[length=1.2mm]}] 
                    (c1) to[bend left=10] node[pos=0.65, above, font=\small, fill=white,rounded corners=1.5pt, inner sep=0.5pt, shift={(-0.1cm, 0cm)}] {$\bR_{14}$} (c4);
            \end{scope}
            
        \end{scope}
    \end{scope}
    
    \node[font=\footnotesize\bfseries] at (\cellW/2, 0.25) {(f) Rotation Averaging};
    
\end{scope}

%% file: fig/tasks/ta.tex
\begin{scope}[shift={(\stepX, \rowOffset)}]
    \fill[relaxColor!8] (0,0) rectangle (\cellW, \cellH);
    
    \begin{scope}[shift={(1.8, 3.5)}, scale=1.15]
        \tdplotsetmaincoords{70}{120}
        \begin{scope}[tdplot_main_coords]
        
            \begin{scope}[shift={(0,-0.3,1.5)}]
            \tdplotsetrotatedcoords{10}{5}{0}
            \begin{scope}[tdplot_rotated_coords,scale=0.8]
                \coordinate (apex1) at (0,0,0);
                \coordinate (bl1) at (-0.25,-0.25,-0.4);
                \coordinate (br1) at (0.25,-0.25,-0.4);
                \coordinate (tr1) at (0.25,0.25,-0.4);
                \coordinate (tl1) at (-0.25,0.25,-0.4);
                
                \fill[domainColor,opacity=0.25] (apex1) -- (bl1) -- (br1) -- cycle;
                \fill[domainColor,opacity=0.25] (apex1) -- (br1) -- (tr1) -- cycle;
                \fill[domainColor,opacity=0.25] (apex1) -- (tr1) -- (tl1) -- cycle;
                \fill[domainColor,opacity=0.25] (apex1) -- (tl1) -- (bl1) -- cycle;
                
                \draw[domainColor,thick,opacity=0.6] (apex1) -- (bl1);
                \draw[domainColor,thick,opacity=0.6] (apex1) -- (br1);
                \draw[domainColor,thick,opacity=0.6] (apex1) -- (tr1);
                \draw[domainColor,thick,opacity=0.6] (apex1) -- (tl1);
                \draw[domainColor,thick,opacity=0.6] (bl1) -- (br1) -- (tr1) -- (tl1) -- cycle;
            \end{scope}
            \end{scope}
            \coordinate (c1) at (0,-0.3,1.5);
            
            \begin{scope}[shift={(1.8,0,-0.1)}]
            \tdplotsetrotatedcoords{20}{10}{15}
            \begin{scope}[tdplot_rotated_coords,scale=0.8]
                \coordinate (apex2) at (0,0,0);
                \coordinate (bl2) at (-0.25,-0.25,-0.4);
                \coordinate (br2) at (0.25,-0.25,-0.4);
                \coordinate (tr2) at (0.25,0.25,-0.4);
                \coordinate (tl2) at (-0.25,0.25,-0.4);
                
                \fill[domainColor,opacity=0.25] (apex2) -- (bl2) -- (br2) -- cycle;
                \fill[domainColor,opacity=0.25] (apex2) -- (br2) -- (tr2) -- cycle;
                \fill[domainColor,opacity=0.25] (apex2) -- (tr2) -- (tl2) -- cycle;
                \fill[domainColor,opacity=0.25] (apex2) -- (tl2) -- (bl2) -- cycle;
                
                \draw[domainColor,thick,opacity=0.6] (apex2) -- (bl2);
                \draw[domainColor,thick,opacity=0.6] (apex2) -- (br2);
                \draw[domainColor,thick,opacity=0.6] (apex2) -- (tr2);
                \draw[domainColor,thick,opacity=0.6] (apex2) -- (tl2);
                \draw[domainColor,thick,opacity=0.6] (bl2) -- (br2) -- (tr2) -- (tl2) -- cycle;
            \end{scope}
            \end{scope}
            \coordinate (c2) at (1.8,0,-0.1);
            
            \begin{scope}[shift={(2.5,2.1,-1)}]
            \tdplotsetrotatedcoords{10}{10}{-5}
            \begin{scope}[tdplot_rotated_coords,scale=0.8]
                \coordinate (apex3) at (0,0,0);
                \coordinate (bl3) at (-0.25,-0.25,-0.4);
                \coordinate (br3) at (0.25,-0.25,-0.4);
                \coordinate (tr3) at (0.25,0.25,-0.4);
                \coordinate (tl3) at (-0.25,0.25,-0.4);
                
                \fill[domainColor,opacity=0.25] (apex3) -- (bl3) -- (br3) -- cycle;
                \fill[domainColor,opacity=0.25] (apex3) -- (br3) -- (tr3) -- cycle;
                \fill[domainColor,opacity=0.25] (apex3) -- (tr3) -- (tl3) -- cycle;
                \fill[domainColor,opacity=0.25] (apex3) -- (tl3) -- (bl3) -- cycle;
                
                \draw[domainColor,thick,opacity=0.6] (apex3) -- (bl3);
                \draw[domainColor,thick,opacity=0.6] (apex3) -- (br3);
                \draw[domainColor,thick,opacity=0.6] (apex3) -- (tr3);
                \draw[domainColor,thick,opacity=0.6] (apex3) -- (tl3);
                \draw[domainColor,thick,opacity=0.6] (bl3) -- (br3) -- (tr3) -- (tl3) -- cycle;
            \end{scope}
            \end{scope}
            \coordinate (c3) at (2.5,2.1,-1);
            
            \begin{scope}[shift={(1.2,2.3,0.1)}]
            \tdplotsetrotatedcoords{100}{20}{15}
            \begin{scope}[tdplot_rotated_coords,scale=0.8]
                \coordinate (apex4) at (0,0,0);
                \coordinate (bl4) at (-0.25,-0.25,-0.4);
                \coordinate (br4) at (0.25,-0.25,-0.4);
                \coordinate (tr4) at (0.25,0.25,-0.4);
                \coordinate (tl4) at (-0.25,0.25,-0.4);
                
                \fill[domainColor,opacity=0.25] (apex4) -- (bl4) -- (br4) -- cycle;
                \fill[domainColor,opacity=0.25] (apex4) -- (br4) -- (tr4) -- cycle;
                \fill[domainColor,opacity=0.25] (apex4) -- (tr4) -- (tl4) -- cycle;
                \fill[domainColor,opacity=0.25] (apex4) -- (tl4) -- (bl4) -- cycle;
                
                \draw[domainColor,thick,opacity=0.6] (apex4) -- (bl4);
                \draw[domainColor,thick,opacity=0.6] (apex4) -- (br4);
                \draw[domainColor,thick,opacity=0.6] (apex4) -- (tr4);
                \draw[domainColor,thick,opacity=0.6] (apex4) -- (tl4);
                \draw[domainColor,thick,opacity=0.6] (bl4) -- (br4) -- (tr4) -- (tl4) -- cycle;
            \end{scope}
            \end{scope}
            \coordinate (c4) at (1.2,2.3,0.1);
            
            \begin{scope}[shift={(0,1.5,1.2)}]
            \tdplotsetrotatedcoords{145}{-5}{-5}
            \begin{scope}[tdplot_rotated_coords,scale=0.8]
                \coordinate (apex5) at (0,0,0);
                \coordinate (bl5) at (-0.25,-0.25,-0.4);
                \coordinate (br5) at (0.25,-0.25,-0.4);
                \coordinate (tr5) at (0.25,0.25,-0.4);
                \coordinate (tl5) at (-0.25,0.25,-0.4);
                
                \fill[domainColor,opacity=0.25] (apex5) -- (bl5) -- (br5) -- cycle;
                \fill[domainColor,opacity=0.25] (apex5) -- (br5) -- (tr5) -- cycle;
                \fill[domainColor,opacity=0.25] (apex5) -- (tr5) -- (tl5) -- cycle;
                \fill[domainColor,opacity=0.25] (apex5) -- (tl5) -- (bl5) -- cycle;
                
                \draw[domainColor,thick,opacity=0.6] (apex5) -- (bl5);
                \draw[domainColor,thick,opacity=0.6] (apex5) -- (br5);
                \draw[domainColor,thick,opacity=0.6] (apex5) -- (tr5);
                \draw[domainColor,thick,opacity=0.6] (apex5) -- (tl5);
                \draw[domainColor,thick,opacity=0.6] (bl5) -- (br5) -- (tr5) -- (tl5) -- cycle;
            \end{scope}
            \end{scope}
            \coordinate (c5) at (0,1.5,1.2);
            
            \begin{scope}
                \draw[-{Stealth[length=2mm]},pathColor,very thick,opacity=0.85] 
                    (c1) -- ($(c1)!0.7!(c2)$) 
                    node[pos=0.55,font=\normalsize,fill=white,rounded corners,inner sep=1pt] {$\mathbf{t}_{12}$};
                
                \draw[-{Stealth[length=2mm]},pathColor,very thick,opacity=0.85] 
                    (c2) -- ($(c2)!0.65!(c3)$)
                    node[pos=0.5,font=\normalsize,fill=white,rounded corners,inner sep=1pt,shift={(0.1cm,-0.05cm)}] {$\mathbf{t}_{23}$};
                
                \draw[-{Stealth[length=2mm]},pathColor,very thick,opacity=0.85] 
                    (c3) -- ($(c3)!0.7!(c4)$)
                    node[pos=0.5,font=\normalsize,fill=white,rounded corners,inner sep=1pt] {$\mathbf{t}_{34}$};
                
                \draw[-{Stealth[length=2mm]},pathColor,very thick,opacity=0.85] 
                    (c4) -- ($(c4)!0.55!(c5)$)
                    node[pos=0.45,font=\normalsize,fill=white,rounded corners,inner sep=1pt] {$\mathbf{t}_{45}$};
                
                \draw[-{Stealth[length=2mm]},pathColor,very thick,opacity=0.85] 
                    (c5) -- ($(c5)!0.65!(c1)$)
                    node[pos=0.5,font=\normalsize,fill=white,rounded corners,inner sep=1pt] {$\mathbf{t}_{51}$};
                
                \draw[-{Stealth[length=2mm]},pathColor,very thick,opacity=0.85] 
                    (c1) -- ($(c1)!0.55!(c4)$)
                    node[pos=0.45,font=\normalsize,fill=white,rounded corners,inner sep=1pt,shift={(0.1cm,-0.15cm)}] {$\mathbf{t}_{14}$};
                
                \draw[-{Stealth[length=2mm]},pathColor,very thick,opacity=0.85] 
                    (c2) -- ($(c2)!0.5!(c5)$)
                    node[pos=0.4,font=\normalsize,fill=white,rounded corners,inner sep=1pt,shift={(0.1cm,0cm)}] {$\mathbf{t}_{25}$};
            \end{scope}
            
            \fill[domainColor] (c1) circle (3pt);
            \node[domainColor,font=\normalsize,shift={(-0.4cm,0.35cm)}] at ($(c1)+(0,0,-0.3)$) {$\mathbf{t}_1$};
            \fill[domainColor] (c2) circle (3pt);
            \node[domainColor,font=\normalsize,shift={(-0.4cm,0.5cm)}] at ($(c2)+(0,0,-0.3)$) {$\mathbf{t}_2$};
            \fill[domainColor] (c3) circle (3pt);
            \node[domainColor,font=\normalsize,shift={(-0.35cm,0.4cm)}] at ($(c3)+(0.1,0,-0.3)$) {$\mathbf{t}_3$};
            \fill[domainColor] (c4) circle (3pt);
            \node[domainColor,font=\normalsize,shift={(0.45cm,-0.3cm)}] at ($(c4)+(0,0,0.3)$) {$\mathbf{t}_4$};
            \fill[domainColor] (c5) circle (3pt);
            \node[domainColor,font=\normalsize,shift={(0.35cm,-0.05cm)}] at ($(c5)+(-0.1,0,0.1)$) {$\mathbf{t}_5$};

        \end{scope}
    \end{scope}
    
    \node[font=\footnotesize\bfseries] at (\cellW/2, 0.25) {(g) Translation Averaging};
\end{scope}

%% file: fig/tasks/tri.tex
\begin{scope}[shift={(2*\stepX, \rowOffset)}]
    \fill[pathColor!8] (0,0) rectangle (\cellW, \cellH);
    
    \begin{scope}[shift={(1.975, 2.32)}, scale=0.52]
        \tdplotsetmaincoords{78}{5}
        \begin{scope}[tdplot_main_coords]          
        
            \coordinate (X) at (0.2, -0.2, 5.5);
            
            \coordinate (O1) at (-2.5, -3.5, -1.3);
            \coordinate (c1A) at (-2.7, -4, 0);
            \coordinate (c1B) at (-2.7, -4, 2.5);
            \coordinate (c1C) at (-1.5, 2, 0.5);
            \coordinate (c1D) at (-1.5, 2, -2);
            
            \draw[domainColor, line width=1.2pt, fill=white, fill opacity=0.92] 
                (c1A) -- (c1B) -- (c1C) -- (c1D) -- cycle;
            \draw[domainColor!80, line width=0.5pt, opacity=0.6, dashed] (O1) -- (c1A);
            \draw[domainColor!80, line width=0.5pt, opacity=0.6, dashed] (O1) -- (c1B);
            \draw[domainColor!80, line width=0.5pt, opacity=0.6, dashed] (O1) -- (c1C);
            \draw[domainColor!80, line width=0.5pt, opacity=0.6, dashed] (O1) -- (c1D);
            
            \coordinate (p11) at (-1.83, -0.3, 0.4);
            
            \coordinate (O2) at (0.5, -3, -1.8);
            \coordinate (c2A) at (-0.9, 2, -2);
            \coordinate (c2B) at (-0.9, 2, 0.5);
            \coordinate (c2C) at (1.4, -4, 2.1);
            \coordinate (c2D) at (1.4, -4, -0.5);
            
            \draw[domainColor, line width=1.2pt, fill=white, fill opacity=0.92] 
                (c2A) -- (c2B) -- (c2C) -- (c2D) -- cycle;
            \draw[domainColor!80, line width=0.5pt, opacity=0.6, dashed] (O2) -- (c2A);
            \draw[domainColor!80, line width=0.5pt, opacity=0.6, dashed] (O2) -- (c2B);
            \draw[domainColor!80, line width=0.5pt, opacity=0.6, dashed] (O2) -- (c2C);
            \draw[domainColor!80, line width=0.5pt, opacity=0.6, dashed] (O2) -- (c2D);
            
            \coordinate (p21) at (0.24, -0.2, -0.2);
            
            \coordinate (O3) at (3.8, -3, -1.3);
            \coordinate (c3A) at (1.5, 2, -1.6);
            \coordinate (c3B) at (1.5, 2, 0.8);
            \coordinate (c3C) at (3.5, -4, 3);
            \coordinate (c3D) at (3.5, -4, 0.5);
            
            \draw[domainColor, line width=1.2pt, fill=white, fill opacity=0.92] 
                (c3A) -- (c3B) -- (c3C) -- (c3D) -- cycle;
            \draw[domainColor!80, line width=0.5pt, opacity=0.6, dashed] (O3) -- (c3A);
            \draw[domainColor!80, line width=0.5pt, opacity=0.6, dashed] (O3) -- (c3B);
            \draw[domainColor!80, line width=0.5pt, opacity=0.6, dashed] (O3) -- (c3C);
            \draw[domainColor!80, line width=0.5pt, opacity=0.6, dashed] (O3) -- (c3D);
            
            \coordinate (p31) at (2.462, -0.2, 0.5);
            
            \draw[optColor, line width=1pt, dashed] (p11) -- (p21);
            \draw[optColor, line width=1pt, dashed] (p21) -- (p31);
            
            \draw[gray, line width=0.6pt, opacity=0.7] (O1) -- (X);
            \draw[gray, line width=0.6pt, opacity=0.7] (O2) -- (X);
            \draw[gray, line width=0.6pt, opacity=0.7] (O3) -- (X);
            
            \fill[observColor] (p11) circle (2.5pt);
            \node[observColor, font=\normalsize, above] at (-2.25, -0.3, 0.2) {$\mathbf{x}_1$};
            
            \fill[observColor] (p21) circle (2.5pt);
            \node[observColor, font=\normalsize, above, shift={(0cm, 0.01cm)}] at (-0.1, -0.2, -0.2) {$\mathbf{x}_2$};
            
            \fill[observColor] (p31) circle (2.5pt);
            \node[observColor, font=\normalsize, above right, shift={(-0.05cm, 0.01cm)}] at (2.3, -0.2, 0.4) {$\mathbf{x}_3$};
            
            \fill[domainColor] (O1) circle (3.5pt);
            \node[domainColor, font=\normalsize\bfseries, below, shift={(0.07cm, 0cm)}] at (O1) {$(\mathbf{R}_1, \mathbf{t}_1)$};
            
            \fill[domainColor] (O2) circle (3.5pt);
            \node[domainColor, font=\normalsize\bfseries, below] at (O2) {$(\mathbf{R}_2, \mathbf{t}_2)$};
            
            \fill[domainColor] (O3) circle (3.5pt);
            \node[domainColor, font=\normalsize\bfseries, below, shift={(-0.2cm, 0cm)}] at (O3) {$(\mathbf{R}_3, \mathbf{t}_3)$};
            
            \fill[boundColor] (X) circle (4.5pt);
            \node[boundColor, font=\normalsize\bfseries, above] at (X) {$\mathbf{X}$};
            
        \end{scope}
    \end{scope}
    
    \node[font=\footnotesize\bfseries] at (\cellW/2, 0.25) {(h) Triangulation};
    
\end{scope}

%% file: fig/tasks/pgo.tex
\begin{scope}[shift={(3*\stepX, \rowOffset)}]
    \fill[optColor!8] (0,0) rectangle (\cellW, \cellH);
    
    \begin{scope}[shift={(1.53, 2.85)}, scale=1.4]
        \tdplotsetmaincoords{70}{120}
        \begin{scope}[tdplot_main_coords]

            \begin{scope}[shift={(0,0.6,0.9)}]
            \tdplotsetrotatedcoords{10}{5}{0}
            \begin{scope}[tdplot_rotated_coords,scale=0.8]
                \coordinate (apex1) at (0,0,0);
                \coordinate (bl1) at (-0.25,-0.25,-0.4);
                \coordinate (br1) at (0.25,-0.25,-0.4);
                \coordinate (tr1) at (0.25,0.25,-0.4);
                \coordinate (tl1) at (-0.25,0.25,-0.4);
                
                \fill[domainColor,opacity=0.3] (apex1) -- (bl1) -- (br1) -- cycle;
                \fill[domainColor,opacity=0.3] (apex1) -- (br1) -- (tr1) -- cycle;
                \fill[domainColor,opacity=0.3] (apex1) -- (tr1) -- (tl1) -- cycle;
                \fill[domainColor,opacity=0.3] (apex1) -- (tl1) -- (bl1) -- cycle;
                \fill[domainColor,opacity=0.2] (bl1) -- (br1) -- (tr1) -- (tl1) -- cycle;
                
                \draw[domainColor,thick] (apex1) -- (bl1);
                \draw[domainColor,thick] (apex1) -- (br1);
                \draw[domainColor,thick] (apex1) -- (tr1);
                \draw[domainColor,thick] (apex1) -- (tl1);
                \draw[domainColor,thick] (bl1) -- (br1) -- (tr1) -- (tl1) -- cycle;
                
                \draw[-{Stealth[length=1.2mm]},semithick,axisRed] (apex1) -- (0.6,0,0);
                \draw[-{Stealth[length=1.2mm]},semithick,axisGreen] (apex1) -- (0,0.3,0);
                \draw[-{Stealth[length=1.2mm]},semithick,axisBlue] (apex1) -- (0,0,-0.4);
                
                \node[domainColor,font=\normalsize] at (0.1,0.1,-0.75) {$\mathbf{T}_5$};
            \end{scope}
            \end{scope}
            \coordinate (c5) at (0,0.6,0.9);
            
            \begin{scope}[shift={(1.8,0.1,1.4)}]
            \tdplotsetrotatedcoords{25}{10}{15}
            \begin{scope}[tdplot_rotated_coords,scale=0.8]
                \coordinate (apex2) at (0,0,0);
                \coordinate (bl2) at (-0.25,-0.25,-0.4);
                \coordinate (br2) at (0.25,-0.25,-0.4);
                \coordinate (tr2) at (0.25,0.25,-0.4);
                \coordinate (tl2) at (-0.25,0.25,-0.4);
                
                \fill[domainColor,opacity=0.3] (apex2) -- (bl2) -- (br2) -- cycle;
                \fill[domainColor,opacity=0.3] (apex2) -- (br2) -- (tr2) -- cycle;
                \fill[domainColor,opacity=0.3] (apex2) -- (tr2) -- (tl2) -- cycle;
                \fill[domainColor,opacity=0.3] (apex2) -- (tl2) -- (bl2) -- cycle;
                \fill[domainColor,opacity=0.2] (bl2) -- (br2) -- (tr2) -- (tl2) -- cycle;
                
                \draw[domainColor,thick] (apex2) -- (bl2);
                \draw[domainColor,thick] (apex2) -- (br2);
                \draw[domainColor,thick] (apex2) -- (tr2);
                \draw[domainColor,thick] (apex2) -- (tl2);
                \draw[domainColor,thick] (bl2) -- (br2) -- (tr2) -- (tl2) -- cycle;
                
                \draw[-{Stealth[length=1.2mm]},semithick,axisRed] (apex2) -- (0.5,0,0);
                \draw[-{Stealth[length=1.2mm]},semithick,axisGreen] (apex2) -- (0,0.4,0);
                \draw[-{Stealth[length=1.2mm]},semithick,axisBlue] (apex2) -- (0,0,-0.4);
                
                \node[domainColor,font=\normalsize] at (0,0.05,-0.85) {$\mathbf{T}_4$};
            \end{scope}
            \end{scope}
            \coordinate (c4) at (1.8,0.1,1.4);
            
            \begin{scope}[shift={(2.5,1.3,-0.05)}]
            \tdplotsetrotatedcoords{50}{10}{-5}
            \begin{scope}[tdplot_rotated_coords,scale=0.8]
                \coordinate (apex3) at (0,0,0);
                \coordinate (bl3) at (-0.25,-0.25,-0.4);
                \coordinate (br3) at (0.25,-0.25,-0.4);
                \coordinate (tr3) at (0.25,0.25,-0.4);
                \coordinate (tl3) at (-0.25,0.25,-0.4);
                
                \fill[domainColor,opacity=0.3] (apex3) -- (bl3) -- (br3) -- cycle;
                \fill[domainColor,opacity=0.3] (apex3) -- (br3) -- (tr3) -- cycle;
                \fill[domainColor,opacity=0.3] (apex3) -- (tr3) -- (tl3) -- cycle;
                \fill[domainColor,opacity=0.3] (apex3) -- (tl3) -- (bl3) -- cycle;
                \fill[domainColor,opacity=0.2] (bl3) -- (br3) -- (tr3) -- (tl3) -- cycle;
                
                \draw[domainColor,thick] (apex3) -- (bl3);
                \draw[domainColor,thick] (apex3) -- (br3);
                \draw[domainColor,thick] (apex3) -- (tr3);
                \draw[domainColor,thick] (apex3) -- (tl3);
                \draw[domainColor,thick] (bl3) -- (br3) -- (tr3) -- (tl3) -- cycle;
                
                \draw[-{Stealth[length=1.2mm]},semithick,axisRed] (apex3) -- (0.6,0,0);
                \draw[-{Stealth[length=1.2mm]},semithick,axisGreen] (apex3) -- (0,0.4,0);
                \draw[-{Stealth[length=1.2mm]},semithick,axisBlue] (apex3) -- (0,0,-0.4);
                
                \node[domainColor,font=\normalsize] at (-1.5,-0.2,-1.1) {$\mathbf{T}_3$};
            \end{scope}
            \end{scope}
            \coordinate (c3) at (2.5,1.3,-0.05);
            
            \begin{scope}[shift={(1.2,2.3,0)}]
            \tdplotsetrotatedcoords{100}{20}{15}
            \begin{scope}[tdplot_rotated_coords,scale=0.8]
                \coordinate (apex4) at (0,0,0);
                \coordinate (bl4) at (-0.25,-0.25,-0.4);
                \coordinate (br4) at (0.25,-0.25,-0.4);
                \coordinate (tr4) at (0.25,0.25,-0.4);
                \coordinate (tl4) at (-0.25,0.25,-0.4);
                
                \fill[domainColor,opacity=0.3] (apex4) -- (bl4) -- (br4) -- cycle;
                \fill[domainColor,opacity=0.3] (apex4) -- (br4) -- (tr4) -- cycle;
                \fill[domainColor,opacity=0.3] (apex4) -- (tr4) -- (tl4) -- cycle;
                \fill[domainColor,opacity=0.3] (apex4) -- (tl4) -- (bl4) -- cycle;
                \fill[domainColor,opacity=0.2] (bl4) -- (br4) -- (tr4) -- (tl4) -- cycle;
                
                \draw[domainColor,thick] (apex4) -- (bl4);
                \draw[domainColor,thick] (apex4) -- (br4);
                \draw[domainColor,thick] (apex4) -- (tr4);
                \draw[domainColor,thick] (apex4) -- (tl4);
                \draw[domainColor,thick] (bl4) -- (br4) -- (tr4) -- (tl4) -- cycle;
                
                \draw[-{Stealth[length=1.2mm]},semithick,axisRed] (apex4) -- (0.25,0,0);
                \draw[-{Stealth[length=1.2mm]},semithick,axisGreen] (apex4) -- (0,0.7,0);
                \draw[-{Stealth[length=1.2mm]},semithick,axisBlue] (apex4) -- (0,0,-0.4);
                
                \node[domainColor,font=\normalsize] at (0.15,0.1,-0.85) {$\mathbf{T}_2$};
            \end{scope}
            \end{scope}
            \coordinate (c2) at (1.2,2.3,0);
            
            \begin{scope}[shift={(0,1.3,1.8)}]
            \tdplotsetrotatedcoords{145}{-5}{-5}
            \begin{scope}[tdplot_rotated_coords,scale=0.8]
                \coordinate (apex5) at (0,0,0);
                \coordinate (bl5) at (-0.25,-0.25,-0.4);
                \coordinate (br5) at (0.25,-0.25,-0.4);
                \coordinate (tr5) at (0.25,0.25,-0.4);
                \coordinate (tl5) at (-0.25,0.25,-0.4);
                
                \fill[domainColor,opacity=0.3] (apex5) -- (bl5) -- (br5) -- cycle;
                \fill[domainColor,opacity=0.3] (apex5) -- (br5) -- (tr5) -- cycle;
                \fill[domainColor,opacity=0.3] (apex5) -- (tr5) -- (tl5) -- cycle;
                \fill[domainColor,opacity=0.3] (apex5) -- (tl5) -- (bl5) -- cycle;
                \fill[domainColor,opacity=0.2] (bl5) -- (br5) -- (tr5) -- (tl5) -- cycle;
                
                \draw[domainColor,thick] (apex5) -- (bl5);
                \draw[domainColor,thick] (apex5) -- (br5);
                \draw[domainColor,thick] (apex5) -- (tr5);
                \draw[domainColor,thick] (apex5) -- (tl5);
                \draw[domainColor,thick] (bl5) -- (br5) -- (tr5) -- (tl5) -- cycle;
                
                \draw[-{Stealth[length=1.2mm]},semithick,axisRed] (apex5) -- (0.4,0,0);
                \draw[-{Stealth[length=1.2mm]},semithick,axisGreen] (apex5) -- (0,0.7,0);
                \draw[-{Stealth[length=1.2mm]},semithick,axisBlue] (apex5) -- (0,0,-0.4);
                
                \node[domainColor,font=\normalsize] at (-0.59,0,-0.3) {$\mathbf{T}_1$};
            \end{scope}
            \end{scope}
            \coordinate (c1) at (0,1.3,1.8);
            
            \begin{scope}
                \draw[relaxColor, semithick, opacity=0.9, -{Stealth[length=1.2mm]}] 
                    (c1) to[bend left=55] node[pos=0.5, font=\normalsize, fill=white, inner sep=0.5pt, rounded corners, shift={(-0.2cm, 0.1cm)}] {$\mathbf{T}_{12}$} (c2);
                \draw[relaxColor, semithick, opacity=0.9, -{Stealth[length=1.2mm]}] 
                    (c2) to[bend right=20] node[pos=0.6, font=\normalsize, fill=white, inner sep=0.5pt, rounded corners, shift={(0.15cm, 0.05cm)}] {$\mathbf{T}_{23}$} (c3);
                \draw[relaxColor, semithick, opacity=0.9, -{Stealth[length=1.2mm]}] 
                    (c3) to[bend right=25] node[pos=0.5, font=\normalsize, fill=white, inner sep=0.5pt, rounded corners, shift={(0.07cm, -0.25cm)}] {$\mathbf{T}_{34}$} (c4);
                \draw[relaxColor, semithick, opacity=0.9, -{Stealth[length=1.2mm]}] 
                    (c4) to[bend left=20] node[pos=0.5, below, font=\normalsize, fill=white, inner sep=0.5pt, rounded corners, shift={(0.05cm, 0.18cm)}] {$\mathbf{T}_{45}$} (c5);
                
                \draw[boundColor, semithick, opacity=0.9, dashed, -{Stealth[length=1.2mm]}] 
                    (c1) to[bend right=40] node[pos=0.65, above, font=\normalsize, fill=white, inner sep=0.5pt, rounded corners, shift={(0.3cm, 0cm)}] {$\mathbf{T}_{14}$} (c4);
                \draw[boundColor, semithick, opacity=0.9, dashed, -{Stealth[length=1.2mm]}] 
                    (c2) to[bend right=30] node[pos=0.4, above right, font=\normalsize, fill=white, inner sep=0.5pt, rounded corners, shift={(-0.26cm, -0.1cm)}] {$\mathbf{T}_{25}$} (c5);
            \end{scope}

        \end{scope}
    \end{scope}

    \node[font=\footnotesize\bfseries] at (\cellW/2, 0.25) {(i) Pose Graph Optimization};
    
\end{scope}

%% file: fig/tasks/ba.tex
\begin{scope}[shift={(4*\stepX, \rowOffset)}]
    \fill[boundColor!8] (0,0) rectangle (\cellW, \cellH);

    \begin{scope}[shift={(1.8, 2.12)}, scale=0.429]
        \tdplotsetmaincoords{78}{5}
        \begin{scope}[tdplot_main_coords]

            \input{fig/tasks/ba_data}
            
            \begin{scope}[transparency group, opacity=0.6]
            \pgfmathtruncatemacro{\maxPoints}{\numPoints-1}
            \foreach \i in {0,...,\maxPoints} {
                \fill[boundColor] (X\i) ++ (0.7, 0, 6) circle (1pt);
            }
            \end{scope}

            \coordinate (O1) at (-2.8, -3.1, -1.1);
            \coordinate (c1A) at (-3.2, -4, 0);
            \coordinate (c1B) at (-3.2, -4, 2);
            \coordinate (c1C) at (-2, 2, 0);
            \coordinate (c1D) at (-2, 2, -2);
            
            \draw[domainColor, line width=1.2pt, fill=white, fill opacity=0.92] 
                (c1A) -- (c1B) -- (c1C) -- (c1D) -- cycle;
            \draw[domainColor!80, line width=0.5pt, opacity=0.6, dashed] (O1) -- (c1A);
            \draw[domainColor!80, line width=0.5pt, opacity=0.6, dashed] (O1) -- (c1B);
            \draw[domainColor!80, line width=0.5pt, opacity=0.6, dashed] (O1) -- (c1C);
            \draw[domainColor!80, line width=0.5pt, opacity=0.6, dashed] (O1) -- (c1D);
            
            \coordinate (p11) at (-2.87, -0.3, -0.4);
            \coordinate (p12) at (-2.23, -0.3, -0.46);
            \coordinate (p13) at (-2.33, -0.3, 0);

            \coordinate (O2) at (0.5, -3, -1.5);
            \coordinate (c2A) at (-0.9, 2, -2);
            \coordinate (c2B) at (-0.9, 2, 0);
            \coordinate (c2C) at (1.4, -4.5, 1.5);
            \coordinate (c2D) at (1.4, -4.5, -0.5);
            
            \draw[domainColor, line width=1.2pt, fill=white, fill opacity=0.92] 
                (c2A) -- (c2B) -- (c2C) -- (c2D) -- cycle;
            \draw[domainColor!80, line width=0.5pt, opacity=0.6, dashed] (O2) -- (c2A);
            \draw[domainColor!80, line width=0.5pt, opacity=0.6, dashed] (O2) -- (c2B);
            \draw[domainColor!80, line width=0.5pt, opacity=0.6, dashed] (O2) -- (c2C);
            \draw[domainColor!80, line width=0.5pt, opacity=0.6, dashed] (O2) -- (c2D);
            
            \coordinate (p21) at (-0.168, -0.3, -0.53);
            \coordinate (p22) at (0.55, -0.3, -0.4);
            \coordinate (p23) at (0.1, -0.2, -0.2);

            \coordinate (O3) at (3.5, -2.9, -0.9);
            \coordinate (c3A) at (2, 2, -1.6);
            \coordinate (c3B) at (2, 2, 0.4);
            \coordinate (c3C) at (4, -4, 2.5);
            \coordinate (c3D) at (4, -4, 0.5);
            
            \draw[domainColor, line width=1.2pt, fill=white, fill opacity=0.92] 
                (c3A) -- (c3B) -- (c3C) -- (c3D) -- cycle;
            \draw[domainColor!80, line width=0.5pt, opacity=0.6, dashed] (O3) -- (c3A);
            \draw[domainColor!80, line width=0.5pt, opacity=0.6, dashed] (O3) -- (c3B);
            \draw[domainColor!80, line width=0.5pt, opacity=0.6, dashed] (O3) -- (c3C);
            \draw[domainColor!80, line width=0.5pt, opacity=0.6, dashed] (O3) -- (c3D);
            
            \coordinate (p31) at (2.53, -0.3, -0.4);
            \coordinate (p32) at (2.755, -0.3, 0.5);
            \coordinate (p33) at (2.47, -0.2, -0.08);
            
            \coordinate (X1) at (-2, -0.2, 6);
            \coordinate (X2) at (1.5, -0.2, 5.2);
            \coordinate (X3) at (-0.3, -0.2, 4.8);
            
            \draw[gray, line width=0.5pt, opacity=0.6] (O1) -- (X1);
            \draw[gray, line width=0.5pt, opacity=0.6] (O2) -- (X1);
            \draw[gray, line width=0.5pt, opacity=0.6] (O3) -- (X1);
            
            \draw[gray, line width=0.5pt, opacity=0.6] (O1) -- (X2);
            \draw[gray, line width=0.5pt, opacity=0.6] (O2) -- (X2);
            \draw[gray, line width=0.5pt, opacity=0.6] (O3) -- (X2);
            
            \draw[gray, line width=0.5pt, opacity=0.6] (O1) -- (X3);
            \draw[gray, line width=0.5pt, opacity=0.6] (O2) -- (X3);
            \draw[gray, line width=0.5pt, opacity=0.6] (O3) -- (X3);
            
            \fill[boundColor] (X1) circle (4.5pt);
            \fill[boundColor] (X2) circle (4.5pt);
            \fill[boundColor] (X3) circle (4.5pt);
            
            \fill[observColor] (p11) circle (2.5pt);
            \fill[observColor] (p21) circle (2.5pt);
            \fill[observColor] (p31) circle (2.5pt);
            
            \fill[observColor] (p12) circle (2.5pt);
            \fill[observColor] (p22) circle (2.5pt);
            \fill[observColor] (p32) circle (2.5pt);
            
            \fill[observColor] (p13) circle (2.5pt);
            \fill[observColor] (p23) circle (2.5pt);
            \fill[observColor] (p33) circle (2.5pt);
            
            \fill[domainColor] (O1) circle (3.5pt);
            \node[domainColor, font=\bfseries\normalsize, below, shift={(0.12cm,0cm)}] at (O1) {$(\mathbf{R}_1, \mathbf{t}_1)$};
            
            \fill[domainColor] (O2) circle (3.5pt);
            \node[domainColor, font=\bfseries\normalsize, below, shift={(0cm,0cm)}] at (O2) {$(\mathbf{R}_2, \mathbf{t}_2)$};
            
            \fill[domainColor] (O3) circle (3.5pt);
            \node[domainColor, font=\bfseries\normalsize, below, shift={(0.05cm,0cm)}] at (O3) {$(\mathbf{R}_3, \mathbf{t}_3)$};
            
            \node[boundColor, font=\bfseries\normalsize, anchor=south] at (4.67,-0.2,4.5) {$\{\mathbf{X}_j\}$};
            \node[observColor, font=\bfseries\normalsize, anchor=north] at (4.67,-0.3,1) {$\{\mathbf{x}_{ij}\}$};

        \end{scope}
    \end{scope}

    \node[font=\footnotesize\bfseries] at (\cellW/2, 0.25) {(j) Bundle Adjustment};
    
\end{scope}

%% file: fig/solver_task_index.tex
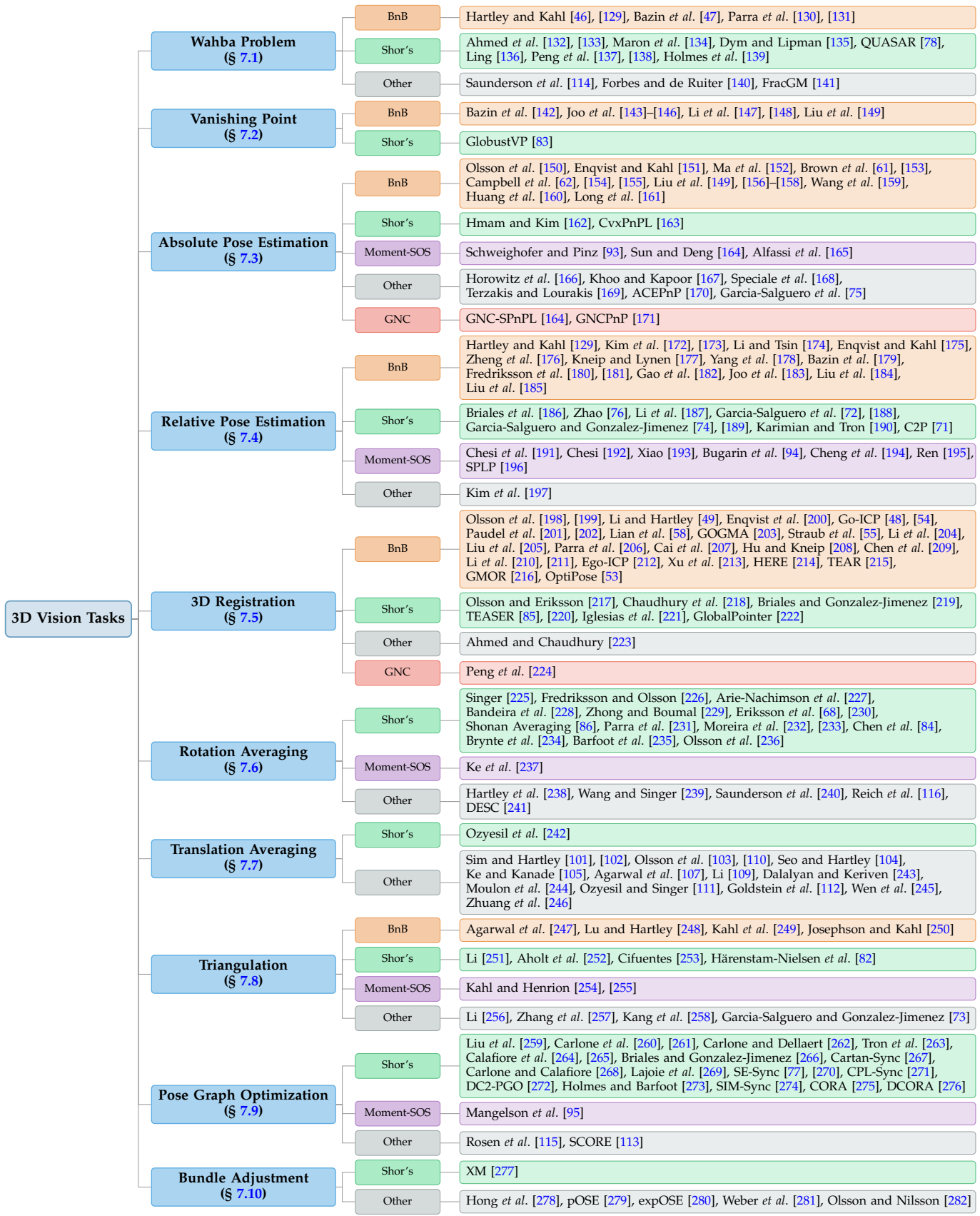
\begin{figure*}[!ht]
\centering
\scriptsize

\definecolor{rootColor}{HTML}{2874A6}
\definecolor{taskColor}{HTML}{3498DB}
\definecolor{taskLight}{HTML}{85C1E9}

\definecolor{bnbColor}{HTML}{E67E22}
\definecolor{bnbLight}{HTML}{F5CBA7}
\definecolor{shorColor}{HTML}{27AE60}
\definecolor{shorLight}{HTML}{ABEBC6}
\definecolor{sosColor}{HTML}{9B59B6}
\definecolor{sosLight}{HTML}{D7BDE2}
\definecolor{gncColor}{HTML}{E74C3C} 
\definecolor{gncLight}{HTML}{F5B7B1}
\definecolor{otherColor}{HTML}{7F8C8D}
\definecolor{otherLight}{HTML}{D5DBDB}

\definecolor{solverVeryLight}{HTML}{F8E6D6}

\forestset{
  tasktree/.style={
    for tree={
      grow'=0,
      parent anchor=east,
      child anchor=west,
      anchor=west,
      edge path'={(!u.parent anchor) -- +(4pt,0) |- (.child anchor)},
      edge={draw, line width=0.4pt, color=black!40},
      rounded corners=2pt,
      draw,
      minimum height=5mm,
      inner xsep=4pt,
      inner ysep=2pt,
      s sep=2.5pt,
      l sep=12pt,
      align=left,
      font=\small
    },
    root/.style={draw, rounded corners=3pt, fill=rootColor!20, 
                 minimum height=8mm, minimum width=10mm,
                 align=center, text centered, font=\bfseries\normalsize,
                 draw=rootColor!70, line width=0.8pt},
    task/.style={draw, rounded corners=2pt, fill=taskLight!70,
                 minimum height=6mm, text width=38mm,
                 align=center, text centered, font=\small\bfseries,
                 draw=taskColor!80, line width=0.6pt},
    bnb/.style={draw, rounded corners=2pt, fill=bnbLight,
                minimum height=5.5mm, text width=16mm,
                align=center, text centered, text centered, font=\scriptsize,
                draw=bnbColor!80, line width=0.4pt},
    shor/.style={draw, rounded corners=2pt, fill=shorLight,
                 minimum height=5.5mm, text width=16mm,
                 align=center, text centered, font=\scriptsize,
                 draw=shorColor!80, line width=0.4pt},
    sos/.style={draw, rounded corners=2pt, fill=sosLight,
                minimum height=5.5mm, text width=16mm,
                align=center, text centered, font=\scriptsize,
                draw=sosColor!80, line width=0.4pt},
    gnc/.style={draw, rounded corners=2pt, fill=gncLight,
                minimum height=5.5mm, text width=16mm,
                align=center, text centered, font=\scriptsize,
                draw=gncColor!80, line width=0.4pt},
    other/.style={draw, rounded corners=2pt, fill=otherLight,
                  minimum height=5.5mm, text width=16mm,
                  align=center, text centered, font=\scriptsize,
                  draw=otherColor!80, line width=0.4pt},
    paper_bnb/.style={draw, rounded corners=2pt, fill=bnbLight!50,
                minimum height=5mm, text width=45.2em,
                align=left, font=\footnotesize,
                draw=bnbColor!80, line width=0.4pt},
    paper_shor/.style={draw, rounded corners=2pt, fill=shorLight!50,
                minimum height=5mm, text width=45.2em,
                align=left, font=\footnotesize,
                draw=shorColor!80, line width=0.4pt},
    paper_sos/.style={draw, rounded corners=2pt, fill=sosLight!50,
                minimum height=5mm, text width=45.2em,
                align=left, font=\footnotesize,
                draw=sosColor!80, line width=0.4pt},
    paper_gnc/.style={draw, rounded corners=2pt, fill=gncLight!50,
                minimum height=5mm, text width=45.2em,
                align=left, font=\footnotesize,
                draw=gncColor!80, line width=0.4pt},
    paper_other/.style={draw, rounded corners=2pt, fill=otherLight!50,
                minimum height=5mm, text width=45.2em,
                align=left, font=\footnotesize,
                draw=otherColor!80, line width=0.4pt},
  }
}

\resizebox{1\textwidth}{!}{
\begin{forest} tasktree
[{3D Vision Tasks}, root
  [{Wahba Problem\\(\S~\ref{subsec:wahba})}, task
    [{BnB}, bnb
      [{Hartley and Kahl~\cite{hartley2007bnb,hartley2009bnb},
        Bazin~\etal~\cite{bazin2012globally},
        Parra~\etal~\cite{parra2014fast,bustos2016fast}}, paper_bnb]
    ]
    [{Shor's}, shor
      [{Ahmed~\etal~\cite{ahmed2011semidefinite,ahmed2012semidefinite},
        Maron~\etal~\cite{maron2016point},
        Dym and Lipman~\cite{dym2017exact},
        QUASAR~\cite{yang2019quaternion},\\
        Ling~\cite{ling2021near},
        Peng~\etal~\cite{peng2022semidefinite,peng2022towards},
        Holmes~\etal~\cite{holmes2024semidefinite}}, paper_shor]
    ]
    [{Other}, other
      [{Saunderson~\etal~\cite{saunderson2015semidefinite},
        Forbes and de Ruiter~\cite{forbes2015linear},
        FracGM~\cite{chen2024fracgm}}, paper_other]
    ]
  ]
  [{Vanishing Point\\(\S~\ref{subsec:vp})}, task
    [{BnB}, bnb
      [{Bazin~\etal~\cite{bazin2012globallyvp},
        Joo~\etal~\cite{joo2016globally,joo2018robust,joo2018globally,joo2019globally},
        Li~\etal~\cite{li2019quasi,li2020quasi},
        Liu~\etal~\cite{liu2020globally}}, paper_bnb]
    ]
    [{Shor's}, shor
      [{GlobustVP~\cite{liao2025convex}}, paper_shor]
    ]
  ]
  [{Absolute Pose Estimation\\(\S~\ref{subsec:pnp})}, task
    [{BnB}, bnb
      [{Olsson~\etal~\cite{olsson2006optimal},
        Enqvist and Kahl~\cite{enqvist2008robust},
        Ma~\etal~\cite{ma2010linear}, Brown~\etal~\cite{brown2015globally,brown2019family},\\
        Campbell~\etal~\cite{campbell2017globally,campbell2018globally,campbell2019alignment},
        Liu~\etal~\cite{liu20182d,liu2019novel,liu2020globally,liu2023absolute},
        Wang~\etal~\cite{wang2021efficient},\\
        Huang~\etal~\cite{huang2024efficientpnp},
        Long~\etal~\cite{long2025bnb}}, paper_bnb]
    ]
    [{Shor's}, shor
      [{Hmam and Kim~\cite{hmam2010optimal},
        CvxPnPL~\cite{agostinho2023cvxpnpl}}, paper_shor]
    ]
    [{Moment-SOS}, sos
      [{Schweighofer and Pinz~\cite{schweighofer2008globally},
        Sun and Deng~\cite{sun2020certifiably},
        Alfassi~\etal~\cite{alfassi2021non}}, paper_sos]
    ]
    [{Other}, other
      [{Horowitz~\etal~\cite{horowitz2014convex},
        Khoo and Kapoor~\cite{khoo2016non},
        Speciale~\etal~\cite{speciale2017consensus},\\
        Terzakis and Lourakis~\cite{terzakis2020consistently},
        ACEPnP~\cite{sun2023efficient},
        Garcia-Salguero~\etal~\cite{garcia2024fast}}, paper_other]
    ]
    [{GNC}, gnc
      [{GNC-SPnPL~\cite{sun2020certifiably},
        GNCPnP~\cite{yi2025outliers}}, paper_gnc]
    ]
  ]
  [{Relative Pose Estimation\\(\S~\ref{subsec:relpose})}, task
    [{BnB}, bnb
      [{Hartley and Kahl~\cite{hartley2007bnb},
        Kim~\etal~\cite{kim2008motion,kim2009motion},
        Li and Tsin~\cite{li2009globally},
        Enqvist and Kahl~\cite{enqvist2009two},\\
        Zheng~\etal~\cite{zheng2011branch},
        Kneip and Lynen~\cite{kneip2013direct},
        Yang~\etal~\cite{yang2014optimal},
        Bazin~\etal~\cite{bazin2014globally},\\
        Fredriksson~\etal~\cite{fredriksson2015practical,fredriksson2016optimal},
        Gao~\etal~\cite{gao2020efficient},
        Joo~\etal~\cite{joo2020globally},
        Liu~\etal~\cite{liu2021globally},\\
        Liu~\etal~\cite{liu2022globallyinlier}}, paper_bnb]
    ]
    [{Shor's}, shor
      [{Briales~\etal~\cite{briales2018certifiably},
        Zhao~\cite{zhao2020efficient},
        Li~\etal~\cite{li2020robot},
        Garcia-Salguero~\etal~\cite{garcia2021certifiable,garcia2022tighter},\\
        Garcia-Salguero and Gonzalez-Jimenez~\cite{garcia2023fast,garcia2024certifiable},
        Karimian and Tron~\cite{karimian2023essential},
        C2P~\cite{tirado2024correspondences}}, paper_shor]
    ]
    [{Moment-SOS}, sos
      [{Chesi~\etal~\cite{chesi2002estimating},
        Chesi~\cite{chesi2008camera},
        Xiao~\cite{xiao2010new},
        Bugarin~\etal~\cite{bugarin2015rank},
        Cheng~\etal~\cite{cheng2015convex},
        Ren~\cite{ren2019non},\\
        SPLP~\cite{sun2022splp}}, paper_sos]
    ]
    [{Other}, other
      [{Kim~\etal~\cite{kim2007visual}}, paper_other]
    ]
  ]
  [{3D Registration\\(\S~\ref{subsec:registration})}, task
    [{BnB}, bnb
      [{Olsson~\etal~\cite{olsson2006registration,olsson2008branch},
        Li and Hartley~\cite{li20073d},
        Enqvist~\etal~\cite{enqvist2009optimal},
        Go-ICP~\cite{yang2013go,yang2015go},\\
        Paudel~\etal~\cite{paudel2015robust,paudel2019robust},
        Lian~\etal~\cite{lian2016efficient},
        GOGMA~\cite{campbell2016gogma},
        Straub~\etal~\cite{straub2017efficient},
        Li~\etal~\cite{li2018fast},\\
        Liu~\etal~\cite{liu2018efficient},
        Parra~\etal~\cite{bustos2019practical},
        Cai~\etal~\cite{cai2019practical},
        Hu and Kneip~\cite{hu2021globally},
        Chen~\etal~\cite{chen2022deterministic},\\
        Li~\etal~\cite{li2023fast,li2024efficient},
        Ego-ICP~\cite{liu2023chebyshev},
        Xu~\etal~\cite{xu2024hierarchical},
        HERE~\cite{huang2024efficient},
        TEAR~\cite{huang2024scalable},\\
        GMOR~\cite{zheng2025robust},
        OptiPose~\cite{ivanov2025fast}}, paper_bnb]
    ]
    [{Shor's}, shor
      [{Olsson and Eriksson~\cite{olsson2008solving},
        Chaudhury~\etal~\cite{chaudhury2015global},
        Briales and Gonzalez-Jimenez~\cite{briales2017convex},\\
        TEASER~\cite{yang2019polynomial,yang2020teaser},
        Iglesias~\etal~\cite{iglesias2020global},
        GlobalPointer~\cite{liao2024globalpointer}}, paper_shor]
    ]
    [{Other}, other
      [{Ahmed and Chaudhury~\cite{ahmed2017global}}, paper_other]
    ]
    [{GNC}, gnc
      [{Peng~\etal~\cite{peng2023convergence}}, paper_gnc]
    ]
  ]
  [{Rotation Averaging\\(\S~\ref{subsec:rot-avg})}, task
    [{Shor's}, shor
      [{Singer~\cite{singer2011angular},
        Fredriksson and Olsson~\cite{fredriksson2012simultaneous},
        Arie-Nachimson~\etal~\cite{arie2012global},\\
        Bandeira~\etal~\cite{bandeira2017tightness},
        Zhong and Boumal~\cite{zhong2018near},
        Eriksson~\etal~\cite{eriksson2018rotation,eriksson2019rotation},\\
        Shonan Averaging~\cite{dellaert2020shonan},
        Parra~\etal~\cite{parra2021rotation},
        Moreira~\etal~\cite{moreira2021rotation,moreira2024rotation},
        Chen~\etal~\cite{chen2021hybrid},\\
        Brynte~\etal~\cite{brynte2022tightness},
        Barfoot~\etal~\cite{barfoot2025certifiably},
        Olsson~\etal~\cite{olsson2025certifiably}}, paper_shor]
    ]
    [{Moment-SOS}, sos
      [{Ke~\etal~\cite{ke2014globally}}, paper_sos]
    ]
    [{Other}, other
      [{Hartley~\etal~\cite{hartley2010rotation},
        Wang and Singer~\cite{wang2013exact},
        Saunderson~\etal~\cite{saunderson2014semidefinite},
        Reich~\etal~\cite{reich2017global},\\
        DESC~\cite{shi2022robust}}, paper_other]
    ]
  ]
  [{Translation Averaging\\(\S~\ref{subsec:trans-avg})}, task
    [{Shor's}, shor
      [{Ozyesil~\etal~\cite{ozyesil2015stable}}, paper_shor]
    ]
    [{Other}, other
      [{Sim and Hartley~\cite{sim2006recovering,sim2006removing},
        Olsson~\etal~\cite{olsson2007efficient,olsson2010outlier},
        Seo and Hartley~\cite{seo2007fast},\\
        Ke and Kanade~\cite{ke2007quasiconvex},
        Agarwal~\etal~\cite{agarwal2008fast},
        Li~\cite{li2009efficient},
        Dalalyan and Keriven~\cite{dalalyan2009l_1},\\
        Moulon~\etal~\cite{moulon2013global},
        Ozyesil and Singer~\cite{ozyesil2015robust},
        Goldstein~\etal~\cite{goldstein2016shapefit},
        Wen~\etal~\cite{wen2018efficient},\\
        Zhuang~\etal~\cite{zhuang2018baseline}}, paper_other]
    ]
  ]
  [{Triangulation\\(\S~\ref{subsec:triangulation})}, task
    [{BnB}, bnb
      [{Agarwal~\etal~\cite{agarwal2006practical},
        Lu and Hartley~\cite{lu2007fast},
        Kahl~\etal~\cite{kahl2008practical},
        Josephson and Kahl~\cite{josephson2008triangulation}}, paper_bnb]
    ]
    [{Shor's}, shor
      [{Li~\cite{li2010multi},
        Aholt~\etal~\cite{aholt2012qcqp},
        Cifuentes~\cite{cifuentes2021convex},
        H{\"a}renstam-Nielsen~\etal~\cite{harenstam2023semidefinite}}, paper_shor]
    ]
    [{Moment-SOS}, sos
      [{Kahl and Henrion~\cite{kahl2005globally,kahl2007globally}}, paper_sos]
    ]
    [{Other}, other
      [{Li~\cite{li2007practical},
        Zhang~\etal~\cite{zhang2010triangulation},
        Kang~\etal~\cite{kang2014robust},
        Garcia-Salguero and Gonzalez-Jimenez~\cite{garcia2023certifiable}}, paper_other]
    ]
  ]
  [{Pose Graph Optimization\\(\S~\ref{subsec:pgo})}, task
    [{Shor's}, shor
      [{Liu~\etal~\cite{liu2012convex},
        Carlone~\etal~\cite{carlone2015lagrangian,carlone2016planar},
        Carlone and Dellaert~\cite{carlone2015duality},
        Tron~\etal~\cite{tron2015inclusion},\\
        Calafiore~\etal~\cite{calafiore2015pose,calafiore2016lagrangian},
        Briales and Gonzalez-Jimenez~\cite{briales2016fast},
        Cartan-Sync~\cite{briales2017cartan},\\
        Carlone and Calafiore~\cite{carlone2018convex},
        Lajoie~\etal~\cite{lajoie2019modeling},
        SE-Sync~\cite{rosen2019se,rosen2020certifiably},
        CPL-Sync~\cite{fan2019efficient},\\
        DC2-PGO~\cite{tian2021distributed},
        Holmes and Barfoot~\cite{holmes2023efficient},
        SIM-Sync~\cite{yu2024sim},
        CORA~\cite{papalia2024certifiably},
        DCORA~\cite{thoms2025distributed}}, paper_shor]
    ]
    [{Moment-SOS}, sos
      [{Mangelson~\etal~\cite{mangelson2019guaranteed}}, paper_sos]
    ]
    [{Other}, other
      [{Rosen~\etal~\cite{rosen2015convex},
        SCORE~\cite{papalia2022score}}, paper_other]
    ]
  ]
  [{Bundle Adjustment\\(\S~\ref{subsec:bundle-adjustment})}, task
    [{Shor's}, shor
      [{XM~\cite{han2025building}}, paper_shor]
    ]
    [{Other}, other
      [{Hong~\etal~\cite{hong2016projective},
        pOSE~\cite{hong2018pose},
        expOSE~\cite{iglesias2023expose},
        Weber~\etal~\cite{weber2024power},
        Olsson and Nilsson~\cite{olsson2025towards}}, paper_other]
    ]
  ]
]
\end{forest}
}

\caption{%
    Comprehensive index of global optimization methods for 3D vision tasks.
    The ten geometric estimation problems are organized by the solver families applied:
    \textcolor{bnbColor}{\textbf{BnB}} (Branch-and-Bound),
    \textcolor{shorColor}{\textbf{Shor's}} (Shor's Relaxation),
    \textcolor{sosColor}{\textbf{Moment-SOS}} (Moment-SOS Relaxation),
    \textcolor{otherColor}{\textbf{Other}} (SOCP, $L_\infty$, alternative convex relaxations, and other global methods), and
    \textcolor{gncColor}{\textbf{GNC}} (Graduated Non-Convexity).
    Representative works are listed for each solver-task combination.
}

\label{fig:solver-task-index}
\end{figure*}

%% file: sec/task_subsec/1_wahba.tex
\subsection{Wahba Problem}
\label{subsec:wahba}

\PAR{Problem definition.}
Given two sets of corresponding vectors $\{\ba_i \!\in\! \bbR^3\}_{i=1}^N$ and $\{\bbb_i \!\in\! \bbR^3\}_{i=1}^N$, \emph{Wahba problem} (a.k.a.\ \emph{rotation search} or \emph{rotation estimation}) seeks to estimate the rotation $\bR \!\in\! \mathrm{SO}(3)$ to align them.
Wahba problem is generally formulated as a least-squares estimation:
\begin{equation}
\label{eq:wahba-ls}
    \min_{\bR\in\mathrm{SO}(3)} \; \sum_{i=1}^N w_i^2 \, \| \bbb_i - \bR \ba_i \|_2^2, \tag{Wahba}
\end{equation}
where $\{w_i \!\in\! \bbR\}_{i=1}^N$ are known confidence weights associated to each pair of vectors.
\eqref{eq:wahba-ls} is algebraically equivalent to maximizing $\trace(\bR^\top \bM)$ with $\bM \!=\! \sum_i w_i^2 \bbb_i \ba_i^\top$.
A number of applications, including satellite attitude determination~\cite{cheng2019total,chin2019star,wahba1965least}, image stitching~\cite{meneghetti2015image}, and motion estimation and 3D reconstruction~\cite{blais1995registering,choi2015robust}, reduce to (or contain) this core problem.

\PAR{Properties and local methods.}
\eqref{eq:wahba-ls} is nonconvex due to the orthogonality constraint on $\bR$.
Iterative schemes~\cite{kruzhilov2014iteration} have been explored for variants of the Wahba problem. While flexible, such local methods often lack global convergence guarantees, may get trapped in local minima, and typically require reliable initialization.
When outliers are present, however, the least-squares formulation becomes sensitive to contaminated correspondences.

\PAR{Closed-form solutions.}
Despite the nonconvexity, the outlier-free Wahba problem admits a closed-form global solution via singular value decomposition (SVD).
Given $\bM \!=\! \sum_i w_i^2 \bbb_i \ba_i^\top$, the optimal rotation is obtained as $\bR^* \!=\! \bU \operatorname{diag}(1,1,\det(\bU\bV^\top)) \bV^\top$ where $\bM \!=\! \bU \bS \bV^\top$ is the SVD of $\bM$~\cite{kabsch1976solution,horn1987closed,arun1987least}.
This solution is exact and globally optimal in the outlier-free case.
Numerous variants of this closed-form approach~\cite{schonemann1966generalized,horn1987closed,horn1988closed,markley1988attitude,arun1987least,khoshelham2016closed} have been developed for different cost functions and constraints, all leveraging SVD-based solutions.

\PAR{Branch-and-Bound (BnB).}
BnB-based rotation search achieves global optimality by recursively partitioning the rotation domain (\eg angle–axis, quaternion, or stereographic parameterizations) and discarding regions that cannot contain the optimum via admissible bounds.
Hartley and Kahl~\cite{hartley2007bnb,hartley2009bnb} develop BnB methods using $L_\infty$ norms, solving the optimization as a series of SOCP feasibility problems over the rotation space.
Bazin~\etal~\cite{bazin2012globally} extend this framework to consensus maximization, explicitly handling outliers by maximizing the number of inliers within a threshold.
To improve efficiency, subsequent work exploits stereographic projection, which provides compact representations and tight bounding functions, leading to faster BnB convergence~\cite{parra2014fast,bustos2016fast}.
Although often presented in the context of 3D registration, these methods typically assume known translation, thereby reducing the problem to pure rotation search.

\PAR{Shor's relaxation.}
Shor's relaxation provides a principled route to certifiable solutions of Wahba problem.

\noindent \emph{Extended Wahba formulations.}
Ahmed~\etal~\cite{ahmed2011semidefinite,ahmed2012semidefinite} formulate robust attitude estimation (\ie Wahba problem) under infinity-norm bounded uncertainties as a nonlinear optimization and derive an SDP relaxation that is provably tight both in theory and in practice.
Yang and Carlone~\cite{yang2019quaternion} introduce QUASAR, an SDP relaxation of the TLS formulation under quaternion parameterization, proving that it certifiably recovers the globally optimal rotation even with a large fraction of outliers. Peng~\etal~\cite{peng2022semidefinite,peng2022towards} further analyze the tightness of TLS-based SDP relaxations, identifying conditions under which exact recovery holds and demonstrating that tightness depends on noise level, truncation parameters, and outlier distribution.
Holmes~\etal~\cite{holmes2024semidefinite} investigate SDP relaxations for robot state-estimation problems including Wahba that have been generalized to include matrix weights arising from anisotropic sensor noise, providing theoretical analysis on the tightness of the relaxation and demonstrating that globally optimal solutions can be certifiably recovered under practical noise levels even when the traditional scalar-weighting assumption is violated.

\noindent \emph{Orthogonal Procrustes extensions.}
The Wahba problem can be viewed as a special case of the Orthogonal Procrustes problem, which is itself subsumed by the more general Procrustes matching formulation.
In the Orthogonal Procrustes problem~\cite{gower2004procrustes}, one seeks an orthogonal transformation (not restricted to $\mathrm{SO}(n)$) with equal weights $w_i {=} 1$.
Maron~\etal~\cite{maron2016point} propose convex SDP relaxations for Procrustes matching that achieve exact global guarantees, while Dym and Lipman~\cite{dym2017exact} rigorously analyze their tightness, establishing recovery guarantees in the noiseless regime.
Ling~\cite{ling2021near} extends this framework to the generalized orthogonal Procrustes problem, showing that a generalized power method with spectral initialization converges to the global optimum of the associated SDP relaxation under a signal–plus–noise model.

\PAR{Other relaxations.}
Beyond Shor's relaxation, alternative convex formulations and specialized optimization techniques provide efficient solutions for Wahba.
Saunderson~\etal~\cite{saunderson2015semidefinite} formulate SDP representations for the convex hull of $\mathrm{SO}(n)$, showing that both the convex hull and its polar are doubly spectrahedral and providing explicit minimal-size SDP descriptions for exact convex relaxations.
Forbes and de Ruiter~\cite{forbes2015linear} recast the Wahba problem as a convex Linear Matrix Inequality (LMI) optimization by relaxing the $\mathrm{SO}(3)$ constraint, proving that under mild assumptions this LMI formulation is equivalent to the original problem and exactly recovers the optimal rotation.
Chen~\etal~\cite{chen2024fracgm} propose FracGM, a fast fractional programming solver for the GM robust estimator that reformulates the nonconvex fractional problem into a convex dual problem plus a linear system, achieving faster convergence and conditional global optimality guarantees for Wahba and related problems including point cloud registration.

\PAR{Robust preprocessing.}
Parra and Chin~\cite{parra2015guaranteed} introduce Guaranteed Outlier Removal (GORE), a deterministic preprocessing technique that leverages the underlying problem geometry to provably discard correspondences that do not belong to the maximum consensus set. By pruning the search space while retaining all inliers, GORE significantly enhances the efficiency of subsequent global optimization methods, such as BnB.  
Peng~\etal~\cite{peng2022arcs} further improve the robustness of rotation search under noise by approximately solving consensus maximization problems using techniques from robust subspace learning~\cite{lerman2018overview} and interval stabbing~\cite{de2008computational}.

%% file: sec/task_subsec/2_vp.tex
\subsection{Vanishing Point Estimation}
\label{subsec:vp}

\PAR{Problem definition.}
Given a set of image lines $\{\ell_j \!\in\! \bbR^3\}_{j=1}^M$ in a calibrated image with known camera calibration matrix $\bK \!\in\! \mathcal{K}$, \emph{vanishing point (VP) estimation} seeks to estimate the vanishing point(s) $\{\bv_i \!\in\! \bbR^3\}_{i=1}^N$, each corresponding to a family of image lines.
VP estimation is generally formulated as a least-squares problem:
\begin{equation}
\label{eq:vp-ls}
    \min_{\bd_i\in\bbR^3} \sum_{i=1}^N \sum_{j \in \mathcal{L}_i} 
    \| \bn_{j}^\top \bd_i \|_2^2,
    \tag{VP}
\end{equation}
where $\bd_i \!=\! \bK^{-1} \bv_i$ represents the 3D direction of the $i$-th VP, $\mathcal{L}_i$ denotes the set of lines associated with the $i$-th VP, and $\bn_{j} \!=\! \bK^{-\top} \ell_{j}$ is the normal vector of the $j$-th line in normalized image coordinates~\cite{hartley2003multiple}.
VP estimation underpins applications in SLAM~\cite{li2023hong}, camera calibration~\cite{lee2013automatic,cipolla1999camera}, single-view geometry~\cite{hedau2009recovering}, and scene understanding~\cite{flint2011manhattan}.

\PAR{Properties and local methods.}
\eqref{eq:vp-ls} is nonconvex on the projective plane and becomes combinatorial in the presence of multiple vanishing points, making globally optimal solutions challenging.
Classical methods for VP estimation are largely heuristic and local.
Hough-transform-based voting~\cite{barnard1982interpreting,lutton1994contribution} provides efficient coarse estimation but is sensitive to discretization artifacts.
RANSAC-style consensus maximization~\cite{zhang2016vanishing,toldo2008robust} offers robustness to outliers but without global optimality guarantees.
Least-squares fitting~\cite{caprile1990using} and expectation–maximization (EM)-based refinements~\cite{antone2000automatic,schindler2004atlanta,tardif2009non} improve accuracy, yet remain prone to local minima and initialization sensitivity, motivating the development of globally optimal methods.

\PAR{Branch-and-Bound (BnB).}
BnB has emerged as the dominant framework for globally optimal VP estimation, with methods typically organized around structured world models such as Manhattan or Atlanta assumptions.

\noindent \emph{Manhattan world methods.}
Under the Manhattan world assumption (three orthogonal dominant directions), several BnB methods provide global optimality guarantees.
Bazin~\etal~\cite{bazin2012globallyvp} formulate VP estimation as a consensus set maximization problem over the rotation search space and solve it by combining BnB and interval analysis theory~\cite{moore1966interval,hansen2003global,moore2009introduction}.
Joo~\etal~\cite{joo2016globally,joo2018robust} relax the consensus set maximization problem using the extended Gaussian image (EGI) representation and define tighter lower and upper bounds to significantly improve BnB efficiency, achieving robust performance in near real-time~\cite{joo2018robust}.
To further enhance efficiency, Li~\etal~\cite{li2019quasi} propose a hybrid approach that computes two degrees of freedom (DoF) of the Manhattan frame rotation via RANSAC-based two-line sampling, then searches for the optimal third DoF using BnB, achieving quasi-global optimality with improved efficiency.
This approach is later extended to near real-time settings with certifiable accuracy~\cite{li2020quasi}.

\noindent \emph{Atlanta world methods.}
The Atlanta world generalizes the Manhattan assumption by allowing multiple vertical dominant directions.
Joo~\etal~\cite{joo2018globally} formulate Atlanta frame estimation as an inlier set maximization problem and solve it using BnB.
To alleviate computational bottlenecks, Joo~\etal~\cite{joo2019globally} present two bound computation strategies, rectangular bound and slice bound in the EGI, and integrate them into the BnB framework, significantly improving efficiency.
Li~\etal~\cite{li2020globally} propose the mine-and-stab (MnS) algorithm embedded in BnB to estimate vanishing points in Atlanta world, achieving global optimality with improved computational efficiency.
Liu~\etal~\cite{liu2020globally} develop a globally optimal method for estimating the vertical direction in Atlanta world using BnB search over the hemisphere parameterized by azimuth and elevation, with tight bounding functions that guarantee convergence without requiring prior knowledge of horizontal frames.

\PAR{Shor's relaxation.}
Inspired by advances in essential matrix estimation~\cite{zhao2020certifiably} and 3D registration~\cite{yang2020teaser}, SDP relaxations have recently been explored for VP estimation.
GlobustVP~\cite{liao2025convex} proposes the first convex relaxation framework for Manhattan-world vanishing point estimation, where a truncated multi-selection error with soft line-VP association enables reformulating the nonconvex QCQP into an SDP with certifiable global optimality while achieving robustness to outliers.

%% file: sec/task_subsec/3_pnp.tex
\subsection{Absolute Pose Estimation}
\label{subsec:pnp}

\PAR{Problem definition.}
Given a set of $N$ putative 3D-2D correspondences between 3D points $\{\bX_i \!\in\! \bbR^3\}_{i=1}^N$ and 2D image keypoints $\{\bx_i \!\in\! \bbR^2\}_{i=1}^N$, \emph{perspective-$n$-point (PnP)} seeks to estimate the absolute camera pose $(\bR,\bt) \!\in\! \mathrm{SE}(3)$ from the 3D-2D correspondence.
PnP is generally formulated as a least-squares estimation minimizing the reprojection errors:
\begin{equation}
\label{eq:pnp-ls}
    \min_{\bR \in \mathrm{SO}(3),\,\bt \in \bbR^3}
    \sum_{i=1}^N \big\| \pi(\bR \bX_i+\bt) - \bx_i \big\|_2^2, \tag{PnP}
\end{equation}
where $\pi\big([v_1;v_2;v_3]\big) \!:=\! [v_1/v_3 ; v_2/v_3]$ is the perspective projection. Alternative geometry-preserving costs replace pixel-space error by angular error between the predicted bearing $\frac{\bR \bX_i+\bt}{\|\bR \bX_i+\bt\|_2}$ and the normalized image bearing $\tilde{\bx}_i$.

\PAR{Properties and local methods.}
\eqref{eq:pnp-ls} is inherently nonconvex due to the rotation manifold and the nonlinear perspective projection.
Classical approaches address this challenge through local optimization, typically alternating between pose parameters or solving linearized subproblems~\cite{lu2000fast,dementhon1995model,david2004softposit,urban2016mlpnp,garro2012solving,schweighofer2006robust,moreno2008pose,zhan2025generalized,zhou2025novel}.
These methods are computationally efficient and effective under moderate noise, yet their performance degrades under poor initialization or high outlier contamination, often leading to convergence to suboptimal local minima.

\PAR{Closed-form and analytical solutions.}
Despite the nonconvexity, exact global solutions can be obtained through direct algebraic methods in outlier-free scenarios.
EPnP~\cite{lepetit2009ep,moreno2007accurate} reduces PnP to retrieving the positions of four control points that span any number of 3D points, enabling a closed-form non-iterative solution.
Hesch and Roumeliotis~\cite{hesch2011direct} introduce the direct least-squares (DLS) method, which computes all stationary points of the first-order Karush--Kuhn--Tucker (KKT) conditions by solving a system of multivariate polynomial equations, guaranteeing discovery of the global optimum among all critical points.
The DLS framework has since been enhanced for improved numerical stability and computational efficiency~\cite{nakano2015globally,zheng2013revisiting}, generalized to non-central and multi-camera systems~\cite{kneip2014upnp,sweeney2014gdls,sweeney2016large,fragoso2020gdls}, extended to diverse 3D registration settings~\cite{wientapper2016unifying,wientapper2018universal}, and adapted to handle unknown intrinsic parameters such as focal length and radial distortion~\cite{nakano2016versatile} as well as line correspondences~\cite{mirzaei2011globally}.
Vakhitov~\etal~\cite{vakhitov2021uncertainty} extend EPnP and DLS with uncertainty-aware variants EPnPU and DLSU, which account for uncertainties in both 3D coordinates and 2D projections during pose estimation.
Wu~\etal~\cite{wu2022quadratic} propose QPEP (quadratic pose estimation problems), which employs a quaternion-based representation and Gr{\"o}bner basis methods to derive globally optimal analytical solutions for a general class of quadratic pose estimation problems including PnP, providing a unified framework that also applies to point-to-plane registration and other geometric estimation tasks.

\PAR{Branch-and-Bound (BnB).}
BnB methods provide globally optimal solutions to PnP by systematically searching over the pose space while pruning suboptimal regions.

\noindent \emph{Early BnB formulations.}
Olsson~\etal~\cite{olsson2006optimal} first integrate convex under-estimators with BnB under the $L_2$ reprojection error, obtaining the first globally optimal PnP solution.
Enqvist and Kahl~\cite{enqvist2008robust} extend this framework to the $L_\infty$ formulation, casting robust PnP with outliers as a mixed-integer program (MIP) and solving it globally by deriving optimality conditions that enable guaranteed outlier rejection.
Ma~\etal~\cite{ma2010linear} perform BnB search over the rotation space, reducing the remaining fixed-rotation problem to LP for translation estimation.

\noindent \emph{Geometric bounds and consensus maximization.}
Brown~\etal~\cite{brown2015globally,brown2019family} introduce nested BnB to achieve the first globally optimal 2D-3D registration without correspondences, applicable to both point and line features.
Campbell~\etal~\cite{campbell2017globally,campbell2018globally} globally maximize the inlier set over $\mathrm{SE}(3)$ and derive tighter geometric bounds for robust pose estimation.
Their follow-up work, GOSMA~\cite{campbell2019alignment}, extends BnB to align 2D-3D mixture models on the unit sphere, using von Mises--Fisher and quasi-Projected Normal mixtures, optimizing the $L_2$ distance between distributions to jointly estimate pose and correspondences without prior matching.
Liu~\etal~\cite{liu20182d} further enhance geometric bounding by deriving four analytical bounds for the projection of 3D points on the image plane, enabling globally optimal 2D-3D point set registration via combined rotation BnB and translation grid search.

\noindent \emph{Decomposition and acceleration.}
Recent works focus on decomposition and acceleration strategies.
RGPnP~\cite{liu2019novel} decouples the original 6-DoF problem into 3-DoF rotation and translation subproblems, each solved globally by a dedicated BnB.
Wang~\etal~\cite{wang2021efficient} introduce a rotation-invariant feature (RIF) representation that separates pose and correspondence search, transforming the 6D optimization into two tractable lower-dimensional BnB stages under an inlier maximization framework.
Liu~\etal~\cite{liu2023absolute} exploit known gravity directions to reduce the search to a 3D translation and a 1D rotation, applying BnB and global voting respectively.
Long~\etal~\cite{long2025bnb} jointly optimize rotation and translation through a consensus maximization cost, reducing the original 6D search space to 3D while deriving tight analytical bounds for efficient BnB pruning.

\noindent \emph{Line-based extensions.}
BnB formulations have also been extended to line-based pose estimation.
Liu~\etal~\cite{liu2020globally} propose a consensus-based BnB framework for line-based pose estimation by decoupling rotation from translation, enabling globally optimal orientation estimation from 2D-3D line correspondences.
To further accelerate convergence, Huang~\etal~\cite{huang2024efficientpnp} incorporate 1D interval stabbing within a 2D BnB search to efficiently estimate rotation.

\PAR{Shor's relaxation.}
SDP relaxations provide certifiable solutions to PnP through convex formulations.
Hmam and Kim~\cite{hmam2010optimal} first cast the PnP based on a quartic object-space error into an SDP, while explicitly incorporating chirality constraints to ensure physically valid camera poses.
More recently, CvxPnPL~\cite{agostinho2023cvxpnpl} reformulates PnP as a QCQP and derives an SDP relaxation that certifies global optimality under mild conditions, providing a unified convex formulation for absolute pose estimation.

\PAR{Moment-SOS relaxation.}
Moment-SOS methods formulate PnP as a POP solved via Lasserre relaxations.
Schweighofer and Pinz~\cite{schweighofer2008globally} first express PnP in the object-space domain using a quaternion-based rotation parameterization, and apply SOS relaxations to obtain certifiably optimal solutions.
Sun and Deng~\cite{sun2020certifiably} extend this formulation to incorporate line correspondences, leveraging a similar quaternion representation and solving the resulting polynomial optimization via Lasserre relaxations with global optimality guarantees.
Alfassi~\etal~\cite{alfassi2021non} further analyze the tightness of these convex relaxations and demonstrate that, in practical PnP instances, low-order Lasserre relaxations (second- or third-order) can be non-tight, thus motivating the use of higher relaxation levels to ensure global correctness.

\PAR{Other relaxations.}
Several methods employ alternative convex relaxations and certification techniques for PnP.

\noindent \emph{Convex hull and QCQP formulations.}
Horowitz~\etal~\cite{horowitz2014convex} exploit explicit LMI representations of the convex hulls of $\mathrm{SE}(2)$ and $\mathrm{SE}(3)$, enabling direct convex relaxations of pose estimation problems including PnP.
Building on this LMI foundation, Speciale~\etal~\cite{speciale2017consensus} derive additional LMI constraints to reformulate consensus maximization as a MIP within a BnB framework, optimally solving absolute pose estimation along with similarity and relative pose problems.
Khoo and Kapoor~\cite{khoo2016non} formulate the 2D-3D registration task as a mixed-integer nonlinear program and relax it into a convex SDP by leveraging the convex hull representation of $\mathrm{SO}(3)$~\cite{saunderson2015semidefinite}, enabling globally optimal solutions via IPMs.
Terzakis and Lourakis~\cite{terzakis2020consistently} propose SQPnP, which reformulates PnP as a QCQP and analytically bounds the feasible set into a finite number of convex regions guaranteed to contain the global optimum, solved via Sequential Quadratic Programming (SQP)~\cite{boggs1995sequential}.
Sun~\etal~\cite{sun2023efficient} present ACEPnP, which expresses PnP as a QCQP and derives an equivalent SOS relaxation, allowing for an analytical and certifiably optimal solution.

\noindent \emph{Optimality certification.}
Garcia-Salguero~\etal~\cite{garcia2024fast} propose a certifiable PnP framework that complements iterative or heuristic pose estimators with an optimality certifier based on Lagrangian duality.
Their method formulates the global optimality verification as an eigenvalue optimization problem solvable via efficient line search, thereby confirming whether a candidate solution is indeed globally optimal.

\PAR{Graduated non-convexity (GNC).}
GNC methods apply continuation strategies to achieve robust solutions without explicit combinatorial search.
Sun and Deng~\cite{sun2020certifiably} introduce GNC-SPnPL, which combines the GNC scheme with the Geman--McClure penalty under the Black--Rangarajan duality~\cite{black1996unification}, leading to a smooth reweighting-based solver that achieves robustness without sacrificing optimality guarantees.
Similarly, Yi~\etal~\cite{yi2025outliers} propose GNCPnP, applying the same duality principle to PnP using the TLS cost, yielding improved resilience to severe outliers.

%% file: sec/task_subsec/4_relpose.tex
\subsection{Relative Pose Estimation}
\label{subsec:relpose}

\PAR{Problem definition.}
Given a set of $N$ putative 2D-2D correspondences between $\{\bx_{0i} \!\in\! \bbR^2\}_{i=1}^N$ and $\{\bx_{1i} \!\in\! \bbR^2\}_{i=1}^N$ from two views, where $\bx_{0i}$ and $\bx_{1i}$ denote homogeneous image coordinates in the first and second cameras, \emph{relative pose estimation} seeks to estimate the relative camera pose $(\bR,\bt) \!\in\! \mathrm{SE}(3)$ between the two cameras.
Depending on whether camera calibration matrices are known, two closely related formulations arise:

\begin{itemize}
    \item \textbf{Uncalibrated case (Fundamental matrix).}
    When camera calibration matrices $\bK_0$, $\bK_1 \!\in\! \mathcal{K}$ the epipolar geometry is encoded by the fundamental matrix 
    $\bF \!\in\! \mathbb{R}^{3 \times 3}$:
    \begin{equation*}
    \label{eq:fundamental}
        \bx_{1i}^\top \bF \bx_{0i} = 0, \quad i = 1,\dots,N,
    \end{equation*}
    generally estimated by minimizing the sum of squared algebraic errors:
    \begin{equation}
    \label{eq:f-ls}
        \min_{\bF}
        \sum_{i=1}^N (\bx_{1i}^\top \bF \bx_{0i})^2,
        \tag{Fundamental}
    \end{equation}
    where $\bF$ has the rank $2$ with $\det(\bF) \!=\! 0$ and is defined up to scale.
    The relative pose can be recovered from $\bF$ up to a projective ambiguity.

    \item \textbf{Calibrated case (Essential matrix).}
    When camera calibration matrices $\bK_0$, $\bK_1 \!\in\! \mathcal{K}$ are known, image points are normalized as $\hat{\bx}_{0i} \!=\! \bK_0^{-1}\bx_{0i}$ and $\hat{\bx}_{1i} \!=\! \bK_1^{-1}\bx_{1i}$, yielding the essential matrix constraint:
    \begin{equation*}
    \label{eq:essential}
        \hat{\bx}_{1i}^\top \bE \hat{\bx}_{0i} = 0, \quad i = 1,\dots,N,
    \end{equation*}
    generally estimated by minimizing the sum of squared algebraic errors:
    \begin{equation}
    \label{eq:e-ls}
        \min_{\bE}
        \sum_{i=1}^N (\hat{\bx}_{1i}^\top \bE \hat{\bx}_{0i})^2,
        \tag{Essential}
    \end{equation}
    where $\bE \!=\! [\bt]_\times \bR$ has rank 2 with two equal nonzero singular values.
    The relative pose can be recovered from $\bE$ up to a cheirality ambiguity~\cite{longuet1981computer}.
\end{itemize}

Relative pose estimation is fundamental to SfM~\cite{schonberger2016structure}, visual localization~\cite{sarlin2019coarse}, and SLAM~\cite{campos2021orb}.

\PAR{Properties and local methods.}
Both formulations~\eqref{eq:f-ls} and~\eqref{eq:e-ls} are inherently nonconvex due to rank constraints and, in the calibrated case, the orthogonality structure of $\mathrm{SO}(3)$.
Classical approaches rely on local optimization methods that refine candidate poses through iterative minimization of reprojection error or Sampson error.
These solvers~\cite{zhang1998determining,kanatani2010unified,ma2001optimization,helmke2007essential,tron2017space,hartley1998textordfeminineminimizing,bartoli2004nonlinear,botterill2011refining,lui2013iterative,tron2014quotient} are computationally efficient and widely deployed in practical systems, but lack global optimality guarantees and are sensitive to initialization, noise, and outlier contamination.
Under heavy outlier rates or degenerate configurations (\eg planar scenes, low baseline), local methods can converge to suboptimal solutions, motivating the development of global optimization approaches.

\PAR{Closed-form and analytical solutions.}
Despite the nonconvexity, minimal formulations of relative pose estimation admit closed-form or analytical methods through polynomial constraint manipulation.

\noindent \emph{Fundamental matrix estimation.} The classical 8-point algorithm~\cite{hartley1997defense,longuet1981computer,faugeras1990motion} provides a linear solution by stacking epipolar constraints from eight correspondences into a homogeneous system, followed by rank-2 enforcement via SVD.

\noindent \emph{Essential matrix estimation.} The celebrated 5-point algorithm~\cite{nister2004efficient} exploits the rank-2 and equal-singular-value constraints on $\bE$ to derive a degree-10 polynomial system, yielding up to 10 candidate solutions from five correspondences.

\noindent Subsequent work refines the polynomial formulation~\cite{stewenius2006recent,kukelova2008polynomial,li2006five} and improves numerical stability through Gr{\"o}bner basis and action matrix techniques~\cite{kukelova2007two,li2020gaps}.
Alternative formulations leverage hidden variable techniques~\cite{kneip2012finding} or incorporate additional constraints from scene structure or motion priors~\cite{trawny2010global,zheng2013practical,ding2021globally,yu2023globally,yu2024globally}.

\PAR{Branch-and-Bound (BnB).}
BnB methods achieve globally optimal relative pose estimation by systematically searching the pose space while pruning suboptimal regions through tight bounds.

\noindent \emph{Early formulations and rotation search.}
Hartley and Kahl~\cite{hartley2007bnb} introduce BnB for essential matrix estimation by searching the rotation space under $L_\infty$ cost, providing the first deterministic globally optimal solver.
Building on this framework, Kim~\etal~\cite{kim2008motion,kim2009motion} extend the approach to multi-camera systems with non-overlapping fields of view, achieving globally optimal 6-DoF motion estimation by branching over rotation and solving for translation via LP.
Kneip and Lynen~\cite{kneip2013direct} propose an eigenvalue formulation equivalent to the algebraic error and solve it using BnB with efficient bounds derived from the rotation manifold structure.

\noindent \emph{Robust formulations with outlier rejection.}
Enqvist and Kahl~\cite{enqvist2009two} present the first BnB framework for two-view geometry estimation with explicit outlier rejection.
Yang~\etal~\cite{yang2014optimal} extend BnB to robust essential matrix estimation by branching over a minimal 5D essential matrix manifold parametrization, achieving inlier-set maximization with global optimality guarantees.
Bazin~\etal~\cite{bazin2014globally} simultaneously optimize over rotation and camera focal length, handling unknown intrinsics within the BnB framework.

\noindent \emph{Specialized geometry and reduced search spaces.}
Several methods exploit problem structure to reduce search complexity.
Li and Tsin~\cite{li2009globally} estimate the affine fundamental matrix from apparent contours with guaranteed global optimality.
Zheng~\etal~\cite{zheng2011branch} develop a branch-and-contract algorithm for globally optimal fundamental matrix estimation in outlier-free scenarios.
Fredriksson~\etal~\cite{fredriksson2015practical} focus on two-view translation estimation with known camera orientation, deriving efficient bounds for globally optimal solutions.
Their subsequent work~\cite{fredriksson2016optimal} addresses relative pose estimation with unknown correspondences, leveraging bipartite matching~\cite{hopcroft1973n} to create efficient bounds.

\noindent \emph{Constrained motion models.}
Recent work exploits motion constraints to further reduce search dimensionality.
Gao~\etal~\cite{gao2020efficient} develop globally optimal correspondence-free registration for planar ground vehicles under constrained non-holonomic motion.
Joo~\etal~\cite{joo2020globally} reduce the generic 6-DoF problem to a 3-DoF search for camera-on-selfie-stick configurations, significantly accelerating BnB convergence.
Liu~\etal~\cite{liu2021globally} exploit known gravity direction to decouple rotation and translation, employing nested BnB~\cite{yang2015go} for consensus maximization.
Similarly, Liu~\etal~\cite{liu2022globallyinlier} leverage general planar motion constraints to decouple the relative pose into relative rotation and translation, enabling simplified bounding strategies for inlier maximization.

\PAR{Shor's relaxation.}
SDP relaxations provide certifiable solutions to relative pose estimation through convex formulations of the algebraically constrained problems.

\noindent \emph{Certifiable solvers.}
Briales~\etal~\cite{briales2018certifiably} reformulate the eigenvalue-based formulation~\cite{kneip2013direct} and demonstrate that tight SDP relaxation can be achieved through redundant constraints, enabling certification of global optimality.
Zhao~\cite{zhao2020efficient} proposes a certifiably globally optimal solver for the over-determined essential matrix problem, later extending the approach to generalized essential matrix estimation~\cite{zhao2020certifiably} for multi-camera systems.
Garcia-Salguero~\etal~\cite{garcia2022tighter} derive additional redundant constraints to improve the tightness of SDP relaxations, building on the framework of~\cite{zhao2020efficient}.

\noindent \emph{Specialized formulations and certifiers.}
Li~\etal~\cite{li2020robot} address 2D robot-to-robot relative pose estimation by reformulating the squared-distance weighted least-squares problem as a QCQP and applying SDP relaxation.
Garcia-Salguero~\etal~\cite{garcia2021certifiable} develop an optimality certifier based on Lagrangian duality to verify whether candidate solutions are globally optimal, later extending this with a family of certifiers and a robust paradigm combining GNC with Black--Rangarajan duality~\cite{garcia2021fast}.
For scenarios with gravity priors, Garcia-Salguero and Gonzalez-Jimenez~\cite{garcia2023fast,garcia2024certifiable} propose iterative algorithms with duality-based certifiers for verifying global optimality.

\noindent \emph{Low-rank and factorized formulations.}
Karimian and Tron~\cite{karimian2023essential} show that under moderate noise levels, tighter SDP relaxations with fewer parameters can be achieved using Burer--Monteiro factorization~\cite{burer2003nonlinear} and Riemannian staircase algorithms~\cite{boumal2016non}.
Tirado-Gar{\'i}n and Civera~\cite{tirado2024correspondences} propose C2P, which ensures that estimated camera pose corresponds to valid 3D geometry and is globally optimal without requiring posterior rotation-translation disambiguation, building on the framework of~\cite{zhao2020efficient}.

\PAR{Moment-SOS relaxation.}
Moment and SOS relaxations address relative pose estimation through polynomial optimization frameworks with hierarchical convex relaxations.

\noindent \emph{SOS relaxations for fundamental matrix.}
Chesi~\etal~\cite{chesi2002estimating} first apply SOS convex relaxation based on homogeneous forms and LMIs to minimize algebraic error with explicit rank-2 constraints for fundamental matrix estimation.
Xiao~\cite{xiao2010new} develops convex LMI relaxation for fundamental matrix estimation using single-epipole parameterization, solving via standard LMI techniques.

\noindent \emph{SOS relaxations for essential matrix.}
Chesi~\cite{chesi2008camera} extends the SOS framework~\cite{chesi2002estimating} to essential matrix estimation for calibrated camera displacement.

\noindent \emph{Moment relaxations and robust formulations.}
Bugarin~\etal~\cite{bugarin2015rank} leverage determinant constraints and advocate moment relaxation hierarchies, solving sequences of increasing-size SDP problems with global optimality verified through moment matrix rank.
Cheng~\etal~\cite{cheng2015convex} combine polynomial optimization with rank minimization to robustly estimate fundamental matrix, introducing binary variables to handle outliers within the moment relaxation framework.

\noindent \emph{Refinements and extensions.}
Ren~\cite{ren2019non} applies SOS techniques to refine the formulations~\cite{briales2018certifiably,zhao2020efficient}, improving accuracy while reducing computational complexity by eliminating redundant constraints.
Sun~\cite{sun2022splp} incorporates line correspondences into SOS relaxations for relative pose estimation, improving estimation accuracy through additional geometric constraints.

\PAR{Other relaxations.}
Kim~\etal~\cite{kim2007visual} address visual odometry for multi-camera systems with non-overlapping fields of view by reducing motion estimation to an $L_\infty$-norm triangulation problem, solved globally via SOCP.

%% file: sec/task_subsec/5_reg.tex
\subsection{3D Registration}
\label{subsec:registration}

\PAR{Problem definition.}
Given $N$ putative correspondences between $\{\bp_i \!\in\! \bbR^3\}_{i=1}^N$ and $\{\bq_i \!\in\! \bbR^3\}_{i=1}^N$ from two sets of 3D points, \emph{3D registration} seeks to estimate the rigid transformation $(s, \bR, \bt) \!\in\! \pR \times \mathrm{SE}(3)$ to align them.
3D registration is generally formulated as a least-squares estimation:
\begin{equation}
\label{eq:reg-ls}
    \min_{s > 0,\,\bR \in \mathrm{SO}(3),\,\bt \in \bbR^3} \; \sum_{i=1}^N \| \bq_i - (s\bR \bp_i + \bt) \|_2^2.
    \tag{Registration}
\end{equation}

3D registration underpins applications in 3D reconstruction~\cite{choi2015robust}, SLAM~\cite{campos2021orb}, panorama stitching~\cite{meneghetti2015image}, and medical imaging~\cite{audette2000algorithmic}.

\PAR{Properties and local methods.}
\eqref{eq:reg-ls} is nonlinear and nonconvex due to the $\mathrm{SO}(3)$ constraint.
Classical approaches rely on alternating optimization between correspondence estimation and transformation estimation.
The Iterative Closest Point (ICP) algorithm~\cite{besl1992method} and its numerous variants~\cite{rusinkiewicz2001efficient} remain the workhorses of 3D registration, iteratively updating correspondences and solving least-squares alignment.
While computationally efficient, these local methods are sensitive to initialization, partial overlap, and outlier contamination, often converging to local minima under challenging conditions.

\PAR{Closed-form solutions.}
In the ideal setting with perfect correspondences and Gaussian noise, \eqref{eq:reg-ls} admits a closed-form solution via SVD.
Horn~\cite{horn1987closed}, Arun~\etal~\cite{arun1987least}, and Umeyama~\cite{umeyama1991least} independently derive SVD-based solutions that compute the optimal rotation and translation in closed form, providing globally optimal alignment when correspondences are correct.
Their global optimality is contingent on the quality of correspondence estimates.

\PAR{Branch-and-Bound (BnB).}
BnB methods guarantee global optimality by systematically partitioning the $\mathrm{SE}(3)$ search space and pruning infeasible regions using admissible bounds.

\noindent \emph{Early formulations and nested BnB.}
Olsson~\etal~\cite{olsson2006registration,olsson2008branch} introduce convex underestimators within BnB to handle point-to-point, point-to-line, and point-to-plane correspondences.
Li and Hartley~\cite{li20073d} jointly parameterize registration in terms of transformations and correspondences, addressing a simplified rotation-only case via BnB combined with Lipschitz optimization~\cite{hansen1995jaumard}.
The influential Go-ICP~\cite{yang2013go,yang2015go} globally optimizes the ICP $L_2$ error through nested BnB over $\mathrm{SE}(3)$, accelerated by local ICP refinement to provide tight upper bounds.
Campbell and Petersson~\cite{campbell2016gogma} extend this framework to Gaussian mixture alignment with GOGMA, enabling globally optimal distribution matching in $\mathrm{SE}(3)$.

\noindent \emph{Alternative problem formulations.}
Several methods reformulate registration to exploit specific structures.
Lian~\etal~\cite{lian2016efficient} reduce registration to a concave quadratic assignment problem solved via rectangular BnB with an efficient linear-assignment lower bound.
Straub~\etal~\cite{straub2017efficient} combine Bayesian nonparametric distribution estimates with decoupling of translation and rotation based on surface normal distributions.
Paudel~\etal~\cite{paudel2015robust,paudel2019robust} integrate BnB with SOS polynomial optimization to globally maximize the inlier set for point-to-plane and plane-to-plane correspondences.
Enqvist~\etal~\cite{enqvist2009optimal} establish pairwise constraints to reformulate 3D-3D and 2D-3D registration as graph vertex cover problems, employing BnB to obtain globally optimal correspondence matching and transformation estimation.
Parra~\etal~\cite{bustos2019practical} design a maximum clique algorithm that integrates tree search with efficient bounding and pruning based on graph coloring.

\noindent \emph{Decomposition strategies.}
Recent work emphasizes decomposing the 6-DoF problem into lower-dimensional subproblems to accelerate search.
Liu~\etal~\cite{liu2018efficient} exploit rotation-invariant constraints to enable BnB search over 3-DoF translation space before estimating optimal rotation.
Li~\etal~\cite{li2018fast} employ translation-invariant vectors to decompose the 6-DoF search into sequential 3-DoF rotation and translation searches, each solved by dedicated BnB, further proposing 3D Integral Volume to accelerate bound evaluation.
Cai~\etal~\cite{cai2019practical} introduce 4-DoF BnB for LiDAR-based registration, later extended by Li~\etal~\cite{li2023fast} with inertial measurement units (IMUs) to reduce search dimensionality.
Chen~\etal~\cite{chen2022deterministic} decompose the 6-DoF problem into two sequential 3-DoF subproblems: first solving for the 2-DoF rotation axis and 1-DoF translation along this axis via 3D BnB, followed by recovery of the remaining 1-DoF rotation angle and 2-DoF planar translation using nested BnB.
Building on this paradigm, Zheng~\etal~\cite{zheng2025robust} propose GMOR, which leverages Chasles' theorem~\cite{joseph2020alternative} to decouple rigid motion and applies two-stage BnB, first on the rotation axis then on the rotation angle, accelerated by interval stabbing~\cite{bustos2017guaranteed} and range maximum query (RMQ), enabling deterministic maximization of inlier overlap under high outlier ratios.

\noindent \emph{Row-vector and constrained formulations.}
Li~\etal~\cite{li2024efficient} propose an $L_{\infty}$-based formulation that reduces registration to three 2-DoF subproblems, each estimating a row vector of the rotation matrix, with residual 1-DoF translation obtained via interval stabbing.
Huang~\etal~\cite{huang2024scalable} extend this row-vector decomposition with TEAR, a two-stage framework that first estimates an initial row vector through 2-DoF BnB in spherical space combined with translation estimation via interval stabbing, then leverages inter-row orthogonality constraints to simplify subsequent search.
Xu~\etal~\cite{xu2024hierarchical} exploit the Atlanta world assumption to reduce the BnB search space via hierarchical translation-transformation decomposition.
Hu and Kneip~\cite{hu2021globally} jointly solve for symmetry plane parameters via BnB, enabling robust performance under extremely low overlap.

\noindent \emph{Acceleration and efficiency improvements.}
Several methods focus on accelerating BnB convergence through tighter bounds and efficient data structures.
Huang~\etal~\cite{huang2024efficient} introduce HERE, which integrates the spatial compatibility graph into BnB search, sampling valid correspondences from the graph and performing progressive outlier removal through three-stage BnB with interval stabbing.
Ego-ICP~\cite{liu2023chebyshev} employs Chebyshev surrogates to approximate alignment error within BnB, reducing computational cost.
Ivanov and Markgraf~\cite{ivanov2025fast} propose OptiPose, which integrates convex relaxations into BnB to bound the TLS objective via interval analysis.

\PAR{Shor's relaxation.}
SDP relaxations provide certifiable solutions to 3D registration by lifting the nonconvex problem into convex form.

\noindent \emph{Lagrangian duality and convex relaxations.}
Olsson and Eriksson~\cite{olsson2008solving} apply Lagrangian duality to enforce orthogonality of the transformation matrix for point-to-plane correspondences.
Chaudhury~\etal~\cite{chaudhury2015global} introduce convex SDP relaxations for globally optimal multiple point cloud registration.
Briales and Gonzalez-Jimenez~\cite{briales2017convex} exploit redundant quadratic rotation constraints to derive tight Lagrangian dual relaxations that handle point-to-point, point-to-line, and point-to-plane correspondences.
Iglesias~\etal~\cite{iglesias2020global} leverage Lagrangian duality to derive certificates that verify global optimality of candidate solutions.

\noindent \emph{Decoupled formulations and robust extensions.}
TEASER~\cite{yang2019polynomial,yang2020teaser} separates rotation, translation, and scale estimation.
The method employs a TLS cost with maximum clique selection~\cite{bustos2019practical} to aggressively prune outliers, then solves rotation via SDP relaxation while recovering translation and scale through adaptive voting schemes.
TEASER++~\cite{yang2020teaser}, a variant presented in the same work, replaces the SDP-based rotation estimation with GNC, significantly improving computational efficiency while maintaining robustness.
Both TEASER and TEASER++ serve as solvers and, using Douglas--Rachford Splitting (DRS)~\cite{yang2020one,henrion2012projection,bauschke1996projection,higham1988computing,combettes2011proximal,jegelka2013reflection}, as certifiable verifiers for candidate solutions from heuristics such as RANSAC.
GlobalPointer~\cite{liao2024globalpointer} decomposes the problem into two subproblems, each reformulated as an SDP relaxation and solved in an alternating fashion until convergence.

\PAR{Other relaxations.}
Ahmed and Chaudhury~\cite{ahmed2017global} reformulate registration as an SDP with explicit rank constraints, addressing it via a tailored nonconvex variant of the Alternating Direction Method of Multipliers (ADMM).

\PAR{Graduated non-convexity (GNC).}
Peng~\etal~\cite{peng2023convergence} incorporate GNC into an IRLS framework to enhance convergence and robustness in outlier-robust registration, analyzing theoretical convergence properties, and demonstrating improved performance under heavy contamination.

\PAR{Robust preprocessing.}
Parra and Chin~\cite{bustos2017guaranteed} extend the rotation-only outlier removal framework of~\cite{parra2015guaranteed} to full 6-DoF point cloud registration, integrating 1-D interval stabbing~\cite{de2008computational} to exploit 3-DoF rotation constraints and effectively reducing the problem dimensionality while guaranteeing elimination of incorrect correspondences.
Li~\etal~\cite{li2023qgore} propose QGORE, a quadratic-time variant that employs voting-based geometric consistency to efficiently compute tight upper bounds while preserving the robustness and optimality of the original method.

%% file: sec/task_subsec/6_ra.tex
\subsection{Rotation Averaging}
\label{subsec:rot-avg}

\PAR{Problem definition.}
Given a measurement graph $\mathcal{G} \!=\! (\mathcal{V},\mathcal{E})$ where each node $i \!\in\! \mathcal{V}$ represents an absolute camera rotation $\bR_i \!\in\! \mathrm{SO}(3)$, $i \!=\! 1,\dots,N$ and each edge $(i,j)\!\in\!\mathcal{E}$ encodes a noisy relative rotation measurement $\bR_{ij} \!\in\! \mathrm{SO}(3)$ satisfying $\bR_{ij} \!\approx\! \bR_i^{-1} \bR_j$, \emph{rotation averaging} seeks to estimate the absolute camera orientations $\{\bR_i\}_{i=1}^N$ up to a global gauge (\eg $\bR_1 \!=\! \bI$).
Rotation averaging is generally formulated to minimize a discrepancy over edges:
\begin{equation}
\label{eq:ra-ls}
    \min_{\bR_i\in\mathrm{SO}(3)} 
    \sum_{(i,j)\in\mathcal{E}} d_\bR(\bR_{ij}, \bR_i^{-1} \bR_j)^p,
    \tag{RA}
\end{equation}
where $d_\bR(\cdot, \cdot)$ is a distance function on $\mathrm{SO}(3)$ (\eg geodesic, chordal, axis-angle, or quaternion distance~\cite{hartley2013rotation}) with $p$-norm, measuring the deviation between measured and estimated relative rotations.

Under a matrix Langevin or von Mises--Fisher noise model~\cite{boumal2014cramer,chen2021cramer}, the maximum-likelihood form is:
\begin{equation}
\label{eq:ra-mle}
    \max_{\bR_i\in\mathrm{SO}(3)} 
    \ \sum_{(i,j)\in\mathcal{E}} \kappa_{ij}\,\trace(\bR_i \bR_{ij} \bR_j^{-1}),
    \tag{RA-MLE}
\end{equation}
with concentration parameters $\kappa_{ij}\!>\!0$.

Rotation averaging underpins SfM~\cite{schonberger2016structure}, multi-view reconstruction~\cite{govindu2006robustness}, and SLAM~\cite{campos2021orb}.

\PAR{Properties and local methods.}
Both formulations \eqref{eq:ra-ls} and \eqref{eq:ra-mle} are highly nonconvex due to manifold constraints ($\bR_i\!\in\!\mathrm{SO}(3)$) and the coupling induced by the measurement graph.
Classical approaches employ locally convergent iterative schemes, including IRLS~\cite{chatterjee2013efficient,chatterjee2017robust,gao2021incremental,li2020pushing,arrigoni2018robust,lee2022hara}, Weiszfeld-type procedures~\cite{hartley2011l1}, and related variants on the manifold $\mathrm{SO}(3)$~\cite{govindu2001combining,moakher2002means,hartley2013rotation,govindu2004lie,tron2012intrinsic,martinec2007robust,sidhartha2021all,govindu2006robustness,shi2020message,boumal2013robust,arrigoni2014robust,zhang2023revisiting}.
These methods are computationally efficient but sensitive to noise and outliers, typically requiring projection back onto $\mathrm{SO}(3)$ and not guaranteeing global correctness~\cite{wilson2016rotations}.
These limitations motivate global optimization techniques that provide certificates of optimality and principled robustness.

\PAR{Shor's relaxation.}
SDP relaxations for rotation averaging lift the problem by parameterizing $\bX \!=\! [\bR_1^\top, \dots, \bR_N^\top]^\top$ and optimizing over $\bZ \!=\! \bX \bX^\top$ with block-orthonormality constraints, yielding a connection-Laplacian view of \eqref{eq:ra-mle}.

\noindent \emph{Early SDP formulations and tightness analysis.}
Singer~\cite{singer2011angular} introduces an SDP relaxation for angular synchronization (\ie rotation averaging), providing one of the first rigorous theoretical guarantees for exact recovery under bounded noise.
Arie-Nachimson~\etal~\cite{arie2012global} derive an SDP relaxation that replaces determinant constraints with equivalent linear inequality constraints, proving theoretical tightness advantages over spectral methods~\cite{martinec2007robust}.
Fredriksson and Olsson~\cite{fredriksson2012simultaneous} develop a Lagrangian dual SDP formulation that certifies global optimality of candidate solutions at low computational cost, without solving a full SDP.
Bandeira~\etal~\cite{bandeira2017tightness} theoretically demonstrate that the SDP relaxation remains tight under relatively high noise, exactly recovering the optimal solution.
Zhong and Boumal~\cite{zhong2018near} further study tightness guarantees for 2D rotation averaging.
Brynte~\etal~\cite{brynte2022tightness} present a theoretical framework for analyzing the power of SDP relaxations for optimization over rotational constraints, investigating the connections between nonnegative and SOS polynomial cones and establishing fundamental relationships to the tightness of SDP relaxations.

\noindent \emph{Lagrangian duality and efficient solvers.}
Eriksson~\etal~\cite{eriksson2018rotation,eriksson2019rotation} apply Lagrangian duality with chordal distance, analytically proving no duality gap under explicit noise conditions (up to approximately $42.9^\circ$) by removing the determinant constraint.
They develop row-by-row block-coordinate descent (BCD) specialized to the dual, achieving $1$--$2$ orders of magnitude speedup compared to standard SDP solvers like SeDuMi~\cite{sturm1999using}.
Leveraging this dual and strong duality, Moreira~\etal~\cite{moreira2021rotation,moreira2024rotation} propose a primal-dual method with spectral updates, akin to recursive spectral initialization~\cite{arrigoni2016spectral}.
Parra~\etal~\cite{parra2021rotation} further avoid forming large SDP matrices by applying rotation-coordinate descent (RCD) directly on $\mathrm{SO}(3)$ while retaining dual certificates when strong duality holds.
Chen~\etal~\cite{chen2021hybrid} propose a practical pipeline that first solves a low-rank SDP relaxation via block-coordinate minimization (BCM)~\cite{tian2019block}, then refines estimates using IRLS~\cite{chatterjee2013efficient}, combining the accuracy of SDP relaxations with the efficiency of local optimization.

\noindent \emph{Manifold lifting and specialized parameterizations.}
Dellaert~\etal~\cite{dellaert2020shonan} introduce Shonan Rotation Averaging, which lifts the problem to higher-dimensional rotation manifolds $\mathrm{SO}(p)^N$ and combines a sequence of semidefinite relaxations with efficient local optimization.
The method provides SDP-derived certificates of global optimality while achieving the efficiency of local manifold methods through adaptive dimension lifting.
Barfoot~\etal~\cite{barfoot2025certifiably} develop a certifiably optimal framework employing the Cayley map to parameterize $\mathrm{SO}(3)$, enabling SDP relaxations for rotation and pose averaging.
Olsson~\etal~\cite{olsson2025certifiably} extend SDP formulations to incorporate anisotropic chordal distances, proposing stronger convex relaxations capable of certifiably finding the global optimum under directionally varying measurement uncertainties.

\PAR{Moment-SOS relaxation.}
The formulation in \eqref{eq:ra-mle} naturally admits a polynomial structure: orthogonality and determinant constraints ($\bR_i^\top \bR_i \!=\! \bI$, $\det(\bR_i) \!=\! 1$) as well as the objective are polynomial functions of the unknowns.
This enables the use of moment and SOS (Lasserre) relaxations~\cite{lasserre2001global}, which can provide certifiably optimal solutions for small- to medium-scale instances.
Ke~\etal~\cite{ke2014globally} demonstrate that single rotation averaging ($N \!=\! 1$)~\cite{hartley2013rotation} can be cast as a POP and solved optimally using low-order Lasserre relaxations, achieving exact solutions on modest problem sizes.
However, due to the rapid growth of moment matrices, these relaxations become computationally prohibitive for large-scale instances, limiting their practical applicability compared to Shor's relaxation-based approaches.

\PAR{Other relaxations.}
Several alternative convex formulations pursue global guarantees while maintaining computational tractability.
Hartley~\etal~\cite{hartley2010rotation} extend the notion of geodesic convexity to $\mathrm{SO}(3)$, introducing the concept of weak convexity and deriving theoretical results on the location of global minima of the rotation averaging cost.
Wang and Singer~\cite{wang2013exact} propose least unsquared deviation (LUD), showing that it admits an equivalent SDP relaxation solvable via the alternating direction augmented Lagrangian method (ADM)~\cite{wen2010alternating}.
Saunderson~\etal~\cite{saunderson2014semidefinite} and Reich~\etal~\cite{reich2017global} relax optimization over $\mathrm{SO}(d)$ to its convex hull, represented with semidefinite constraints, thereby broadening the scope of convex relaxations.
More recently, Shi~\etal~\cite{shi2022robust} introduce the DESC framework, which casts robust rotation averaging as a quadratic program leveraging cycle consistency, proving that its global optimum can exactly recover ground-truth structure under favorable conditions.

%% file: sec/task_subsec/7_ta.tex
\subsection{Translation Averaging}
\label{subsec:trans-avg}

\PAR{Problem definition.}
Given a measurement graph $\mathcal{G} \!=\! (\mathcal{V}, \mathcal{E})$ where each node $i \!\in\! \mathcal{V}$ represents an absolute camera position $\bt_i \!\in\! \bbR^3$, $i \!=\! 1,\dots,N$ and each edge $(i,j) \!\in\! \mathcal{E}$ encodes a noisy translation direction measurement $\bt_{ij} \!\in\! \mathcal{S}^{2}$ satisfying $\frac{\bt_j - \bt_i}{\|\bt_j - \bt_i\|_2} \!\approx\! \bt_{ij}$, \emph{translation averaging} seeks to estimate the absolute camera locations $\{\bt_i\}_{i=1}^N$ up to a global translation and scale.
Translation averaging is generally formulated to minimize the angular or Euclidean discrepancy between estimated and observed directions:
\begin{equation}
\label{eq:ta-ls}
    \min_{\bt_i \in \bbR^3} \sum_{(i,j) \in \mathcal{E}} \left\| \frac{\bt_j - \bt_i}{\|\bt_j - \bt_i\|_2} - \bt_{ij} \right\|_2^2.
    \tag{TA}
\end{equation}

Translation averaging underpins 3D reconstruction~\cite{choi2015robust}, SLAM~\cite{campos2021orb}, and SfM~\cite{schonberger2016structure,pan2024global}.

\PAR{Properties and local methods.}  
Translation averaging is significantly more challenging than rotation averaging because only pairwise directions (up to an unknown global scale) are observable, introducing both scale and gauge ambiguities.
\eqref{eq:ta-ls} is inherently nonconvex due to the normalization, and is further complicated by sensitivity to outliers and degenerate camera configurations (\eg collinear arrangements).
Early approaches~\cite{brand2004spectral,govindu2001combining,govindu2004lie,jiang2013global} cast the problem as least-squares minimization over linear equations derived from pairwise direction measurements, often employing spectral methods or iterative refinement.
While computationally efficient, these methods are prone to drift accumulation and do not guarantee global consistency, especially under noise, outliers, and scale ambiguities.
These limitations motivate convex relaxations and global optimization approaches that provide theoretical guarantees.

\PAR{Shor's relaxation.}
Ozyesil~\etal~\cite{ozyesil2015stable} propose the first certifiably correct formulation of translation averaging by introducing a convex SDP relaxation based on pairwise line measurements.
Their approach leverages parallel rigidity theory~\cite{whiteley1989matroid} to establish theoretical guarantees for exact recovery, and employs an augmented Lagrangian dual algorithm~\cite{wen2010alternating} to scale the SDP solver to large camera networks.

\PAR{Other relaxations.}
Several alternative convex formulations provide practical solutions to translation averaging through specialized relaxations.
Sim and Hartley~\cite{sim2006recovering} formulate translation averaging as a quasi-convex $L_\infty$ optimization solvable via SOCP.
Moulon~\etal~\cite{moulon2013global} improve practical efficiency by replacing SOCP with LP using simplified Euclidean-distance constraints, yielding substantial speedups in large-scale pipelines.
Ozyesil and Singer~\cite{ozyesil2015robust} recast the problem as a convex LUD formulation, solved iteratively through QP approximations to enhance outlier resilience.
Goldstein~\etal~\cite{goldstein2016shapefit} propose ShapeFit, an alternative convex SOCP formulation that demonstrates scalable optimization using the ADMM.
Zhuang~\etal~\cite{zhuang2018baseline} introduce an explicit normalization variable to obtain baseline-insensitive angular error, admitting efficient optimization via BCD.

\PAR{Translation averaging with known rotations.}
A closely related problem is the joint estimation of camera translations and 3D structure given known rotations, often termed \emph{SfM with known rotations} or \emph{known rotations problem}, and more recently referred to as \emph{Global Positioning}~\cite{pan2024global}.
This setting arises naturally when rotation averaging has been solved first, leaving translation and structure estimation as a separate subproblem.

\noindent \emph{Quasi-convex $L_\infty$ formulations.}
Early work formulates this task as quasi-convex $L_\infty$ optimization solvable via SOCP~\cite{hartley2004sub,kahl2005multiple,kahl2008multiple}.
Olsson~\etal~\cite{olsson2007efficient} demonstrate that the reprojection objective is pseudoconvex (stronger than quasi-convexity) and propose two solvers: an IPM based on KKT conditions, and a bisection scheme solving a sequence of SOCPs.
Seo and Hartley~\cite{seo2007fast} accelerate $L_\infty$ minimization by operating on carefully selected data subsets, while Agarwal~\etal~\cite{agarwal2008fast} reduce the number of SOCP iterations using a Newton scheme rooted in fractional programming.
Li~\cite{li2009efficient} introduces a reduction that decomposes large $L_\infty$ problems into a finite set of minimal ``primitive'' subproblems, which are solvable analytically or in closed form, greatly improving scalability.

\noindent \emph{Robust formulations.}
To enhance robustness against outliers, several works advocate convex $L_1$ formulations solvable by LP~\cite{ke2007quasiconvex,dalalyan2009l_1,olsson2010outlier,wen2018efficient}.
Sim and Hartley~\cite{sim2006removing} employ residual trimming within the quasi-convex optimization loop to explicitly handle contaminated measurements.

%% file: sec/task_subsec/8_tri.tex
\subsection{Triangulation}
\label{subsec:triangulation}

\PAR{Problem definition.} 
Given a set of 2D projections $\{\bx_i \!\in\! \bbR^2\}_{i=1}^N$ in $N$ views with known camera calibration matrices $\{\bK_i \!\in\! \mathcal{K}\}_{i=1}^N$ and camera poses $\{(\bR_i, \bt_i) \!\in\! \mathrm{SE}(3)\}_{i=1}^N$, \emph{triangulation} seeks to estimate the corresponding 3D point $\bX \!\in\! \bbR^3$.
Triangulation is generally formulated as a least-squares estimation minimizing the reprojection errors:
\begin{equation}
\label{eq:triangulation-ls}
    \min_{\bX \in \bbR^3} \sum_{i=1}^N \| \pi \big(\bK_i (\bR_i \bX + \bt_i) \big) - \bx_i \|_2^2,
    \tag{Triangulation}
\end{equation}
where $\pi\big([v_1;v_2;v_3]\big) \!:=\! [v_1/v_3 ; v_2/v_3]$ is the perspective projection.
Triangulation is a cornerstone problem in multi-view geometry and SfM~\cite{schonberger2016structure}, as it underlies 3D point estimation from correspondences and is repeatedly solved within bundle adjustment~\cite{triggs1999bundle}.

\PAR{Properties and local methods.}
\eqref{eq:triangulation-ls} minimizes geometric reprojection error over a single 3D point and is nonconvex due to the perspective projection $\pi(\cdot)$.
The problem is ill-conditioned under near-degenerate configurations (\eg small baseline, large noise, near-epipolar arrangements), where numerical difficulty increases significantly.
Iterative local optimization methods refine initial estimates by minimizing reprojection error through nonlinear least squares.
Kanatani~\etal~\cite{kanatani2008triangulation} develop an improved higher-order method that satisfies the epipolar constraint and converges to a local extremum of the reprojection error, extending the first-order optimal correction~\cite{kanatani2005statistical}.
These local methods are fast and effective under moderate noise, but remain sensitive to initialization and may converge to local minimum.

\PAR{Closed-form and analytical solutions.}
Despite the nonconvexity, triangulation admits closed-form or analytical solutions for small numbers of views through polynomial constraint manipulation and algebraic methods.
Direct linear triangulation (DLT)~\cite{hartley2003multiple} provides a fast closed-form solution by linearizing the problem into a homogeneous system solvable via SVD.
For the two-view case, Hartley and Sturm~\cite{hartley1997triangulation} demonstrate that optimal $L_2$ triangulation reduces to solving a sixth-degree polynomial, yielding an efficient closed-form procedure that minimizes geometric reprojection error.
Alternatively, for angular reprojection errors, closed-form optimal two-view solutions are derived by Oliensis~\cite{oliensis2002exact} and Lee and Civera~\cite{lee2019closed}.
For three-view triangulation, Gr{\"o}bner-basis solvers provide $L_2$-optimal estimation by formulating the problem as a system of polynomial equations solved via elimination templates~\cite{kukelova2013fast,byrod2007fast,stewenius2005hard,hedborg2014robust}.

\PAR{Branch-and-Bound (BnB).}
Agarwal~\etal~\cite{agarwal2006practical} and Kahl~\etal~\cite{kahl2008practical} propose globally optimal BnB solvers based on fractional programming~\cite{tawarmalani2001semidefinite,sturm1999using} for minimizing both the $L_2$-norm and robust $L_1$-norm of reprojection errors, applicable to multiview triangulation and other geometric vision problems.
Lu and Hartley~\cite{lu2007fast} develop a simpler BnB algorithm for $L_2$ triangulation, combining a simplified BnB tree with LP relaxations and iterative methods (\eg Newton or Gauss-Newton~\cite{bjorck2024numerical}), achieving average runtimes of 0.02 seconds per triangulation.
Josephson and Kahl~\cite{josephson2008triangulation} propose a unified framework for triangulating points, lines, and conics using similar BnB strategies~\cite{agarwal2006practical}.

\PAR{Shor's relaxation.}
SDP relaxations cast triangulation as QCQPs, enabling convex formulations with certification guarantees.
Li~\cite{li2010multi} develops an SDP approach for multi-view structure computation without explicitly estimating motion, leveraging graph rigidity theory~\cite{schulze2017rigidity,eren2002closing}.
Aholt~\etal~\cite{aholt2012qcqp} cast triangulation as a QCQP using bilinear and quadratic epipolar constraints, then relax it as an SDP solvable with polynomial-time certificates of global optimality.
Cifuentes~\cite{cifuentes2021convex} generalizes this approach by developing larger SDP relaxations for the closest rank-deficient matrix problem, proving exact recovery in low noise regimes and applying it to triangulation.
H{\"a}renstam-Nielsen~\etal~\cite{harenstam2023semidefinite} extend these formulations~\cite{aholt2012qcqp,cifuentes2021convex} by incorporating TLS cost functions, constructing two SDP relaxations, one based on epipolar constraints and another on fractional reprojection constraints, demonstrating strong performance under high noise and outlier contamination.

\PAR{Moment-SOS relaxation.}
Kahl and Henrion~\cite{kahl2005globally,kahl2007globally} propose a general framework based on a hierarchy of convex LMI relaxations~\cite{lasserre2001global} for computing globally optimal estimates in geometric vision problems, including triangulation.

\PAR{Other relaxations.}
Several works exploit alternative convex formulations for globally optimal triangulation.

\noindent \emph{Quasi-convex and SOCP formulations.}
A series of works reformulate multiview triangulation under the $L_\infty$ norm as quasi-convex optimization problems, enabling globally optimal solutions via LP or SOCP~\cite{hartley2004sub,kahl2005multiple,kahl2008multiple,ke2007quasiconvex,min2010infinity}.
Li~\cite{li2007practical} develops a practical algorithm for $L_\infty$ triangulation with LP that explicitly handles outliers while maintaining global optimality guarantees.
Zhang~\etal~\cite{zhang2010triangulation} extend quasi-convex $L_\infty$ triangulation to omnidirectional vision, developing three frameworks tailored to different panoramic camera models (mirror, sphere, and cylinder).
Kang~\etal~\cite{kang2014robust} combine SOCP-based optimization with optimal inlier selection for robust multiview $L_2$ triangulation, achieving resilience to outlier contamination.

\noindent \emph{Certification and approximation methods.}
Garcia-Salguero and Gonzalez-Jimenez~\cite{garcia2023certifiable} propose a fast and certifiable solver that approximates the QCQP feasible set with linear constraints, enabling efficient solutions via complete orthogonal decomposition (COD) or SVD, with solutions certified as optimal via the Hessian of Lagrangian multipliers from the dual problem.

%% file: sec/task_subsec/9_pgo.tex
\subsection{Pose Graph Optimization}
\label{subsec:pgo}

\PAR{Problem definition.}
Given a pose graph $\mathcal{G} \!=\! (\mathcal{V}, \mathcal{E})$ where each node $i \!\in\! \mathcal{V}$ represents an absolute camera pose $\bT_i \!=\! (\bR_i, \bt_i) \!\in\! \mathrm{SE}(3)$, $i \!=\! 1,\dots,N$ and each edge $(i,j) \!\in\! \mathcal{E}$ encodes a noisy relative pose measurement $\bT_{ij} \!=\! (\bR_{ij}, \bt_{ij}) \!\in\! \mathrm{SE}(3)$, \emph{pose graph optimization (PGO)} seeks to estimate the absolute camera poses $\{(\bR_i, \bt_i)\}_{i=1}^N$ that are maximally consistent with the relative constraints.
PGO is generally formulated as a maximum-likelihood estimation:
\begin{equation}
\label{eq:pgo-ls}
    \begin{aligned}
        \min_{\bR_i \in \mathrm{SO}(3), \bt_i\in\bbR^3} \sum_{(i,j) \in \mathcal{E}} 
            & \kappa_{ij} \, d_\bR(\bR_{ij}, \bR_i^{-1} \bR_j)^2
            + \\
            & \tau_{ij} \, \| \bt_j - (\bR_i \bt_{ij} + \bt_i)\|_2^2,
    \end{aligned}
    \tag{PGO}
\end{equation}
where $d_\bR(\cdot, \cdot)$ denotes a distance function on $\mathrm{SO}(3)$ (\eg geodesic, chordal, or quaternion distance), and $\kappa_{ij}, \tau_{ij}$ are confidence weights.
PGO is a fundamental back-end task in 3D vision, central to SLAM~\cite{campos2021orb} and SfM~\cite{schonberger2016structure}.

\PAR{Properties and local methods.}
\eqref{eq:pgo-ls} is nonlinear and nonconvex due to $\mathrm{SO}(3)$ constraints and the coupling between rotations and translations.
Classical approaches rely on local iterative optimization, typically Gauss-Newton or Levenberg-Marquardt on $\mathrm{SE}(3)$ manifolds~\cite{grisetti2011g2o,kummerle2011g}.
These methods are efficient and form the backbone of modern SLAM libraries (\eg g2o~\cite{grisetti2011g2o}, GTSAM~\cite{dellaert2022borglab}), but are sensitive to initialization and susceptible to convergence to poor local minima, particularly under high noise or outlier contamination.
Several heuristics improve convergence, including chordal distance relaxations~\cite{olson2006fast}, spectral initializations~\cite{martinec2007robust,carlone2015initialization,zhao2026dual}, and incremental smoothing approaches~\cite{kaess2008isam,kaess2012isam2,rosen2014rise}.
However, without global optimality guarantees, reliability remains a key challenge, motivating the development of global methods.

\PAR{Shor's relaxation.}
SDP relaxations have become particularly influential in PGO, providing certifiable solutions through convex formulations based on Lagrangian duality and synchronization frameworks.

\noindent \emph{Lagrangian duality and planar formulations.}
Carlone~\etal~\cite{carlone2015lagrangian,carlone2015duality} formulate PGO as an SDP dual problem and exploit Lagrangian relaxation to recover globally optimal solutions, providing the first certifiable guarantees for PGO.
Tron~\etal~\cite{tron2015inclusion} extend this framework by including determinant constraints in the dual derivation, demonstrating that relaxing determinant constraints does not cause certification failures.
Building on this duality perspective, Carlone~\etal~\cite{carlone2016planar} analyze the planar case, proving exactness conditions and offering the first rigorous certification guarantees in practice.
Briales and Gonzalez-Jimenez~\cite{briales2016fast} improve efficiency by establishing a smaller dual SDP problem for optimality verification, and later extend these ideas to 3D with a dual-based initialization strategy~\cite{briales2017initialization}, showing that even when relaxation is not tight, the recovered solution provides high-quality initialization for iterative refinement.

\noindent \emph{Complex domain formulations.}
Calafiore~\etal~\cite{calafiore2015pose,calafiore2016lagrangian} reformulate PGO in the complex domain, establishing an SDP relaxation whose tightness is characterized by the Single Zero Eigenvalue Property (SZEP).
Fan~\etal~\cite{fan2019efficient} introduce CPL-Sync to formulate planar PGO in the complex domain, combining SDP relaxations with Riemannian trust-region (RTR) methods~\cite{absil2008optimization,absil2007trust} and proving tightness guarantees under low noise while achieving significant computational efficiency.

\noindent \emph{SE-Sync and synchronization frameworks.}
Rosen~\etal~\cite{rosen2019se,rosen2020certifiably} introduce SE-Sync, demonstrating that the rotation subproblem of PGO can be lifted into an SDP yielding exact recovery when measurement noise is not too large.
SE-Sync employs the Riemannian Staircase~\cite{boumal2015riemannian}, which leverages symmetric low-rank Burer--Monteiro factorization~\cite{burer2003nonlinear} combined with truncated-Newton Riemannian trust-region optimization, enabling certifiably globally optimal solutions at speeds comparable to standard local methods (\eg Gauss-Newton~\cite{bjorck2024numerical} and Levenberg-Marquardt~\cite{more2006levenberg}).
Building upon SE-Sync, Briales and Gonzalez-Jimenez~\cite{briales2017cartan} propose Cartan-Sync, which directly relaxes the problem without first marginalizing translations, reformulating sparse SDP pose synchronization as Riemannian optimization on the Cartan motion group~\cite{ozyesil2018synchronization}.
Yu and Yang~\cite{yu2024sim} further generalize synchronization to the similarity group with SIM-Sync, bridging certifiable optimality with learned depth priors~\cite{piccinelli2024unidepth,yang2024depth,bochkovskii2024depth}, illustrating promising synergy between geometric optimization and data-driven approaches.

\noindent \emph{Robust formulations.}
Carlone and Calafiore~\cite{carlone2018convex} enhance robustness by incorporating three robust cost functions, least unsquared deviation, least absolute deviation, and Huber loss, into SDP relaxations, preserving global optimality while handling outliers.
Lajoie~\etal~\cite{lajoie2019modeling} adopt a TLS cost for PGO to address perceptual aliasing and outlier correlation in SLAM via discrete-continuous graphical models.

\noindent \emph{Extensions to SLAM.}
Liu~\etal~\cite{liu2012convex} formulate 2D pose SLAM as a QCQP and relax it into a convex SDP problem, providing early foundations for certifiable SLAM.
Holmes and Barfoot~\cite{holmes2023efficient} derive SDP-based global optimality certificates for landmark-based SLAM with three-dimensional landmark measurements, preserving polynomial structure.
Papalia~\etal~\cite{papalia2024certifiably} present CORA, a certifiably correct range-aided SLAM method following the SE-Sync framework, demonstrating robustness and certification in range-augmented scenarios.

\noindent \emph{Distributed methods.}
Tian~\etal~\cite{tian2021distributed} propose DC2-PGO to address multi-robot settings with a distributed certifiably correct algorithm that combines sparse semidefinite relaxation with Riemannian block-coordinate descent (RBCD), inspired by low-rank SDP factorizations~\cite{wang2017mixing,tian2019block,erdogdu2022convergence}, enabling scalable deployment in collaborative SLAM.
Thoms~\etal~\cite{thoms2025distributed} develop DCORA, the first distributed certifiable solver for range-aided SLAM, building on CORA~\cite{papalia2024certifiably} via the Riemannian Staircase for scalable multi-robot applications.

\PAR{Moment-SOS relaxation.}
Mangelson~\etal~\cite{mangelson2019guaranteed} formulate Pose-Graph SLAM and Landmark SLAM as POPs solvable via SOS convex relaxations.
They employ a sparse bounded-degree variant of SOS relaxation to solve these problems to global minima, providing certifiable guarantees.

\PAR{Other relaxations.}
Rosen~\etal~\cite{rosen2015convex} relax optimization over the special orthogonal group $\mathrm{SO}(d)$ to its convex hull, represented using semidefinite constraints, providing approximate global guarantees.
Papalia~\etal~\cite{papalia2022score} introduce SCORE, which formulates range-aided SLAM initialization as an SOCP, providing a certifiable geometry-aware initializer that substantially improves convergence and accuracy of PGO.

%% file: sec/task_subsec/10_ba.tex
\subsection{Bundle Adjustment}
\label{subsec:bundle-adjustment}

\PAR{Problem definition.}
Given a set of 2D observations $\{\bx_{ij} \!\in\! \bbR^2\}_{(i,j) \in \mathcal{V}}$ across a visibility set $\mathcal{V}$ relating $N$ cameras and $M$ 3D points, known camera calibration matrices $\{\bK_i \!\in\! \mathcal{K}\}_{i=1}^N$, and initial estimates of camera poses $\{(\bR_i, \bt_i) \!\in\! \mathrm{SE}(3)\}_{i=1}^N$ and 3D points $\{\bX_j \!\in\! \bbR^3\}_{j=1}^M$, \emph{bundle adjustment (BA)} seeks to jointly refine the camera parameters and 3D scene structure.
BA is generally formulated as a least-squares estimation minimizing the reprojection errors:
\begin{equation}
\label{eq:ba-ls}
    \min_{\substack{(\bR_i, \bt_i) \in \mathrm{SE}(3) \\ \bX_j \in \bbR^3}} \sum_{(i,j)\in\mathcal{V}} \|\pi \big(\bK_i (\bR_i \bX_j + \bt_i) \big) - \bx_{ij}\|_2^2,
    \tag{BA}
\end{equation}
where $\pi\big([v_1;v_2;v_3]\big) \!:=\! [v_1/v_3 ; v_2/v_3]$ is the perspective projection.
BA~\cite{triggs1999bundle} is the canonical back-end in SfM~\cite{pan2024global,schonberger2016structure} and SLAM~\cite{campos2021orb}, producing accurate reconstructions and calibrated camera poses.

\PAR{Properties and local methods.}
\eqref{eq:ba-ls} is intuitive to formulate but extremely difficult to optimize due to two main challenges:
(i) the problem is highly nonconvex, arising from both the $\mathrm{SE}(3)$ feasible set and the nonconvex perspective projection $\pi(\cdot)$, and
(ii) the scale can be enormous, with the number of cameras $N$ and 3D points $M$ ranging from hundreds to tens of thousands, making \eqref{eq:ba-ls} a large-scale, sparse, nonlinear, and nonconvex optimization problem.

Classical BA solvers exploit problem sparsity, \ie each observation connects a single camera to a single 3D point, to form a sparse measurement Jacobian.
The canonical algorithmic backbone is iterative local optimization (\eg Gauss-Newton~\cite{bjorck2024numerical} or Levenberg-Marquardt~\cite{more2006levenberg}) with Schur-complement marginalization of 3D points, yielding a reduced camera normal equation solvable efficiently using sparse direct or iterative linear solvers~\cite{triggs1999bundle,lourakis2009sba,agarwal2010bundle}.
Widely-used implementations (\eg SBA~\cite{lourakis2009sba}, g2o~\cite{grisetti2011g2o,kummerle2011g}, GTSAM/iSAM2~\cite{dellaert2022borglab,kaess2012isam2}, Ceres~\cite{agarwal2012ceres}) combine robust linear algebra (sparse Cholesky, multifrontal solvers, preconditioned conjugate gradient), damping strategies, careful relinearization, and automatic differentiation to scale to millions of observations in practice.
Despite their empirical success, local methods may converge to poor local minima when initialization is inaccurate or observations are heavily contaminated by outliers.

Due to the extreme scale and sparsity structure of bundle adjustment, the field has been dominated by local optimization methods for decades.
Global optimization methods for BA are a recent development and remain relatively rare compared to other geometric problems, reflecting the fundamental computational challenges of certifiable large-scale nonconvex optimization.

\PAR{Shor's relaxation.}
Direct SDP liftings of \eqref{eq:ba-ls} are intractable at scale due to the problem dimension, but recent formulations exploit factorization and convex relaxations to regain scalability.
Inspired by SE-Sync~\cite{rosen2019se} and SIM-Sync~\cite{yu2024sim}, Han and Yang~\cite{han2025building} relax scaled bundle adjustment as a convex SDP and solve it efficiently with Burer--Monteiro factorization~\cite{burer2003nonlinear} and a CUDA-based Riemannian trust-region optimizer~\cite{absil2008optimization,absil2007trust}, achieving certifiable global optimality at large scales while incorporating learned depth priors~\cite{piccinelli2024unidepth,yang2024depth,bochkovskii2024depth}.

\PAR{Other global methods.}
Unlike most geometric problems covered in this survey where alternative convex relaxations are available, bundle adjustment's extreme scale precludes direct relaxation approaches beyond the recent SDP work discussed above.
Instead, a complementary line of work develops initialization-free bundle adjustment using variable projection, factorization, and surrogate objectives, providing practical global methods without requiring careful initialization.
Hong~\etal~\cite{hong2016projective} demonstrate that projective bundle adjustment from arbitrary initializations becomes tractable using the Variable Projection (VarPro) algorithm~\cite{golub1973differentiation}, which analytically eliminates structure variables and optimizes only over camera parameters, providing a principled alternative to purely iterative reprojection minimization.
Building on this, Hong and Zach~\cite{hong2018pose} introduce pOSE, a pseudo object-space error defined as a convex combination of the object space error and the affine projection error.
As a surrogate objective for the BA cost, pOSE preserves the bilinear factorization structure of the residual and retains a wide basin of convergence, enabling BA to achieve global minima from arbitrary initializations.
Iglesias~\etal~\cite{iglesias2023expose} complement pOSE with a quadratic approximation of exponential regularization in expOSE, making it suitable for optimization with VarPro and yielding more accurate initialization-free BA.
Weber~\etal~\cite{weber2024power} adapt power series expansion to the VarPro algorithm and extend this to Riemannian manifold optimization~\cite{absil2008optimization}, enabling large-scale bundle adjustment without initialization.
Most recently, Olsson and Nilsson~\cite{olsson2025towards} combine pOSE with rotation averaging by incorporating pairwise relative rotation estimates into the objective function, enabling BA to converge to global optima from random initializations.

%% file: sec/8_discussion.tex
\section{Open Challenges and Future Directions}
\label{sec:dis}

Despite substantial progress in global solvers for 3D vision, several fundamental challenges remain that limit the practical adoption of these methods in real-world systems.
We identify three key areas for future research, each presenting both significant challenges and promising opportunities.

\PAR{Scalability and algorithmic advances.}
\textit{How can we develop the next generation of global solvers that scale to larger problems while maintaining optimality guarantees and robustness?}

\PAR{Integration with deep learning.}
\textit{How can we effectively combine data-driven priors from deep learning with certifiable global solvers, both to accelerate optimization and to provide guarantees for learned representations?}

\PAR{Standardized evaluation.}
\textit{How can we establish comprehensive benchmarks and evaluation protocols that enable fair comparison across methods and track progress toward practical deployment?}

\subsection{Scalability and Algorithmic Advances}
\label{subsec:scalability}

\PAR{Scalability challenges.}
The global optimality guarantees provided by BnB and convex relaxation methods come at significant computational cost, with complexity growing prohibitively for large-scale problems~\cite{yang2022certifiably,zhang2024accelerating}.
Addressing this scalability bottleneck while preserving certifiability remains a central challenge.

\noindent \emph{Scaling BnB.}
BnB methods can benefit from tighter bounds derived through problem-specific structure, adaptive branching strategies that focus computational effort on promising regions~\cite{zhang2024accelerating}, and decomposition techniques that exploit sparsity~\cite{wentz1978branch} or symmetry~\cite{campbell2017globally,hu2021globally}.
Learning-based approaches for predicting search priorities or selecting subdivision strategies offer potential acceleration while preserving global guarantees.

\noindent \emph{Scaling CR.}
Significant progress has been made in scalable convex relaxations through techniques such as block-coordinate descent~\cite{tian2019block}, random projections~\cite{angell2023fast}, incremental or distributed formulations~\cite{liao2024globalpointer,liao2025convex}, and GPU/TPU-parallel solvers~\cite{han2025building}, which reduce runtime and memory demands.
However, these advances do not readily extend to outlier-robust formulations, where problem size grows with the number of observations and the need to model consensus or robust costs introduces combinatorial structure.
Promising directions include learning-driven outlier detection~\cite{brachmann2019neural,yan2025hemora} to reduce problem size, noise-aware relaxations that maintain tractability under outliers, and hybrid pipelines that couple robust pruning with efficient continuous solvers.

\noindent \emph{Scaling GNC.}
GNC achieves superior scalability but lacks global guarantees.
A promising direction is to design hybrid pipelines where efficient GNC-style solvers are combined with lightweight optimality certifiers, balancing speed and certifiability~\cite{yang2020teaser,yang2022certifiably}.

\noindent \emph{Emerging paradigms.}
Quantum-inspired optimization~\cite{chin2020quantum,doan2022hybrid,yang2024robust,wang2026quantum} and neuromorphic computing offer potential for dramatic acceleration of global search, though their practical realization for geometric vision problems remains an open challenge.
Advances in these directions may ultimately enable certifiable solvers for online and safety-critical applications such as autonomous driving and medical robotics.

\PAR{Relaxation tightness.}
Convex relaxations play a pivotal role in bridging nonconvexity and tractable optimization~\cite{peng2022semidefinite,cifuentes2022local,dumbgen2024toward}.
However, generic SDP or SOS hierarchies are often too loose at low levels and computationally intractable at high levels.
Recent work has explored problem-specific tightening strategies, including exploiting algebraic invariants such as adjugate constraints for essential matrices~\cite{garcia2022tighter,tirado2024correspondences}, enforcing low-rank structures~\cite{boumal2015riemannian,boumal2016non,wang2017mixing,rosen2020scalable,burer2003nonlinear,papalia2024overview}, leveraging sparsity~\cite{lasserre2006convergent,dumbgen2024exploiting}, and adding redundant constraints~\cite{zhao2020certifiably,yang2020teaser}.
Future work may focus on adaptive hierarchies that selectively refine only critical regions of the solution space, dynamically balancing tightness and computational cost.

\subsection{Integration with Deep Learning}
\label{subsec:learning}

\PAR{Deep learning for global solvers.}
Deep learning offers opportunities to complement global optimization by providing data-driven priors and heuristics~\cite{holmes2025sdprlayers,sungatullina2024minbackprop,pyatov2024robust,talak2024outlier,yang2021self}, not to replace principled optimization but to guide and accelerate it.

For BnB, learned priors can identify and prune large portions of the search space early, significantly enhancing scalability.
For CR, learned priors can decouple complex problems into more tractable formulations~\cite{han2025building}, enabling more efficient optimization. Alternatively, if networks provide warm-start solutions~\cite{angell2023fast}, relaxations can serve as verifiers to certify their global optimality.
For GNC, learned priors can inform homotopy schedules or step-size strategies, improving convergence to near-global solutions.

\PAR{Global solvers for deep learning.}
An emerging trend in geometric vision is to leverage large-scale pretraining and foundation models (\eg VGGT~\cite{wang2025vggt}, DUSt3R~\cite{wang2024dust3r}) as feed-forward data-driven predictors.
These models can implicitly encode scene geometry and multi-view relationships from data, providing high-quality predictions without explicit optimization.
However, learned predictions may violate multi-view geometric consistency constraints (\eg epipolar geometry, triangulation consistency) and lack global optimality certificates.
One promising direction is to embed certifiable constraints into model architectures or loss functions.
A more integrated approach is to incorporate convex relaxations as differentiable layers within learning pipelines~\cite{holmes2025sdprlayers,agrawal2019differentiable,pineda2022theseus}, enabling global solvers to function as implicit layers that provide optimality certificates while allowing end-to-end gradient-based training.
This synergy between learning and optimization may enable robust and certifiable perception systems that combine the flexibility of data-driven methods with the reliability of global solvers.

\subsection{Standardized Benchmarks and Evaluation}
\label{subsec:bench}

The field lacks standardized benchmarks and evaluation protocols for comparing global solvers across different problem domains and scales, hindering systematic progress and fair comparison of methods.

\PAR{Comprehensive benchmark suites.}
Current evaluations often rely on synthetic data or small-scale scenarios, making cross-method comparison difficult and limiting insights into real-world performance.
Large-scale, realistic benchmarks~\cite{agarwal2010bundle,liao2024globalpointer} covering diverse geometric vision tasks (\eg pose estimation, registration, SLAM, SfM) with varying noise levels, outlier ratios, and problem scales are essential for measuring progress.
These benchmarks should include both controlled synthetic scenarios for reproducibility and real-world datasets for practical validation, spanning the spectrum from small-scale controlled experiments to large-scale deployment scenarios.

\PAR{Evaluation metrics beyond accuracy.}
Beyond solution quality~\cite{sturm2012benchmark}, certifiable methods require metrics that capture the optimality-efficiency trade-off and the practical value of certification.
Important dimensions include:
(i)~the fraction of instances solved to certified global optimality,
(ii)~certification success rate and relaxation tightness gaps, (iii)~runtime-accuracy curves across problem scales,
(iv)~robustness under adversarial perturbations or extreme outlier ratios, and
(v)~memory footprint and computational requirements for large-scale deployment.
Such multidimensional evaluation would better capture the practical trade-offs inherent in different global optimization approaches.

\PAR{Reproducibility and standardization.}
Establishing common evaluation protocols, including standardized data formats, reference implementations with consistent interfaces, and accessible computational infrastructure, will enable fair comparison and accelerate community-wide progress.
Public repositories of benchmark datasets, baseline solver implementations, and evaluation scripts would lower barriers to entry and facilitate systematic comparison of new methods against established approaches.

\medskip
\noindent
In summary, realizing the full potential of global solvers in practical vision systems requires coordinated advances across all three fronts: developing scalable algorithms that maintain guarantees at realistic problem scales, integrating data-driven and optimization-based approaches to leverage their complementary strengths, and establishing rigorous evaluation protocols that track progress toward practical deployment.
Success in these directions will enable the next generation of robust, certifiable perception systems for safety-critical applications in autonomous driving~\cite{geiger2012we}, surgical robotics~\cite{sunmola2025surgical}, aerial navigation~\cite{burri2016euroc}, and beyond.

%% file: sec/9_social_impact.tex
\section{Social Impacts}
\label{sec:social}

The deployment of global optimization methods in real-world systems raises critical questions beyond technical performance.
As these methods transition from research prototypes to safety-critical applications, autonomous vehicles~\cite{geiger2012we}, medical robotics~\cite{sunmola2025surgical}, infrastructure monitoring~\cite{wegner2016cataloging}, their societal impacts become increasingly important.
The certifiability provided by global solvers offers unique opportunities for building trustworthy AI systems, but also introduces responsibilities regarding deployment, validation, and ethical use.
This section examines the broader implications of global optimization for 3D vision, addressing safety considerations, transparency requirements, integration with learning-based systems, and ethical responsibilities.

\subsection{Safety-Critical Applications}

Certifiable global solvers play an increasingly critical role in safety-critical systems where failures can have severe consequences, including autonomous driving~\cite{geiger2012we}, surgical robotics~\cite{sunmola2025surgical}, and aerospace navigation~\cite{berton2024earthloc}.
In these contexts, decision errors may lead to injury, property damage, or loss of life, making formal guarantees highly desirable.
For example, globally optimal pose estimation and triangulation provide guarantees that perception modules are not misled by local minima, reducing the risk of catastrophic failures in downstream decision-making.

In contrast to heuristic methods~\cite{fischler1981random} that may silently fail in challenging conditions, such as low texture, extreme viewpoints, or adversarial perturbations, certifiable solvers can either guarantee correctness or explicitly signal when guarantees cannot be provided, enabling fail-safe system design.
As these systems move toward real-time deployment in uncontrolled environments, certifiable optimization can act as a crucial safeguard, providing guarantees that classical heuristics~\cite{fischler1981random} or purely learning-based methods~\cite{wang2024dust3r} cannot offer.
This capability becomes especially valuable in scenarios where human oversight is limited or impossible, such as deep-space exploration~\cite{gallet2024exploring} or remote surgical procedures~\cite{hu2024ophnet}.

\subsection{Trust, Transparency, and Accountability}

Beyond safety, global methods play an important role in improving transparency and accountability in automated decision systems.
Certifiable solvers produce not only solutions but also verifiable certificates of global optimality or explicit bounds on suboptimality, enabling practitioners to assess solution quality without extensive empirical validation.
This transparency is crucial in applications ranging from industrial inspection~\cite{bai2023vision} to infrastructure monitoring~\cite{wegner2016cataloging}, where reproducibility and verifiability are central to decision-making and liability assessment.

The interpretability of optimization-based methods also provides a level of transparency that is often lacking in purely data-driven systems~\cite{wang2025vggt}.
Unlike black-box learned models whose internal reasoning is opaque, optimization methods make explicit their objective functions, constraints, and solution procedures, facilitating debugging, validation, and post-deployment auditing.
This transparency becomes especially important in regulated domains where algorithmic decisions must be auditable and explainable, such as medical diagnosis support~\cite{chen2025multi} or structural safety assessment~\cite{wegner2016cataloging}.
The mathematical certificates provided by global solvers can serve as evidence of due diligence, supporting regulatory compliance and establishing accountability in case of system failures.

\subsection{Integration with Learning-Based Systems}

The integration of learning and optimization raises both opportunities and concerns for societal impact.
On one hand, data-driven warm-starts and learned priors can dramatically improve scalability, enabling global methods to be deployed in complex real-world systems where pure optimization would be computationally prohibitive.
On the other hand, reliance on large-scale training data introduces risks related to dataset bias, privacy, and fairness.
For instance, if learned outlier detectors or problem decompositions systematically fail on underrepresented populations or environments, the overall system may achieve certification on biased subsets while failing on critical edge cases.

A critical challenge is developing hybrid architectures where learned components enhance efficiency without compromising the certifiability of the overall system.
This requires careful validation to ensure that data-driven acceleration does not introduce unverified failure modes in safety-critical applications.
Particular attention must be paid to distribution shift: a learned component trained on one environment may fail silently when deployed in novel conditions, potentially invalidating any downstream certification guarantees.
Addressing these concerns requires not only technical solutions, such as uncertainty-aware learning and robust verification, but also careful documentation of training data characteristics and deployment domain assumptions.

\subsection{Computational Resources and Accessibility}

The computational intensity of global solvers raises considerations about resource access and environmental impact.
While certification provides valuable guarantees, the associated computational cost can limit accessibility to well-resourced organizations with substantial computing infrastructure.
Ongoing research into scalable algorithms (\cref{sec:dis}) aims to democratize access by reducing computational barriers, but attention must be paid to ensuring that advanced certifiable perception capabilities remain available beyond elite institutions.

Additionally, the energy footprint of compute-intensive global optimization must be considered alongside its benefits.
Future work should explore efficient algorithms and specialized hardware that provide certification guarantees while minimizing environmental impact, particularly as these methods scale to large-scale deployment in autonomous systems~\cite{yang2025instadrive} and robotics platforms~\cite{bono2024learning}.

\subsection{Ethical and Societal Considerations}

As global optimization methods transition from research to deployment, ethical considerations must guide their application.
While certifiable 3D reconstruction and localization enable beneficial applications, assistive robotics for healthcare, cultural heritage preservation through accurate digitization, disaster response and search-and-rescue operations, the same technologies could be misused for invasive surveillance, unauthorized tracking, or military targeting systems.

The dual-use nature of these technologies places responsibility on the research community to consider potential misuse during development.
This includes being transparent about capabilities and limitations, engaging with ethicists and policymakers to establish appropriate use guidelines, and prioritizing research directions that demonstrably benefit society while considering potential harms.
Publication of algorithmic advances should be accompanied by discussion of both beneficial applications and potential misuse scenarios, enabling informed societal discourse about appropriate deployment contexts.

Moreover, access to certifiable methods should be democratized to avoid concentration of advanced capabilities in a few well-resourced entities.
Open-source implementations, transparent evaluation benchmarks (\cref{subsec:bench}), and educational resources can help ensure that the benefits of robust, certifiable perception are broadly distributed rather than restricted to organizations with substantial computational infrastructure.
Efforts to improve algorithmic efficiency (\cref{subsec:scalability}) serve not only technical goals but also equity objectives by reducing barriers to adoption.

Ultimately, the technical guarantees provided by global solvers must be complemented by institutional and regulatory frameworks that govern their deployment, ensuring that innovation serves societal welfare while respecting individual rights and dignity.

\medskip
\noindent
In summary, addressing these societal dimensions is essential for responsible advancement of global optimization methods in 3D vision.
By proactively considering safety, transparency, fairness, and ethics alongside technical performance, the research community can guide these powerful tools toward applications that enhance human welfare while minimizing potential harms.
The path forward requires ongoing dialogue among researchers, practitioners, ethicists, policymakers, and affected communities to ensure that the pursuit of global optimality serves not only technical excellence but also societal good.

%% file: sec/10_conclusion.tex
\section{Conclusion}
\label{sec:conclusion}

Global solvers have evolved from theoretical curiosities to practical tools for certifiable 3D perception, offering provable guarantees that distinguish them from heuristic and learning-based alternatives.
This survey unifies the landscape through a systematic taxonomy of three core paradigms, Branch-and-Bound, Convex Relaxation, and Graduated Non-Convexity, examining their theoretical foundations, practical enhancements for robustness and scalability, and comprehensive deployment across ten fundamental geometric vision tasks from pose estimation to bundle adjustment.
Our comparative analysis reveals the fundamental trade-offs governing method selection: certifiable methods provide global guarantees at significant computational cost, while continuation approaches sacrifice guarantees for efficiency, with the optimal choice depending critically on application requirements, problem structure, and deployment constraints.
Realizing the full potential of global solvers requires coordinated advances in algorithmic scalability, integration with data-driven methods, and standardized evaluation protocols, alongside careful consideration of their societal impacts in safety-critical applications.
As 3D vision systems transition from controlled environments to real-world deployment, global solvers provide an essential foundation for building trustworthy, certifiable perception, bridging the gap between theoretical guarantees and practical reliability.

%% file: sec/acknowledgements.tex
\section*{Acknowledgments}
This work was supported by Spanish grants PID2021-127685NB-I00 and PID2024-155886NB-I00 and Arag{\'o}n grant T45\_23R.